%% file: main.tex
\pdfoutput=1
\documentclass{article} %
\usepackage[T1]{fontenc}
\usepackage{iclr2027_conference,times}
\input{math_commands.tex}

\usepackage{url}
\usepackage{graphicx}
\usepackage{wrapfig}
\usepackage{float}
\usepackage{booktabs}
\usepackage{multirow}
\usepackage{longtable}
\usepackage{placeins}
\usepackage{needspace}
\usepackage{etoc}
\usepackage{amsmath,amssymb}
\usepackage{xcolor}
\usepackage{colortbl}
\usepackage{hyperref}
\definecolor{linkblue}{RGB}{40,112,173}
\hypersetup{
    colorlinks=true,
    allcolors=linkblue,
    pdftitle={Purify Before You Align: Improving Representation Alignment for Diffusion Models},
    pdfauthor={Yingheng Wang, Yaoqiang Li, Yaqin Wu, Junwen Bai, Jiatao Gu, Christopher De Sa, Volodymyr Kuleshov}
}
\definecolor{projectlink}{RGB}{200,93,49}
\newcommand{\projecturl}[1]{{\hypersetup{urlcolor=projectlink}\href{#1}{#1}}}
\usepackage[most]{tcolorbox}
\usepackage[capitalise,nameinlink,noabbrev,sort&compress]{cleveref}

\definecolor{findingborder}{RGB}{91,124,130}
\definecolor{findingbg}{RGB}{248,250,250}
\definecolor{lappurple}{HTML}{7943C2}
\colorlet{lapshade}{lappurple!12}
\newcommand{\ours}{\cellcolor{lapshade}}
\newcommand{\sdev}[1]{\,{\scriptsize\ensuremath{\pm#1}}}

\newtcolorbox{findingbox}{
    enhanced,
    colback=findingbg,
    colframe=findingborder,
    boxrule=0.65pt,
    arc=1.5mm,
    left=2.2mm,
    right=2.2mm,
    top=1.2mm,
    bottom=1.2mm,
    before skip=5pt,
    after skip=7pt,
    fontupper=\small
}

\newtcolorbox{introhighlight}{
    enhanced,
    colback=findingborder!9!white,
    frame hidden,
    boxrule=0pt,
    sharp corners,
    borderline west={1.2pt}{0pt}{findingborder},
    left=10pt,
    right=10pt,
    top=7pt,
    bottom=7pt,
    before skip=9pt,
    after skip=9pt,
    fontupper=\normalsize\raggedright
}

\newcommand{\Rtwo}{R^2}
\newcommand{\sg}{\mathrm{sg}}

\newcommand{\MI}{\mathcal{I}}
\title{%
\fontsize{16.2}{19.5}\selectfont\raggedright
\texorpdfstring{\raisebox{-0.17em}{\includegraphics[height=1.12em]{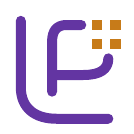}}\hspace{0.3em}}{}%
Purify Before You Align: Improving\\
Representation Alignment for Diffusion Models}

\author{%
\begin{minipage}[t]{\textwidth}
\centering
\textbf{%
Yingheng Wang\textsuperscript{1,*},
Yaoqiang Li\textsuperscript{†},
Yaqin Wu\textsuperscript{†},
Junwen Bai\textsuperscript{1}}\\[0.25em]
\textbf{%
Jiatao Gu\textsuperscript{2},
Christopher De Sa\textsuperscript{1},
Volodymyr Kuleshov\textsuperscript{1}}\\[0.75em]
{\normalfont
\textsuperscript{1}Cornell University
\qquad
\textsuperscript{2}University of Pennsylvania}
\end{minipage}%
}

\newcommand{\authornotes}{
    \begingroup
    \renewcommand{\thefootnote}{}
    \footnotetext{
        \textsuperscript{*}Corresponding author:
        \texttt{yingheng@cs.cornell.edu}.
    }
    \footnotetext{
        \textsuperscript{†}Independent researcher.
    }
    \footnotetext{
        Code: \projecturl{https://github.com/yingheng-wang/LAP}
    }
    \endgroup
}

\iclrfinalcopy

\begin{document}
\vspace*{-24pt}
\begingroup
\setlength{\tabcolsep}{0pt}
\maketitle
\endgroup
\authornotes
\lhead{Preprint}
\etocdepthtag.toc{main}

\vspace{-6pt}
\begin{abstract}
Representation alignment improves and accelerates diffusion training by encouraging intermediate features to match a pretrained visual encoder. However, standard alignment matches pretrained representations without distinguishing what information may be more useful for generation. We introduce \emph{local appearance purification} (LAP), which improves generation quality and makes encoders such as MAE more effective alignment targets by removing locally predictable content and aligning to the residual. LAP modifies only the alignment target, and requires no changes to diffusion training or inference.
Through controlled interventions within a fixed encoder, we show that effective alignment depends on what information the target emphasizes, not simply how much image information it contains.
Our theory gives a variational interpretation of standard alignment and, for fixed local predictors, a sum of token-wise information bounds conditioned on local appearance. To understand why purification benefits some encoders more than others, we derive an exact constant-readout test that measures how well the target can be matched without using the diffusion model's hidden state. LAP improves generation quality across five pretrained representations (MAE, AIMv2, CLIP, DINOv2, DINOv3), with the largest gain for MAE: on ImageNet-256 with SiT-B/2, it reduces FID from $53.1$ to $42.3$. Its gains persist with stronger alignment recipes and tokenizers. Computed before diffusion training, the constant-readout score predicts the benefit of purification (Spearman $\rho=0.94$ across 13 encoders).
Code and checkpoints are on the project page:\par\vspace{-\parskip}
{\centering\projecturl{https://yingheng-wang.com/LAP}\par}
\end{abstract}

\begin{figure}[H]
    \centering
    \vspace{-15pt}
    \includegraphics[width=\textwidth]{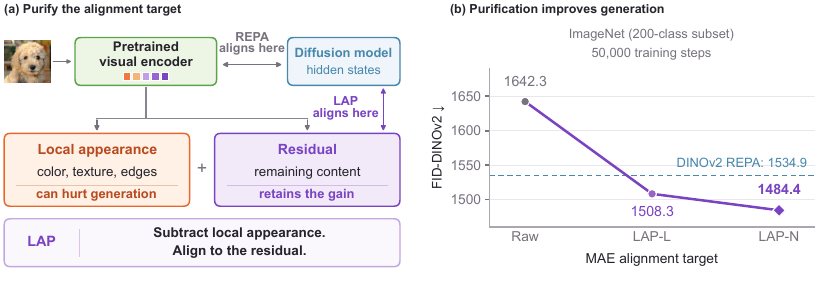}
    \vspace{-30pt}
    \caption{
\textbf{Purification improves generation by changing the alignment target.}
(a) LAP subtracts a local prediction and aligns to the residual.
(b) Both LAP-L and LAP-N improve MAE-based generation for SiT-B/2 and surpass DINOv2 REPA in the development setting.
}
    \label{fig:what_matters}
    \vspace{-10pt}
\end{figure}

\section{Introduction}
\label{sec:intro}

\begin{figure}[!t]
    \centering
    \setlength{\parskip}{0pt}
    \setlength{\abovecaptionskip}{5pt}
    \setlength{\belowcaptionskip}{0pt}
    \vspace{-5pt}
    \includegraphics[width=\textwidth]{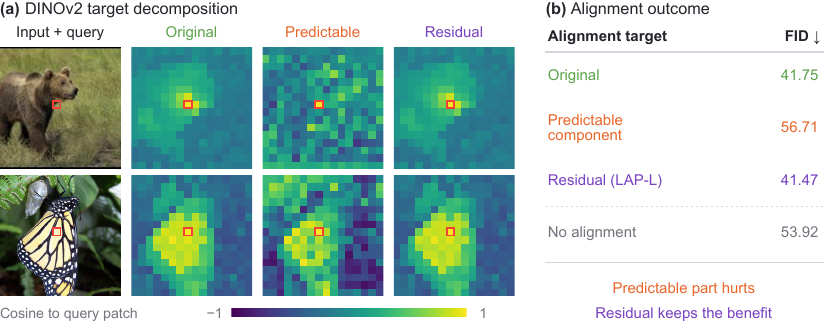}
    \vspace{-18pt}
    \caption{\textbf{One encoder, opposite alignment effects.}
    (a) DINOv2 features split into a locally predictable component (about 2\% of centered feature variance and has rank at most 17) and a residual. Each map shows the similarity of every patch to the red-boxed patch (\Cref{app:controls}).
    (b) Aligning to the locally predictable component hurts generation, whereas the residual (LAP-L) keeps the benefit.
    }
    \vspace{-14pt}
    \label{fig:motivating}
\end{figure}

Representation alignment has become part of the standard recipe for training diffusion models. REPA~\citep{yu2025repa} encourages intermediate denoiser features to match those of a frozen pretrained encoder, substantially accelerating training and improving sample quality. A growing line of work builds on this idea~\citep{leng2025repae,wang2025haste,reg2025,varepa2026}. However, one important choice remains largely empirical: which encoder should provide the alignment target? Different pretrained encoders lead to very different generation quality, and we still lack a clear principle for what makes a representation useful for alignment.

The puzzle is sharpest for MAE~\citep{he2022mae}. Since diffusion models learn to recover clean images from noise, features that preserve reconstructive information might seem especially suitable. MAE features predict clean-image content far better than DINOv2~\citep{oquab2023dinov2}, yet yield smaller generation gains in our development comparison (\Cref{fig:encoder_utility}). Recent work shows that target processing matters, including spatial operations in iREPA~\citep{singh2025irepa} and other target transformations~\citep{pixelrepa2026,repadog2026}. We ask a more basic question:

\begin{introhighlight}
\vspace{-5pt}
\emph{What information should an alignment target ask the denoiser to match?}
\vspace{-5pt}
\end{introhighlight}

To answer this, we hold the pretrained encoder fixed and intervene on the information selected for alignment. For DINOv2, aligning only to the locally predictable component performs worse than no alignment, while its residual retains the full benefit (\Cref{fig:motivating}). Likewise, making the aligned representation more predictive of the denoising target improves prediction but worsens generation (\Cref{sec:f2}). Together, these results show that effective alignment depends on what information the target emphasizes, not simply how much image information it contains.

These results suggest a simple view: the denoising objective already provides a strong reconstruction signal, so alignment is most useful when it emphasizes information beyond what denoising already encourages. This view also suggests a simple fix without replacing the encoder. \emph{Local appearance purification} (LAP) removes target content predictable from local appearance and aligns the denoiser to the residual, using linear or nonlinear local predictors (\Cref{fig:framework}). Across five visual teachers (MAE, AIMv2, CLIP, DINOv2, DINOv3), LAP improves generation quality, with the largest gains for MAE. On ImageNet-256 with SiT-B/2, LAP reduces MAE-based FID from $53.1$ to $42.3$. Its gains persist when combined with stronger alignment recipes and tokenizers, showing that target purification complements improvements elsewhere in the alignment pipeline. Beyond natural images, LAP could make self-supervised reconstructive encoders useful alignment teachers in domains such as medical and scientific imaging, where suitable discriminative teachers may be unavailable or poorly matched to the data.

We also develop an information-theoretic view of purified alignment. Standard alignment maximizes a variational lower bound on the information that the hidden state carries about the teacher target. For a fixed local predictor, purified alignment instead bounds a sum of token-wise information quantities, each conditioned on that token's local appearance. However, a high alignment score need not imply an informative hidden state: if target tokens share a common direction, even a constant readout that ignores the hidden state can achieve a high cosine score. We derive the exact optimum of this constant readout in closed form, giving a diagnostic that can be computed before diffusion training. In a matched comparison of 13 encoders, this score strongly predicts the benefit of purification (Spearman $\rho=0.94$), supporting its use to identify teachers likely to benefit from LAP (\Cref{sec:target_diagnostics}).

\textbf{Contributions.}
(1) Through fixed-encoder interventions, we show that reconstruction predictability alone does not determine alignment quality. In particular, locally predictable content can hurt generation even when it is easy to match.
(2) We introduce LAP, a family of target transformations that removes locally predictable content before alignment. LAP improves generation across multiple visual teachers, composes with stronger alignment recipes and tokenizers, and requires no encoder change or inference overhead.
(3) We develop an information-theoretic view of raw and purified alignment and derive a closed-form constant-readout diagnostic for alignment achievable without the hidden state. Computed before diffusion training, this score predicts the benefit of purification across 13 encoders.

\vspace{-5pt}
\section{Background}
\label{sec:setup}

\vspace{-5pt}
\subsection{Representation Alignment in Diffusion Training}
\label{sec:prelim}

We follow the latent flow-matching setup of REPA~\citep{yu2025repa,ma2024sit,peebles2023dit,lipman2023flow,albergo2023interpolants}. A frozen SD-VAE~\citep{rombach2022ldm} encodes an image $I$ into a clean latent $X$. The denoiser receives
\(
x_t=(1-t)X+t\varepsilon,~\varepsilon\sim\mathcal{N}(0,\mathbf{I}),
\)
and predicts the velocity $v^*=\varepsilon-X$ with the flow-matching loss
\(
\mathcal{L}_{\mathrm{flow}}
=
\mathbb{E}\left\|v_\theta(x_t,t)-v^*\right\|^2
\).
REPA adds supervision from a frozen visual encoder $f$: a trainable projector $g_\psi$ maps the hidden state at block $l$ to an aligned representation $a_t=g_\psi(h_t^l)$ and matches it token-wise to features of the clean image,
\begin{equation}
\mathcal{L}
=
\mathcal{L}_{\mathrm{flow}}
+
\lambda_a\mathcal{L}_{\mathrm{align}},
\qquad
\mathcal{L}_{\mathrm{align}}
=
-\mathbb{E}\left[
\frac{1}{N}\sum_{i=1}^{N}
\cos\!\left(a_t^{(i)},y^{(i)}\right)
\right],
\label{eq:align_loss}
\end{equation}
where $N$ is the number of patch tokens. Standard REPA uses the raw encoder output $y=z=f(I)$. The raw or purified teacher target occupies the same slot $y$. See \Cref{fig:framework} for illustration.

\vspace{-5pt}
\subsection{Measuring Representation Content and Use}
\label{sec:diagnostics}

To understand alignment, we distinguish the information available in a representation from the information the denoiser actually uses. We measure the former with linear probes and the latter with hidden-state subspace ablations.
For features $U$ and responses $V$, let $\Rtwo(U\!\to\!V)$ denote the held-out coefficient of determination of a ridge regression,
\begin{equation}
\Rtwo(U\!\to\!V)
=
1-\frac{\sum_{\mathrm{eval}}\lVert V-\hat W^\top U-\hat b\rVert^2}
{\sum_{\mathrm{eval}}\lVert V-\bar V\rVert^2},
\label{eq:probe_r2}
\end{equation}
where $(\hat W,\hat b)$ is fit on a separate set of images from those used for evaluation, and $\bar V$ is the evaluation-set mean.
Let $X_p$ and $v_p^*$ denote the patchified clean latent and velocity target. We use three probes. \emph{Prediction utility}, $\Rtwo(y\!\to\!X_p)$, measures how much clean-image information is available in the frozen alignment target and can therefore be evaluated before diffusion training. \emph{Presence}, $\Rtwo(h_t^l\!\to\!z)$, measures how much of the teacher representation can be decoded from the denoiser hidden state. \emph{Velocity sufficiency}, $\Rtwo(a_t\!\to\!v_p^*)$, measures how predictive the aligned representation is of the denoising target.

Linear probes establish that information is decodable, but not that the denoiser uses it. To measure reliance, we ablate the hidden-state subspace associated with the aligned representation. Let $\Pi$ project onto the span of the top $r$ left singular vectors of the probe from $h_t^l$ to $a_t$, and replace
\(
h_t^l
\leftarrow
(\mathbf I-\Pi)h_t^l+\Pi\bar h_t,
\)
where $\bar h_t$ is the mean hidden state in the same noise-level bin. Let $\mathcal E_\Pi$, $\mathcal E_{\mathrm{full}}$, and $\mathcal E_0$ denote the velocity MSE after subspace ablation, full-layer mean filling, and no ablation, respectively. We normalize the effect of the subspace ablation by the effect of ablating the full layer:
\(
U_\Pi
=
(\mathcal E_\Pi-\mathcal E_0)/
(\mathcal E_{\mathrm{full}}-\mathcal E_0).
\label{eq:causal_use}
\)
We compare $U_\Pi$ against rank-matched random subspaces.
The detailed experimental setup for these diagnostics is summarized in \Cref{sec:scales}.

\vspace{-5pt}
\section{Why Purify the Alignment Target?}
\label{sec:mechanism}

\begin{wrapfigure}[]{r}{0.4\textwidth}
\centering
\vspace{-20pt}
\includegraphics[width=\linewidth]{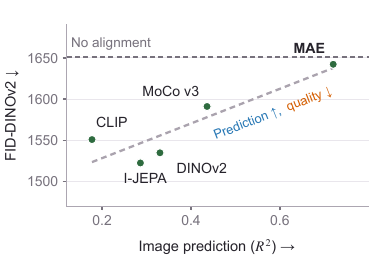}
\vspace{-25pt}
\caption{Better clean-image prediction generally accompanies worse generation. Dev-200/50K, FID-DINOv2 on 10K samples. The dashed line is a descriptive linear fit (\Cref{tab:ladder}).}
\label{fig:encoder_utility}
\vspace{-10pt}
\end{wrapfigure}

Across MAE, MoCo v3, DINOv2, I-JEPA, and CLIP, better clean-image prediction generally accompanies worse generation (\Cref{fig:encoder_utility}). MAE predicts clean-latent patches best but yields smaller generation gains than the other four encoders in this development comparison. This raises a question about what the denoiser needs from its teacher: reconstruction information, or a different constraint on its intermediate representation?
A cross-encoder comparison cannot identify the cause because architecture, data, and pretraining objective change together. We therefore hold DINOv2 fixed and intervene in two ways. First, we encourage the aligned representation to predict the denoising target more accurately. Second, we separate locally predictable target content from its residual and align to each alone. These interventions test the role of reconstruction content without replacing the encoder.

\vspace{-5pt}
\subsection{More Velocity-Predictive Aligned Representations Can Hurt Generation}
\label{sec:f2}

We first encourage the aligned representation $a_t$ to predict the denoising target. A small auxiliary head $r_\omega$ adds the loss $\lambda_s\mathbb{E}\|r_\omega(a_t,t)-v^*\|^2$ to REPA. At full scale, it raises held-out velocity-prediction $\Rtwo$ from $0.11$ to $0.53$, yet worsens FID-50K from $41.75$ to $46.79$ (\Cref{tab:f2,tab:mech_full}). Better prediction through the aligned representation therefore does not translate into better generation (loss sweep in \Cref{fig:causal_tests}). Applying the same auxiliary loss through a fixed random or separately learned projection also worsens generation in development controls, so this degradation is not specific to the aligned projection (\Cref{app:controls}).
The unmodified REPA model shows why prediction alone is insufficient. DINOv2 features are decodable from the hidden state (presence $\Rtwo=0.207$), but the denoiser relies only weakly on the aligned subspace: ablating it increases velocity error by only $12.2\%$ of the full-layer effect, compared with $22.7\%$ for rank-matched random subspaces and $14.0\%$ for the untrained projector subspace in the unaligned model (\Cref{fig:causal_tests}). Decodability does not imply reliance. This ablation tests reliance on the specific aligned subspace, but does not imply that all reconstruction-related information is unimportant.

\vspace{-5pt}
\subsection{Locally Predictable Content Does Not Drive the Alignment Benefit}
\label{sec:dissociation}

We next ask whether locally predictable content is responsible for the benefit of a strong alignment target. We use an affine predictor from clean-latent patches to DINOv2 tokens to decompose each target feature into a locally predictable component and a residual, then align to each component separately while keeping the encoder and training recipe fixed. The predictable component has rank at most 17 and accounts for about 2\% of DINOv2's centered variance at full scale (\Cref{app:controls}). The residual corresponds to the linear LAP variant defined in \Cref{sec:purifier}.
The two components have sharply different effects (\Cref{fig:motivating}). Aligning only to the locally predictable component worsens FID from $53.92$ without alignment to $56.71$, whereas aligning to the residual reaches $41.47$, preserving the gain of raw DINOv2 at $41.75$. This difference is not explained by alignment difficulty: the predictable component is easier to match despite producing worse generation. 
A fixed random rank-17 projection remains within the REPA seed range in a development control, while a frozen dataset-level predictable component loses the benefit. These controls argue against rank reduction or batch refitting alone explaining the result (\Cref{app:controls}).
Together with the previous result, this supports a view in which alignment is useful not because it provides more reconstruction information, but because it can shape intermediate representations beyond what denoising already encourages. This motivates purification.
We next formalize this idea as a target construction, develop its information-theoretic interpretation, and introduce a diagnostic for evaluating alignment targets.

\begin{figure}[!t]
    \centering
    \setlength{\parskip}{0pt}
    \setlength{\abovecaptionskip}{5pt}
    \setlength{\belowcaptionskip}{0pt}
    \includegraphics[width=1\linewidth]{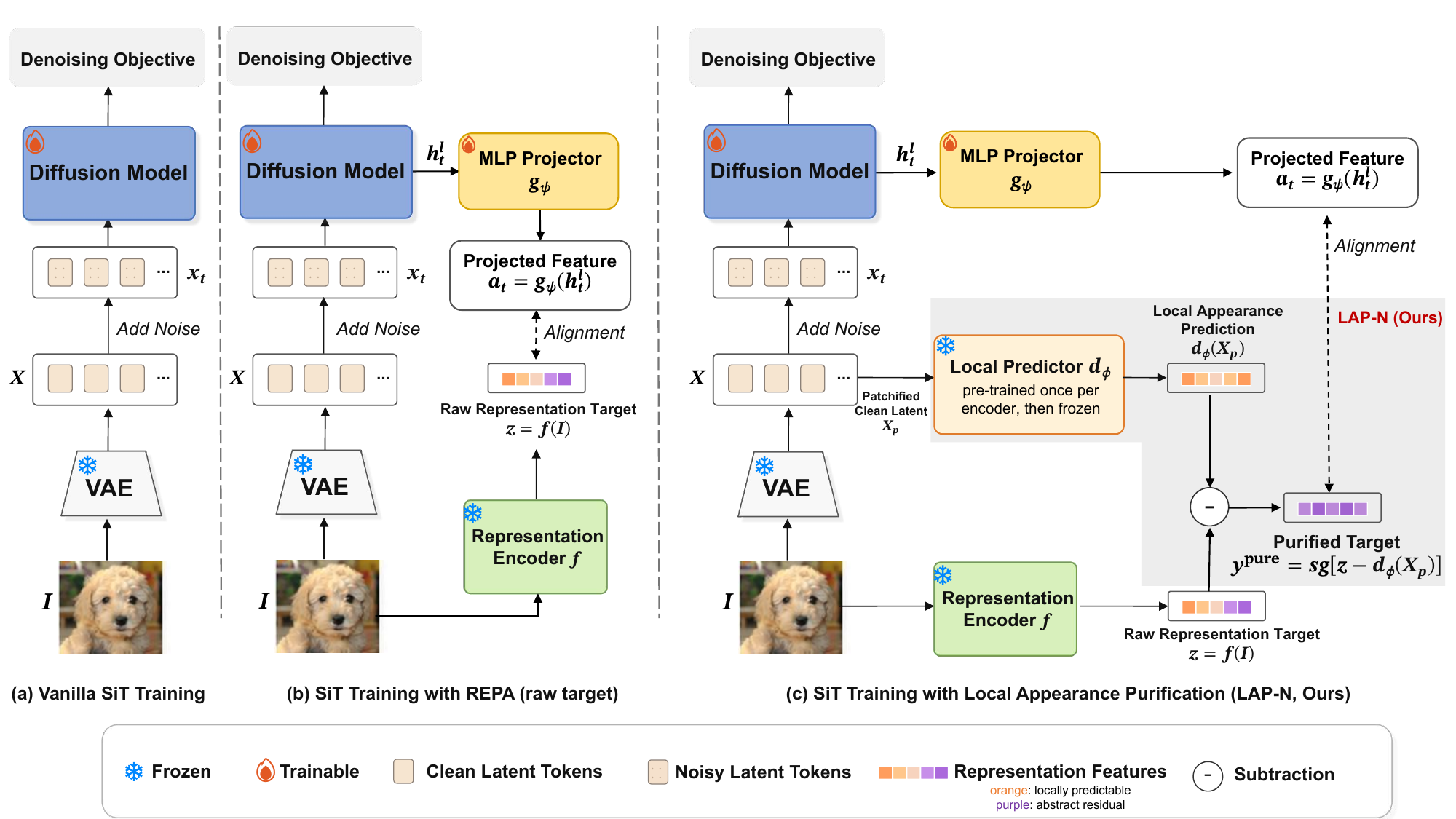}
    \vspace{-10pt}
    \caption{
    \textbf{Our LAP-N framework.}
    Compared with SiT training with REPA, LAP-N modifies only the alignment target by subtracting the reconstruction content predicted by a frozen local predictor $d_{\phi}$. Training and inference are otherwise unchanged.
    }
    \vspace{-10pt}
    \label{fig:framework}
\end{figure}

\vspace{-5pt}
\section{Local Appearance Purification for Representation Alignment}
\label{sec:method}

The results above suggest a simple design principle: alignment should emphasize information that complements the reconstruction signal already provided by denoising. We turn this idea into \emph{local appearance purification} (LAP), which removes locally predictable content from a target representation and aligns the denoiser to the residual. To analyze this objective, we develop a variational interpretation of what alignment encourages the hidden state to encode and a constant-readout diagnostic that measures how much alignment can be achieved even without using the hidden state.

\vspace{-5pt}
\subsection{Local Appearance Purification}
\label{sec:purifier}

We define locally predictable content as the part of an encoder representation predictable from local clean-image appearance. Let $z=f(I)$ be the frozen encoder features and $X_p$ the patchified clean latent. We fit a predictor $d_\phi$ from $X_p$ to $z$ and align to the residual:
\begin{equation}
\mathcal{L}_{\mathrm{pred}}
=\mathbb{E}\left\|z-d_\phi(X_p)\right\|^2,
\qquad
y^{\mathrm{pure}}=\sg\!\left[z-d_\phi(X_p)\right].
\label{eq:lap_target}
\end{equation}
Here $\sg$ denotes stop-gradient. Replacing $y$ with $y^{\mathrm{pure}}$ in \Cref{eq:align_loss} gives the LAP objective. The predictor selects what to remove and receives no gradients from diffusion training. The denoising objective, model architecture, and inference procedure remain unchanged (\Cref{fig:framework}). We consider two variants of $d_\phi$: an affine patch predictor and a nonlinear local predictor.

\textbf{LAP-L: linear purification.}
We fit an affine map from each clean-latent patch to its corresponding target token and subtract its prediction. This gives the residual used in \Cref{sec:dissociation}. The affine predictor is refitted independently on each minibatch, with no gradient through the fit (\Cref{app:controls}).

\textbf{LAP-N: nonlinear purification.}
We train a small convolutional predictor to recover the target from a local neighborhood of clean-latent patches. The predictor is fitted before diffusion training, frozen, and reused throughout training. Details are given in \Cref{app:purifier}.

What LAP removes depends on the predictor: local patches may still contain semantic information~\citep{brendel2019bagnet}, and a limited predictor may not capture all locally predictable content. We therefore do not interpret the residual as a purely semantic representation, but instead evaluate its content and effect on generation empirically in \Cref{sec:results}. Purification is used only to construct training targets and adds no inference cost.

\vspace{-5pt}
\subsection{A Variational View of Purified Alignment}
\label{sec:information}

The denoising objective trains the model to predict velocity, while alignment encourages the hidden state to preserve information about the teacher. We interpret this alignment objective probabilistically, which lets us connect cosine alignment to a variational information bound. For fixed \(l,t\), write \(h=h_t^l\) and \(a=g_\psi(h)\). For each nonzero token, let \(\hat y^{(i)}=y^{(i)}/\lVert y^{(i)}\rVert\) and define \(\hat a^{(i)}\) analogously. We interpret cosine alignment as predicting the direction of each teacher token from the hidden state. Specifically, we use a von Mises--Fisher decoder~\citep{banerjee2005clustering} whose likelihood increases with the cosine similarity between the normalized readout $\hat a^{(i)}$ and target $\hat y^{(i)}$:
\[
q_\psi(\hat y^{(i)}\mid h)\propto
\exp\!\left(\kappa\,\hat a^{(i)\top}\hat y^{(i)}\right),
\qquad \kappa>0.
\]
Under this decoder, maximizing the expected log-likelihood is equivalent, up to constants and scaling, to minimizing the cosine alignment loss. Thus, lower alignment loss corresponds to a better probabilistic prediction of the teacher representation from the hidden state.
For a fixed local predictor, write $S_i=X_{\mathcal N_k(i)}$ for the clean-latent neighborhood used to predict token $z^{(i)}$. The denoiser state $h$ still has access to the entire noisy latent. Applying the variational bound of \citet{barber2003im} gives
\begin{equation}
\begin{aligned}
\MI(h;z)
&\ge
c_\sigma-\kappa N\,\mathcal L_{\mathrm{align}}^{(t)}(z), \\
\sum_{i=1}^{N}\MI(h;z^{(i)}\mid S_i)
&\ge
c'_\sigma-\kappa N\,\mathcal L_{\mathrm{align}}^{(t)}(y^{\mathrm{pure}}).
\end{aligned}
\label{eq:info_bounds}
\end{equation}
Here $\MI$ denotes mutual information, with losses evaluated at fixed $t$. The first bound concerns the full teacher target. For LAP-N, the second aggregates information about each teacher token beyond its predictor's local input. The sum can count shared information more than once and is not a joint mutual information. Since $\kappa N>0$, minimizing either loss maximizes its lower bound.
The constants \(c_\sigma\) and \(c'_\sigma\) are independent of the denoiser and projector, and $\sigma>0$ is the smoothing scale used only for the analysis. For batchwise LAP-L, use $S_i=(X_p^{(i)},\widehat W)$, including the fitted affine coefficients (\Cref{app:variational}). That appendix gives the derivations and assumptions. These bounds may be loose and do not by themselves establish information separation or explain generation gains. We next introduce a directly measurable diagnostic of alignment targets.

\vspace{-5pt}
\subsection{The Constant-Readout Test}
\label{sec:baseline}

The variational view links alignment score to how well the teacher target can be predicted from the hidden state. However, a high score can also arise from structure shared across the targets themselves. If many target tokens point in a similar direction, a readout that ignores the hidden state and always outputs the same vector can still achieve a high cosine score. We therefore ask how well such a constant readout can align to the target.
For projector \(g_\psi\), define the alignment score
\begin{equation}
C_\psi(y)
=
\mathbb{E}\left[
\frac{1}{N}
\sum_{i=1}^{N}
\cos\left(g_\psi(h)^{(i)}, y^{(i)}\right)
\right].
\label{eq:alignment_score}
\end{equation}
Now consider a readout that outputs the same unit vector $\mu$ for every image and token. Let
\(
m_y
=
\mathbb{E}\left[
\frac{1}{N}
\sum_{i=1}^{N}
\hat y^{(i)}
\right].
\label{eq:target_mean_direction}
\)
The expected cosine score of this constant readout is $\mu^\top m_y$. Maximizing over all unit vectors gives the exact baseline
\begin{equation}
C_0(y)
=
\max_{\|\mu\|=1}
\mu^\top m_y
=
\|m_y\|.
\label{eq:constant_baseline}
\end{equation}
Thus, $C_0$ measures how much cosine alignment can be obtained from shared target structure alone, without using the hidden state. Since it depends only on the target distribution, $C_0$ can be computed before diffusion training by normalizing the target tokens, averaging them across images and positions, and taking the norm. After training, comparing $C_\psi$ with $C_0$ measures how much the learned readout gains beyond this constant baseline.
Under the same von Mises-Fisher decoder, the optimal gain over $C_0$ also admits an interpretation as predictive $\mathcal V$-information~\citep{xu2020vinfo} (derivations in \Cref{app:var_usable}). Motivated by this baseline, we empirically test whether raw-target $C_0$ predicts the generation benefit of LAP-L in \Cref{sec:target_diagnostics}.

\vspace{-5pt}
\section{Experiments}
\label{sec:results}

We now evaluate LAP across teacher encoders, alignment settings, training scales, and datasets, and ask when purification provides the largest generation gains. We also test whether the constant-readout score can identify promising targets before diffusion training.

\vspace{-5pt}
\subsection{Experimental Setup}
\label{sec:scales}

Unless stated otherwise, we train class-conditional SiT-B/2 on ImageNet-1k at $256\times256$ resolution for 400K steps with batch size 64, the SD-VAE, and representation alignment at block 8. We evaluate generation without guidance under the ADM protocol~\citep{dhariwal2021adm}, reporting FID-50K, Inception Score (IS), precision, and recall. Results are averaged over two training seeds. Evaluation details and individual runs are provided in \Cref{app:adm,app:seed_results}.
For development experiments, we use a 200-class subset (\emph{dev-200}) and report FID-DINOv2 on 10K samples~\citep{stein2023fddino}. These values are not directly comparable to FID-50K. Hidden-state diagnostics use disjoint probe-fitting and evaluation images, noise levels $t\in[0.3,0.7)$, and ablation rank $256$ (\Cref{app:probes}).

\begin{table}[t]
\centering\small
\setlength{\tabcolsep}{4.5pt}
\renewcommand{\arraystretch}{1.05}
\caption{\textbf{Purification across five teachers.} SiT-B/2 on ImageNet-256, mean $\pm$ sample SD over two training seeds. Teachers are ordered by raw-target $C_0$ on dev-200. Shaded rows use LAP variants. Bold marks the best mean per teacher and metric. Individual FIDs are in \Cref{tab:seeds_adm}.}
\vspace{3pt}
\label{tab:adm}
\begin{tabular}{lcl@{\hspace{8pt}}rrrr}
\toprule
Teacher & $C_0$ & Method & FID $\downarrow$ & IS $\uparrow$ & Prec. $\uparrow$ & Rec. $\uparrow$ \\
\midrule
\multicolumn{3}{l}{No alignment} & 53.92\sdev{0.03} & 25.34\sdev{0.16} & 0.435\sdev{0.002} & 0.602\sdev{0.006} \\
\midrule
\multirow{3}{*}{MAE ViT-L} & \multirow{3}{*}{0.93} & REPA & 53.05\sdev{0.90} & 26.30\sdev{0.45} & 0.438\sdev{0.004} & 0.599\sdev{0.001} \\
 &  & \ours $+$LAP-L & \ours 45.15\sdev{0.48} & \ours 31.44\sdev{0.25} & \ours 0.480\sdev{0.005} & \ours 0.617\sdev{0.003} \\
 &  & \ours $+$LAP-N & \ours \textbf{42.30}\sdev{1.08} & \ours \textbf{33.85}\sdev{0.44} & \ours \textbf{0.496}\sdev{0.006} & \ours \textbf{0.622}\sdev{0.010} \\
\midrule
\multirow{3}{*}{AIMv2 ViT-L} & \multirow{3}{*}{0.66} & REPA & 47.97\sdev{0.15} & 29.88\sdev{0.01} & 0.464\sdev{0.000} & 0.615\sdev{0.007} \\
 &  & \ours $+$LAP-L & \ours \textbf{44.31}\sdev{0.13} & \ours \textbf{32.67}\sdev{0.32} & \ours \textbf{0.484}\sdev{0.000} & \ours \textbf{0.622}\sdev{0.006} \\
 &  & \ours $+$LAP-N & \ours 44.56\sdev{0.82} & \ours 32.40\sdev{0.57} & \ours 0.483\sdev{0.004} & \ours 0.621\sdev{0.005} \\
\midrule
\multirow{3}{*}{CLIP ViT-L} & \multirow{3}{*}{0.61} & REPA & 44.74\sdev{0.01} & 32.21\sdev{0.13} & 0.481\sdev{0.003} & 0.615\sdev{0.007} \\
 &  & \ours $+$LAP-L & \ours 42.69\sdev{0.87} & \ours 33.91\sdev{0.65} & \ours 0.493\sdev{0.004} & \ours 0.621\sdev{0.004} \\
 &  & \ours $+$LAP-N & \ours \textbf{42.21}\sdev{0.16} & \ours \textbf{34.38}\sdev{0.23} & \ours \textbf{0.495}\sdev{0.001} & \ours \textbf{0.624}\sdev{0.009} \\
\midrule
\multirow{3}{*}{DINOv3 ViT-B} & \multirow{3}{*}{0.52} & REPA & 42.73\sdev{0.86} & 34.25\sdev{1.01} & 0.486\sdev{0.004} & 0.623\sdev{0.003} \\
 &  & \ours $+$LAP-L & \ours \textbf{42.07}\sdev{1.63} & \ours \textbf{34.91}\sdev{1.14} & \ours \textbf{0.490}\sdev{0.011} & \ours \textbf{0.633}\sdev{0.000} \\
 &  & \ours $+$LAP-N & \ours 43.27\sdev{0.24} & \ours 34.02\sdev{0.23} & \ours 0.485\sdev{0.002} & \ours 0.628\sdev{0.003} \\
\midrule
\multirow{3}{*}{DINOv2 ViT-B} & \multirow{3}{*}{0.22} & REPA & 41.75\sdev{0.56} & 35.50\sdev{0.62} & 0.492\sdev{0.001} & 0.630\sdev{0.005} \\
 &  & \ours $+$LAP-L & \ours \textbf{41.47}\sdev{1.21} & \ours \textbf{35.53}\sdev{1.07} & \ours \textbf{0.495}\sdev{0.010} & \ours 0.621\sdev{0.002} \\
 &  & \ours $+$LAP-N & \ours 42.11\sdev{0.30} & \ours 35.26\sdev{0.34} & \ours 0.493\sdev{0.002} & \ours \textbf{0.632}\sdev{0.002} \\
\bottomrule
\end{tabular}
\vspace{-10pt}
\end{table}

\vspace{-5pt}
\subsection{Generation Quality across Teachers}
\label{sec:generation}

Raw MAE barely improves over no alignment, whereas LAP-N reduces FID from $53.05$ to $42.30$, approaching DINOv2-based REPA at $41.75$ (\Cref{tab:adm}). AIMv2 and CLIP also improve in both seeds, with lower FID accompanied by higher IS, precision, and recall. Using LAP-L as a common comparison across all five teachers, mean FID improves by $7.9$, $3.7$, and $2.0$ for MAE, AIMv2, and CLIP, but by only $0.7$ and $0.3$ for DINOv3 and DINOv2, respectively. The latter differences are within seed variation. Purifier capacity also matters differently across teachers: for MAE, LAP-L reaches about $45.2$ FID, while LAP-N further improves to $42.3$. For CLIP, LAP-L captures most of the gain, whereas DINOv2 changes little across variants. Overall, purification can substantially improve otherwise weak alignment targets, but the magnitude of the benefit and the useful purifier capacity depend on the teacher.

\vspace{-5pt}
\subsection{LAP Complements Different REPA-style Recipes and Tokenizers}
\label{sec:composition}

Since LAP changes only the alignment target, we ask whether its gains persist when other parts of the alignment pipeline are improved. We combine LAP with spatial operations in iREPA~\citep{singh2025irepa}, the alignment schedule in VA-REPA~\citep{varepa2026}, the entangled class token in REG~\citep{reg2025}, and the structural loss in sREPA~\citep{structra2026}. We also evaluate LAP on different latent tokenizers such as SD-VAE~\citep{rombach2022ldm}, REPA-E VAE~\citep{leng2025repae}, and EQ-VAE~\citep{kouzelis2025eqvae}.
Our REG and sREPA implementations use the shared block-8, batch-64 recipe. Our REG implementation follows the official code, while sREPA is our reimplementation of the paper because no official code was available (\Cref{app:adm}).

\textbf{LAP adds to stronger alignment recipes.}
For MAE, LAP-N improves FID from $43.06$ to $41.32$ with spatial operations from iREPA and from $52.82$ to $41.60$ with the VA-REPA schedule (\Cref{tab:compose}). Spatial operations substantially improve MAE on their own, yet they do not subsume purification. REG and sREPA leave raw MAE relatively weak ($52.92$ and $50.43$ FID), whereas adding LAP-N lowers them to $41.65$ and $38.46$, respectively. For DINOv2, the selected LAP variants reduce mean FID across all four recipes by $0.3$--$1.2$, with lower FID in both seeds and higher mean IS and precision. Thus, purification remains useful across changes to the alignment loss, schedule, and architecture, although the gains are smaller for an already strong DINOv2 target.

\textbf{LAP makes MAE effective across tokenizers.}
Stronger tokenizers substantially improve the generative backbone, but do not solve the target-selection problem: raw MAE remains ineffective with SD-VAE, EQ-VAE, and REPA-E VAE (\Cref{tab:tokenizers}). LAP changes this consistently. It reduces FID from $53.05$ to $42.30$ with SD-VAE, from $52.41$ to $40.01$ with EQ-VAE, and from $33.74$ to $27.56$ with REPA-E VAE. Notably, with the stronger REPA-E tokenizer, raw MAE is worse than no alignment ($33.74$ vs.\ $32.30$), whereas LAP turns it into a substantial gain ($27.56$), while also improving IS from $39.32$ to $47.43$ and precision from $0.574$ to $0.606$. These results show that LAP remains valuable even after large improvements to the latent representation: the tokenizer improves the generative substrate, while purification makes MAE useful as an alignment target. The benefit also extends beyond MAE, with LAP improving DINOv2 from $26.10$ to $24.94$ FID on REPA-E VAE (\Cref{tab:tok_full}). All tokenizer comparisons use the released VAEs frozen, without joint VAE training.

\begin{table}[t]
\centering\small
\setlength{\tabcolsep}{4.5pt}
\renewcommand{\arraystretch}{1.05}
\vspace{-10pt}
\caption{\textbf{LAP on top of REPA-family recipes.} Same protocol as \Cref{tab:adm}. Each $+$LAP row uses the variant with the best two-seed mean FID. For REG/sREPA, these are LAP-N/LAP-N for MAE and LAP-L/LAP-N for DINOv2. Bold marks the better one. Full results are in \Cref{tab:adm_full}.}
\vspace{3pt}
\label{tab:compose}
\begin{tabular}{ll@{\hspace{8pt}}rrrr}
\toprule
Teacher & Method & FID $\downarrow$ & IS $\uparrow$ & Prec. $\uparrow$ & Rec. $\uparrow$ \\
\midrule
\multirow{8}{*}{MAE ViT-L} & iREPA & 43.06\sdev{0.62} & 33.23\sdev{0.66} & 0.491\sdev{0.004} & 0.617\sdev{0.006} \\
 & \ours $+$LAP & \ours \textbf{41.32}\sdev{0.28} & \ours \textbf{34.70}\sdev{0.37} & \ours \textbf{0.502}\sdev{0.002} & \ours \textbf{0.626}\sdev{0.001} \\
 & VA-REPA & 52.82\sdev{0.90} & 26.32\sdev{0.58} & 0.437\sdev{0.001} & 0.606\sdev{0.006} \\
 & \ours $+$LAP & \ours \textbf{41.60}\sdev{0.91} & \ours \textbf{34.06}\sdev{0.53} & \ours \textbf{0.503}\sdev{0.004} & \ours \textbf{0.622}\sdev{0.004} \\
 & REG & 52.92\sdev{0.53} & 26.33\sdev{0.41} & 0.434\sdev{0.002} & 0.608\sdev{0.002} \\
 & \ours $+$LAP & \ours \textbf{41.65}\sdev{0.06} & \ours \textbf{34.31}\sdev{0.16} & \ours \textbf{0.500}\sdev{0.003} & \ours \textbf{0.626}\sdev{0.002} \\
 & sREPA & 50.43\sdev{0.53} & 27.41\sdev{0.22} & 0.454\sdev{0.002} & 0.606\sdev{0.009} \\
 & \ours $+$LAP & \ours \textbf{38.46}\sdev{0.62} & \ours \textbf{36.89}\sdev{0.30} & \ours \textbf{0.519}\sdev{0.003} & \ours \textbf{0.622}\sdev{0.002} \\
\midrule
\multirow{8}{*}{DINOv2 ViT-B} & iREPA & 41.32\sdev{0.19} & 35.65\sdev{0.01} & 0.496\sdev{0.004} & \textbf{0.635}\sdev{0.003} \\
 & \ours $+$LAP & \ours \textbf{40.55}\sdev{0.39} & \ours \textbf{36.29}\sdev{0.16} & \ours \textbf{0.503}\sdev{0.003} & \ours 0.632\sdev{0.004} \\
 & VA-REPA & 40.71\sdev{0.14} & 36.27\sdev{0.24} & 0.500\sdev{0.002} & 0.635\sdev{0.001} \\
 & \ours $+$LAP & \ours \textbf{40.39}\sdev{0.02} & \ours \textbf{36.54}\sdev{0.04} & \ours \textbf{0.502}\sdev{0.001} & \ours \textbf{0.637}\sdev{0.006} \\
 & REG & 29.24\sdev{0.82} & 57.08\sdev{2.10} & 0.559\sdev{0.003} & 0.610\sdev{0.001} \\
 & \ours $+$LAP & \ours \textbf{28.67}\sdev{0.26} & \ours \textbf{57.49}\sdev{0.07} & \ours \textbf{0.567}\sdev{0.005} & \ours \textbf{0.615}\sdev{0.004} \\
 & sREPA & 38.13\sdev{0.01} & 39.55\sdev{0.01} & 0.515\sdev{0.005} & \textbf{0.635}\sdev{0.004} \\
 & \ours $+$LAP & \ours \textbf{36.92}\sdev{0.42} & \ours \textbf{41.11}\sdev{0.46} & \ours \textbf{0.522}\sdev{0.002} & \ours 0.631\sdev{0.005} \\
\bottomrule
\end{tabular}
\vspace{-10pt}
\end{table}

\begin{table}[t]
\centering\small
\setlength{\tabcolsep}{4.5pt}
\renewcommand{\arraystretch}{1.05}
\caption{\textbf{LAP with other tokenizers.} MAE ViT-L teacher, same protocol as \Cref{tab:adm}. Each $+$LAP row uses LAP-N. Bold marks the best.}
\vspace{3pt}
\label{tab:tokenizers}
\begin{tabular}{ll@{\hspace{8pt}}rrrr}
\toprule
Tokenizer & Method & FID $\downarrow$ & IS $\uparrow$ & Prec. $\uparrow$ & Rec. $\uparrow$ \\
\midrule
\multirow{3}{*}{SD-VAE} & No alignment & 53.92\sdev{0.03} & 25.34\sdev{0.16} & 0.435\sdev{0.002} & 0.602\sdev{0.006} \\
 & REPA & 53.05\sdev{0.90} & 26.30\sdev{0.45} & 0.438\sdev{0.004} & 0.599\sdev{0.001} \\
 & \ours $+$LAP & \ours \textbf{42.30}\sdev{1.08} & \ours \textbf{33.85}\sdev{0.44} & \ours \textbf{0.496}\sdev{0.006} & \ours \textbf{0.622}\sdev{0.010} \\
\midrule
\multirow{3}{*}{EQ-VAE} & No alignment & 52.50\sdev{0.04} & 25.99\sdev{0.05} & 0.449\sdev{0.003} & 0.599\sdev{0.004} \\
 & REPA & 52.41\sdev{1.37} & 26.31\sdev{0.91} & 0.452\sdev{0.004} & 0.597\sdev{0.004} \\
 & \ours $+$LAP & \ours \textbf{40.01}\sdev{0.25} & \ours \textbf{35.24}\sdev{0.36} & \ours \textbf{0.523}\sdev{0.003} & \ours \textbf{0.608}\sdev{0.005} \\
\midrule
\multirow{3}{*}{REPA-E VAE} & No alignment & 32.30\sdev{0.04} & 40.61\sdev{0.09} & 0.584\sdev{0.002} & 0.604\sdev{0.005} \\
 & REPA & 33.74\sdev{0.45} & 39.32\sdev{0.88} & 0.574\sdev{0.005} & \textbf{0.611}\sdev{0.003} \\
 & \ours $+$LAP & \ours \textbf{27.56}\sdev{0.40} & \ours \textbf{47.43}\sdev{0.44} & \ours \textbf{0.606}\sdev{0.003} & \ours 0.607\sdev{0.000} \\
\bottomrule
\end{tabular}
\vspace{-10pt}
\end{table}

\begin{figure}[t]
\centering
\includegraphics[width=1\linewidth]{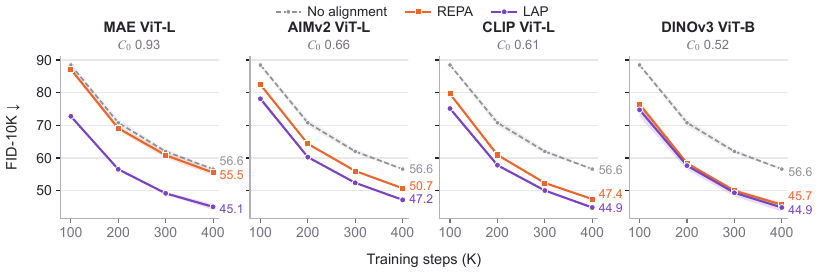}
\vspace{-25pt}
\caption{\textbf{Purification improves throughout training.} Two-seed FID-10K means and ranges for SiT-B/2, using LAP-N for MAE/CLIP and LAP-L for AIMv2/DINOv3. Full trajectories and evaluation details: \Cref{app:trajectories}.}
\label{fig:training_rescue}
\vspace{-6pt}
\end{figure}

\vspace{-5pt}
\subsection{LAP Gains Persist Across Training, Scale, and Datasets}
\label{sec:scaling}

\textbf{LAP improves generation early and throughout training.}
By 100K steps, LAP lowers FID-10K by $14.4$ for MAE, $4.3$ for AIMv2, and $4.5$ for CLIP (\Cref{fig:training_rescue}). These gains persist through 400K steps: MAE REPA remains close to the unaligned model, whereas MAE LAP reaches the same final FID as DINOv2 REPA (\Cref{tab:curves_fid10k}). Purification therefore improves the full training trajectory.

\begin{table}[t]
\vspace{-10pt}
\centering\small
\setlength{\tabcolsep}{4.5pt}
\renewcommand{\arraystretch}{1.05}
\caption{\textbf{LAP on aerial-scene datasets.} Mean $\pm$ sample SD over four training seeds for RESISC45 and two for AID. FID uses 10K samples and 10K reference images. LAP uses LAP-N. Bold marks the best. Details: \Cref{app:domains}.}
\vspace{3pt}
\label{tab:domains_main}
\begin{tabular}{llrrrr}
\toprule
Dataset & Method & FID $\downarrow$ & IS $\uparrow$ & Prec. $\uparrow$ & Rec. $\uparrow$ \\
\midrule
\multirow{4}{*}{RESISC45} & No alignment & 68.06\sdev{5.10} & 4.75\sdev{0.17} & 0.421\sdev{0.014} & 0.200\sdev{0.020} \\
 & REPA (DINOv2) & 66.34\sdev{3.70} & 4.80\sdev{0.15} & 0.426\sdev{0.008} & 0.184\sdev{0.016} \\
 & REPA (MAE) & 64.65\sdev{4.38} & 4.94\sdev{0.23} & 0.414\sdev{0.013} & 0.205\sdev{0.024} \\
 & \ours LAP (MAE) & \ours \textbf{61.47}\sdev{2.33} & \ours \textbf{4.96}\sdev{0.04} & \ours \textbf{0.433}\sdev{0.010} & \ours \textbf{0.228}\sdev{0.009} \\
\midrule
\multirow{4}{*}{AID} & No alignment & 62.41\sdev{2.06} & 6.47\sdev{0.02} & 0.365\sdev{0.003} & \textbf{0.222}\sdev{0.003} \\
 & REPA (DINOv2) & 63.68\sdev{1.35} & 6.36\sdev{0.02} & 0.376\sdev{0.002} & 0.200\sdev{0.003} \\
 & REPA (MAE) & 66.13\sdev{1.47} & 6.40\sdev{0.06} & 0.368\sdev{0.007} & 0.191\sdev{0.003} \\
 & \ours LAP (MAE) & \ours \textbf{60.13}\sdev{3.29} & \ours \textbf{6.52}\sdev{0.09} & \ours \textbf{0.382}\sdev{0.001} & \ours 0.214\sdev{0.026} \\
\bottomrule
\end{tabular}
\vspace{-14pt}
\end{table}

\textbf{LAP remains effective across training settings.}
With MAE, LAP continues to improve generation quality at larger batch size, with SiT-L/2 and SiT-XL/2, and at a different alignment depth (details in \Cref{app:regime}). At block 4, the depth \citet{singh2025irepa} recommend for SiT-B/2, LAP also improves CLIP and AIMv2 by $3.7$ and $4.6$ FID, more than at block 8 (\Cref{app:depth}). The gains also persist under classifier-free guidance, where MAE, AIMv2, and CLIP improve in both seeds (\Cref{tab:cfg}). Thus, the benefit of purification is not specific to the base training configuration.

\begin{wrapfigure}{R}{0.45\textwidth}
\vspace{-20pt}
\centering
\setlength{\abovecaptionskip}{5pt}
\includegraphics[width=\linewidth]{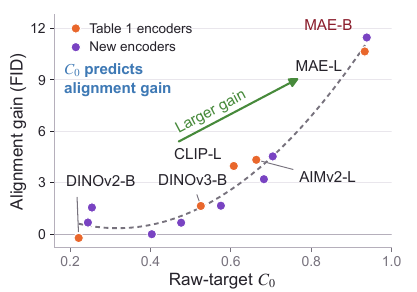}
\vspace{-18pt}
\caption{\textbf{$C_0$ predicts purification gain.} Raw-to-LAP-L FID-10K improvement at 100K steps. Dashed line: quadratic fit.}
\label{fig:sweep}
\vspace{-30pt}
\end{wrapfigure}

\textbf{LAP also transfers beyond ImageNet.}
LAP with MAE achieves the best FID on two aerial-scene datasets: RESISC45 and AID (\Cref{tab:domains_main}). It improves over no alignment by $6.6$ FID on RESISC45 and $2.3$ on AID, and outperforms DINOv2-based REPA by $4.9$ and $3.6$ FID, respectively. Notably, on AID, neither raw MAE nor DINOv2 improves over no alignment, whereas purified MAE does. These results suggest that purification can recover useful alignment guidance even when the pretrained teacher is transferred outside its original image domain.

\vspace{-5pt}
\subsection{The Constant-Readout Score Predicts Purification Benefit}
\label{sec:target_diagnostics}

\textbf{$C_0$ predicts which teachers benefit from LAP-L.} Across 13 encoders evaluated under matched protocols, higher raw-target $C_0$ is associated with larger FID gains from LAP-L (\Cref{fig:sweep,fig:sweep_400k}). The Spearman correlation is $\rho=0.94$ at 100K steps (FID-10K, one-sided permutation $p\approx3.0\times10^{-6}$) and $0.79$ at 400K steps (FID-50K, $p=0.0010$). After controlling for raw-target FID, the corresponding partial rank correlations remain $0.83$ and $0.69$, indicating that raw-target generation quality alone does not explain the association. MAE illustrates why this diagnostic is useful: its raw-target $C_0=0.93$ is close to the trained projector's cosine score of $0.957$, so much of its alignment score is attainable without using the hidden state. These results support $C_0$ as a cheap diagnostic for linear-purification benefit under this recipe. The separate pre-specified eight-encoder test shows the same direction (\Cref{app:sweep}).

\textbf{$C_0$ does not explain the full benefit of LAP.} For MAE, LAP-L and LAP-N both reduce $C_0$ to about $0.01$--$0.02$, yet LAP-N achieves better generation quality. This suggests that suppressing the shared direction measured by $C_0$ is not sufficient to explain LAP-N's additional gain. Instead, LAP-N removes substantially more target variation: it eliminates 35\% of held-out target variance while preserving class information. On the 200-class pooled linear probe, the LAP-N residual reaches 68.7\% accuracy, compared with 66.1\% for raw MAE and 68.5\% for the LAP-L residual (\Cref{app:purifier}). Together, these results suggest that LAP-N removes additional image-dependent local content beyond the shared component captured by $C_0$.

\vspace{-5pt}
\section{Conclusion}
\label{sec:conclusion}

Prediction utility and alignment benefit need not coincide. Our fixed-encoder interventions motivate LAP, which removes locally predictable content before alignment and makes otherwise weak targets such as MAE effective. LAP complements existing REPA recipes and latent tokenizers without changing inference. Variational information bounds clarify what the alignment objective rewards, while the constant-readout score $C_0$ measures alignment attainable without consulting the hidden state and empirically identifies teachers likely to benefit from LAP-L under our shared recipe.

\textbf{Limitations.} Evaluation centers on SiT and ImageNet. Larger-model checks use two seeds and cross-domain runs are short. Gains depend on the teacher and recipe. LAP does not guarantee a semantic residual, and $C_0$ does not provide calibrated gain predictions.

\subsection*{Reproducibility Statement}
All experiments use the public REPA codebase as the base recipe with the modifications described in \Cref{sec:setup,sec:mechanism,sec:method,sec:results}. Hyperparameters, probe and ablation protocols, seed conventions, purifier training, control specifications, and the pre-specified sweep design are given in \Cref{app:theory,app:variational,app:probes,app:controls,app:purifier,app:regime,app:arc,app:adm,app:domains,app:seed_results}, with per-run compute disclosed in \Cref{app:compute}. Code and checkpoints are on the project page.

\subsection*{AI Use Statement}
In this work, we used generative AI tools, specifically large language models (LLMs), to assist in the derivation of the mathematical formulation. Generative AI was used in developing code to run experiments and in producing the tables and figures that report our results. We also used LLMs to summarize existing literature and format references. We further used LLMs to provide feedback on research methodology and implementation. We take responsibility for the final content of this work, including text, claims or artifacts produced with the aid of generative AI.

\bibliography{references}
\bibliographystyle{iclr2027_conference}
\clearpage
\appendix
\crefalias{section}{appendix}
\crefalias{subsection}{appendix}
\etocdepthtag.toc{appendix}
\begingroup
\renewcommand{\contentsname}{Appendix Contents}
\etocsettagdepth{main}{none}
\etocsettagdepth{appendix}{subsection}
\tableofcontents
\endgroup
\medskip
\noindent
For the theoretical distinction between decodability and use, start with \Cref{app:theory}.
\Cref{app:variational} reads the alignment losses as variational information bounds.
For experimental definitions and reliability checks, see \Cref{app:probes,app:controls}.
Purifier validation and robustness results are in \Cref{app:purifier,app:regime}.
\Cref{app:seed_results} collects the individual runs behind the main comparisons.
\clearpage
\begingroup
\raggedbottom
\setlength{\textfloatsep}{12pt}
\setlength{\intextsep}{10pt}
\renewcommand{\arraystretch}{1.12}
\input{appendix}
\clearpage
\endgroup
\end{document}

%% file: math_commands.tex
\usepackage{amsmath,amsfonts,bm}

\def\1{\bm{1}}

\DeclareMathAlphabet{\mathsfit}{\encodingdefault}{\sfdefault}{m}{sl}
\SetMathAlphabet{\mathsfit}{bold}{\encodingdefault}{\sfdefault}{bx}{n}

%% file: appendix.tex
\section{Linear-Gaussian Analysis of Decodability and Use}
\label{app:theory}

This analysis asks a limited question: can a feature be easy to decode without helping predict a second target? The answer is yes, even with optimal predictors. The model below explains that distinction. It does not model diffusion training or prove that an auxiliary loss harms optimization.

\subsection{Model and Prediction Error}

Let $\xi\in\mathbb{R}^k$ be a latent factor vector. Define
\begin{equation}
\xi\sim\mathcal{N}(0,I_k),\qquad
h=\xi+\eta,\qquad
y=C\xi,\qquad v=B\xi,
\end{equation}
where $\eta\sim\mathcal{N}(0,\sigma^2 I_k)$ is independent of $\xi$, $\sigma^2\geq0$, and $B$ and $C$ are fixed matrices. Here $y$ represents an alignment target and $v$ a prediction target. The hidden state $h$ is fixed by this construction, rather than learned.

For an orthogonal projector $P=P^\top=P^2$, let $\mathcal{R}(P)$ be the minimum mean-squared error for predicting $v$ from $Ph$. Joint Gaussianity makes the conditional mean linear. Writing $s=1+\sigma^2$, we obtain
\begin{align}
\mathbb{E}[v\mid Ph]&=\frac{1}{s}BPh,\\
\mathcal{R}(P)&=\operatorname{tr}(BB^\top)
-\frac{1}{s}\operatorname{tr}(BPB^\top).
\label{eq:gaussian_risk}
\end{align}
To see this, $\operatorname{Cov}(v,Ph)=BP$ and $\operatorname{Cov}(Ph)=sP$. The pseudoinverse of $sP$ is $P/s$. Substituting these quantities into the Gaussian conditional covariance gives $BB^\top-BPB^\top/s$. Its trace is the scalar prediction error.

\subsection{Decodability Can Be High When Use Is Zero}

Assume $C\neq0$, and let $P_y$ project onto the row space of $C$. We define presence using a variance-weighted population coefficient of determination:
\begin{equation}
R^2(h\to y)=1-
\frac{\mathbb{E}\|y-\mathbb{E}[y\mid h]\|^2}{\operatorname{tr}(CC^\top)}
=\frac{1}{1+\sigma^2}.
\label{eq:gaussian_presence}
\end{equation}
This score depends on observation noise but not on $B$.

\textbf{Proposition 1.} Define information-level use as the extra optimal prediction error after removing a subspace:
\begin{equation}
U_{\mathrm{opt}}(P)=\mathcal{R}(I-P)-\mathcal{R}(I)
=\frac{\|BP\|_F^2}{1+\sigma^2}.
\label{eq:gaussian_use}
\end{equation}
For $B\neq0$, the normalized use of the target subspace is therefore
\begin{equation}
\frac{U_{\mathrm{opt}}(P_y)}{U_{\mathrm{opt}}(I)}
=\frac{\|BP_y\|_F^2}{\|B\|_F^2}.
\end{equation}
If $\operatorname{rank}(C)<k$, a nonzero $B$ can be chosen with rows in the orthogonal complement of $\operatorname{row}(C)$. Then $BP_y=0$, so target-subspace use is zero while presence approaches one as $\sigma^2\to0$. If $C$ has full column rank, $P_y=I$ and its normalized use is instead one.

\emph{Proof.} Apply \Cref{eq:gaussian_risk} to $I-P$ and $I$, then subtract. Since $P$ is symmetric and idempotent, $\operatorname{tr}(BPB^\top)=\|BP\|_F^2$. \Cref{eq:gaussian_presence} follows by replacing $B$ with $C$ and retaining the full hidden state. The zero-use construction requires a nontrivial orthogonal complement, which is exactly the stated rank condition. $\square$

\subsection{Prediction Through a Restricted Subspace}

\textbf{Proposition 2.} For a unit vector $u$, predicting $v$ only from $u^\top h$ reduces error relative to predicting the mean by
\begin{equation}
\operatorname{tr}(BB^\top)-\mathcal{R}(uu^\top)
=\frac{\|Bu\|^2}{1+\sigma^2}.
\end{equation}
Thus, a direction helps prediction exactly when $Bu\neq0$. A subspace $P$ retains all of the full-state prediction benefit exactly when $B(I-P)=0$.

\emph{Proof.} The first statement is \Cref{eq:gaussian_risk} with $P=uu^\top$. For the second, the error gap from restricting prediction to $P$ is $\|B(I-P)\|_F^2/(1+\sigma^2)$. This vanishes precisely under the stated condition. $\square$

The relevant notion is non-orthogonal overlap, not a literal intersection of row spaces. For example, two distinct one-dimensional subspaces can have zero intersection but still have a nonzero projection onto one another. The formula above captures that case.

\subsection{Scope and Numerical Verification}

The experimental use score holds the trained network fixed and measures its response to an intervention. In contrast, $U_{\mathrm{opt}}$ compares optimal predictors under an isotropic Gaussian model. In this special model, evaluating the full-state optimal predictor on $(I-P)h$ also attains the retained-subspace optimum. That equivalence need not hold for a nonlinear denoiser. Experimental use scores need not be nonnegative or bounded by the full-layer effect in every possible setting.

The model has no learning dynamics. Adding an auxiliary task cannot change $h$, and an unrestricted optimal velocity predictor already uses all available information. Consequently, these propositions do not establish gradient competition, semantic content, or the safety of residual alignment during training. Those interpretations must be assessed using the experiments.

The accompanying verification script checks the covariance identities, the rank condition, and orthogonal, shared, and oblique subspace examples. It also compares the analytical errors with held-out Monte Carlo predictions using $4\times10^5$ samples. These checks verify the implementation of the algebra, rather than adding empirical evidence about diffusion models.

\FloatBarrier
\section{Alignment Targets as Variational Information Bounds}
\label{app:variational}

This appendix derives the information bounds and constant-readout identity of \Cref{sec:information,sec:baseline}. We first construct a directional decoder whose negative log-likelihood is the cosine alignment loss, up to a positive scale and a constant (\Cref{app:var_decoder}). We then derive the raw and purified bounds, including the smoothing correction needed for the actual training targets (\Cref{app:var_bounds,app:var_targets}). In both cases, minimizing the loss maximizes the lower bound. Finally, we solve for the best constant readout and derive the exact predictive-score identity (\Cref{app:var_usable}). The distinction between optimizing a bound and increasing information, and the connection to InfoDiffusion, are discussed in \Cref{app:var_scope}.

\subsection{Setup and Notation}
\label{app:var_setup}

\textbf{Mutual information.} The main text writes $I$ for the image, so we write $\MI(\cdot\,;\cdot)$ for mutual information. All random variables take values in Euclidean spaces or in products of spheres, so regular conditional distributions exist. For random variables $U$, $W$, and $S$, with joint law $P_{UW}$ and marginals $P_U$, $P_W$,
\begin{align}
\MI(U;W)&=\mathrm{KL}\!\left(P_{UW}\,\middle\|\,P_U\otimes P_W\right)\in[0,\infty],
\label{eq:mi_def}\\
\MI(U;W\mid S)&=\int \mathrm{KL}\!\left(P_{UW\mid S=s}\,\middle\|\,P_{U\mid S=s}\otimes P_{W\mid S=s}\right)P_S(ds).
\label{eq:cmi_def}
\end{align}
We use two standard facts, which hold in this generality~\citep{cover2006elements,gray2011entropy}. The first is the chain rule,
\begin{equation}
\MI\big(U;(W,S)\big)=\MI(U;S)+\MI(U;W\mid S).
\label{eq:chain_rule}
\end{equation}
The second is the data-processing inequality: for probability measures $P$, $R$ and a measurable map $T$,
\begin{equation}
\mathrm{KL}\!\left(P\circ T^{-1}\,\middle\|\,R\circ T^{-1}\right)\le\mathrm{KL}(P\,\|\,R),
\label{eq:dpi}
\end{equation}
with equality when $T$ is a bijection onto its image with measurable inverse.

\textbf{Entropy relative to a reference measure.} Let $\nu$ be a $\sigma$-finite measure on the space of $W$. If $P_W$ has density $p$ with respect to $\nu$, we write $H_\nu(W)=-\mathbb{E}\log p(W)$. If, for $P_S$-almost every $s$, the conditional law $P_{W\mid S=s}$ has density $p(\cdot\mid s)$ with respect to $\nu$, we write $H_\nu(W\mid S)=-\mathbb{E}\log p(W\mid S)$. Both are used only when the expectation is finite.

\textbf{Alignment variables.} Fix a noise time $t\in(0,1]$. The hidden state $h=h_t^l$ is a deterministic function of $x_t$ and the conditioning vector $e$, with parameters $\theta$. Here $e$ is the sum of the time embedding and the class embedding after label dropout. The aligned representation is $a=g_\psi(h)\in\mathbb{R}^{N\times D}$ with tokens $a^{(i)}$, which we assume nonzero. The encoder features are $z=f(I)\in\mathbb{R}^{N\times D}$. Since $f$ is frozen, the law of $z$ does not depend on $(\theta,\psi)$. For any $y\in\mathbb{R}^{N\times D}$ with nonzero tokens, let $\hat y^{(i)}=y^{(i)}/\|y^{(i)}\|$ and $\hat y=(\hat y^{(1)},\dots,\hat y^{(N)})\in(\mathbb{S}^{D-1})^N$. Let $\omega$ be the surface measure on the unit sphere $\mathbb{S}^{D-1}$ and $\nu=\omega^{\otimes N}$ the product measure on $(\mathbb{S}^{D-1})^N$. We assume $D\ge2$.

\textbf{Smoothed targets.} Let $\xi\in\mathbb{R}^{N\times D}$ have independent standard normal entries and be independent of all other variables. For $\sigma>0$ and any array $y$, we write $y_\sigma=y+\sigma\xi$. The tokens of $y_\sigma$ are nonzero almost surely, so $\hat y_\sigma$ is well defined. Smoothing is a device of the analysis. Training uses the unsmoothed targets.

\textbf{Local side information.} For token $i$, let $\mathcal N_k(i)$ be the predictor's neighborhood on the clean-latent token grid, including its fixed boundary-padding convention, and let $S_i=X_{\mathcal N_k(i)}$ collect only those patches. A fixed local predictor satisfies $d_\phi(X_p)^{(i)}=d_{\phi,i}(S_i)$; the index allows for boundary effects. In contrast, $h$ denotes the full hidden-state array, whose tokens can incorporate information from the whole noisy image. We study the sum of token-wise conditional information terms
\begin{equation}
\MI_{\mathrm{loc}}(h;z)=\sum_{i=1}^{N}\MI\big(h;z^{(i)}\mid S_i\big).
\label{eq:local_information}
\end{equation}
Each term conditions on that token's local neighborhood, not on the full clean latent $X_p$. This sum can count shared information more than once; it is not joint mutual information about the target array. Its definition depends on the chosen neighborhoods.

\subsection{From the vMF Decoder to the Alignment Loss}
\label{app:var_decoder}

\textbf{A likelihood for directions.} Cosine alignment compares directions and ignores token norms. A natural probabilistic decoder therefore assigns a density on the unit sphere. For a mean direction $\mu\in\mathbb{S}^{D-1}$ and concentration $\kappa>0$, the von Mises--Fisher (vMF) density with respect to $\omega$ is~\citep{mardia2000directional}
\begin{equation}
\mathrm{vMF}(w;\mu,\kappa)=c_D(\kappa)\exp\!\left(\kappa\,\mu^\top w\right),
\qquad
c_D(\kappa)=\frac{\kappa^{D/2-1}}{(2\pi)^{D/2}\,\mathcal{B}_{D/2-1}(\kappa)},
\end{equation}
where $\mathcal{B}_\alpha$ is the modified Bessel function of the first kind. The normalizer satisfies $c_D(\kappa)^{-1}=\int\exp(\kappa\mu^\top w)\,\omega(dw)$ and is independent of $\mu$ by rotation invariance. We keep $\kappa$ fixed throughout the analysis. The denoiser and projector choose only the mean direction of each token. This gives the product decoder
\begin{equation}
q_\psi(w\mid h)=\prod_{i=1}^{N}\mathrm{vMF}\!\left(w^{(i)};\,\frac{a^{(i)}}{\|a^{(i)}\|},\,\kappa\right),
\qquad a=g_\psi(h),
\label{eq:vmf_decoder}
\end{equation}
which is a probability density with respect to $\nu$ for every $h$. Factorizing the decoder is a modeling choice, not an assumption that the true target tokens are independent.

\textbf{Recovering the cosine objective.} Taking the logarithm turns the product over tokens into a sum. Substituting the normalized readout and target gives
\begin{align}
\log q_\psi(\hat y\mid h)
&=\sum_{i=1}^{N}\left[\log c_D(\kappa)+\kappa\frac{(a^{(i)})^\top y^{(i)}}{\|a^{(i)}\|\,\|y^{(i)}\|}\right] \notag\\
&=N\log c_D(\kappa)+\kappa\sum_{i=1}^{N}\cos\!\left(a^{(i)},y^{(i)}\right).
\label{eq:vmf_loglik}
\end{align}
At a fixed noise time $t$, write $\mathcal{L}_{\mathrm{align}}^{(t)}(y)=-\mathbb{E}[\tfrac1N\sum_i\cos(a^{(i)},y^{(i)})]$ for the alignment loss of \Cref{eq:align_loss} with target $y$, evaluated at noise time $t$. Taking the negative expectation in \Cref{eq:vmf_loglik} yields
\begin{equation}
\mathbb{E}\big[-\log q_\psi(\hat y\mid h)\big]
=-N\log c_D(\kappa)+\kappa N\,\mathcal{L}_{\mathrm{align}}^{(t)}(y).
\label{eq:vmf_nll}
\end{equation}
Thus minimizing the cosine loss is exactly equivalent to maximizing the decoder's expected log-likelihood for fixed $\kappa$. This identity requires no density for the target distribution. Also, $|\log q_\psi(\hat y\mid h)|\le N|\log c_D(\kappa)|+\kappa N$, so its expectations are finite.

\textbf{From likelihood to an information bound.} Relating this score to Shannon information requires an additional step. For a smoothed target $W=\hat y_\sigma$ with the finite marginal and conditional entropies established below (Lemma~4 with $S$ constant and with $S=h$ supplies the densities, and Lemma~1 gives the gap), adding and subtracting $\mathbb{E}\log q_\psi(W\mid h)$ gives the Barber--Agakov decomposition~\citep{barber2003im},
\begin{align}
\MI(h;W)
&=H_\nu(W)+\mathbb{E}\log q_\psi(W\mid h)
  +\mathbb{E}_{h}\,\mathrm{KL}\!\left(P_{W\mid h}\,\middle\|\,q_\psi(\cdot\mid h)\nu\right) \notag\\
&\ge H_\nu(W)+N\log c_D(\kappa)
  -\kappa N\,\mathcal{L}_{\mathrm{align}}^{(t)}(y_\sigma).
\label{eq:vmf_ba_walkthrough}
\end{align}
The inequality drops a nonnegative decoder-approximation gap. The target entropy is fixed because the encoder and target construction do not change during diffusion training. Consequently, decreasing the alignment loss raises the lower bound. Training uses $y$, not $y_\sigma$, so \Cref{app:var_targets} bounds the change in cosine caused by smoothing and absorbs it into a model-independent constant. Applying the conditional argument separately to each target token given $S_i$, then summing, gives the purified-target bound. The lemmas below justify these steps even when the original targets have no density.

\subsection{General Lemmas}
\label{app:var_bounds}

The first lemma is the bound of \citet{barber2003im}. We prove it without assuming that $W$ has a conditional density given $U$, because $h$ is a deterministic function of the noisy input and such a density need not exist.

\textbf{Lemma 1.} Let $P_W$ have density $p$ with respect to $\nu$, with $\mathbb{E}|\log p(W)|<\infty$. Let $q(w\mid u)\ge0$ be jointly measurable with $\int q(w\mid u)\,\nu(dw)=1$ for every $u$, and $\mathbb{E}|\log q(W\mid U)|<\infty$. Then
\begin{equation}
\MI(U;W)\;\ge\;H_\nu(W)+\mathbb{E}\log q(W\mid U).
\label{eq:ba_bound}
\end{equation}
If, in addition, $P_{W\mid U=u}$ has a density $p(\cdot\mid u)$ with respect to $\nu$ for $P_U$-almost every $u$ and $\MI(U;W)<\infty$, the gap in \Cref{eq:ba_bound} equals $\mathbb{E}_{U}\,\mathrm{KL}\big(p(\cdot\mid U)\,\|\,q(\cdot\mid U)\big)$.

\emph{Proof.} Write $P=P_{UW}$ and $R=P_U\otimes P_W$. If $P$ is not absolutely continuous with respect to $R$, then $\MI(U;W)=\infty$ by \Cref{eq:mi_def} and the claim holds, since the right-hand side of \Cref{eq:ba_bound} is finite. Otherwise let $\rho=dP/dR$, so that $\MI(U;W)=\mathbb{E}_P\log\rho$. If this is infinite, the claim again holds. Assume it is finite.

Define $g(u,w)=q(w\mid u)/p(w)$ on $\{p(w)>0\}$. Since $p(W)>0$ almost surely, $g$ is defined $P$-almost everywhere and $R$-almost everywhere. The integrability assumptions give $\mathbb{E}_P|\log g|<\infty$ and
\begin{equation}
\mathbb{E}_P\log g=\mathbb{E}\log q(W\mid U)-\mathbb{E}\log p(W)=\mathbb{E}\log q(W\mid U)+H_\nu(W).
\label{eq:ba_split}
\end{equation}
Since $\rho>0$ holds $P$-almost surely and both $\mathbb{E}_P\log\rho$ and $\mathbb{E}_P\log g$ are finite,
\begin{equation}
\MI(U;W)-\mathbb{E}_P\log g=-\mathbb{E}_P\log\frac{g}{\rho}\;\ge\;-\log\mathbb{E}_P\!\left[\frac{g}{\rho}\right],
\end{equation}
by Jensen's inequality for the concave logarithm. Changing measure from $P$ to $R$,
\begin{align}
\mathbb{E}_P\!\left[\frac{g}{\rho}\right]
&=\int_{\{\rho>0\}}\frac{g}{\rho}\,\rho\,dR
=\int_{\{\rho>0\}}g\,dR
\;\le\;\int g\,dR \notag\\
&=\int P_U(du)\int_{\{p>0\}}\frac{q(w\mid u)}{p(w)}\,p(w)\,\nu(dw)
\;\le\;\int P_U(du)\int q(w\mid u)\,\nu(dw)=1.
\end{align}
The first inequality uses $g\ge0$, the next equality is Tonelli's theorem, and the last inequality enlarges the domain of integration. Therefore $\MI(U;W)-\mathbb{E}_P\log g\ge-\log1=0$, and \Cref{eq:ba_bound} follows from \Cref{eq:ba_split}.

For the second statement, the additional assumption gives $\rho(u,w)=p(w\mid u)/p(w)$, so $\MI(U;W)=\mathbb{E}\log p(W\mid U)+H_\nu(W)$. Subtracting the right-hand side of \Cref{eq:ba_bound} leaves $\mathbb{E}\big[\log p(W\mid U)-\log q(W\mid U)\big]=\mathbb{E}_U\,\mathrm{KL}\big(p(\cdot\mid U)\,\|\,q(\cdot\mid U)\big)$. $\square$

\textbf{Lemma 2.} Suppose that, for $P_S$-almost every $s$, the conditional law $P_{W\mid S=s}$ has density $p(\cdot\mid s)$ with respect to $\nu$, with $\mathbb{E}|\log p(W\mid S)|<\infty$. Let $q(w\mid u,s)\ge0$ be jointly measurable with $\int q(w\mid u,s)\,\nu(dw)=1$ for every $(u,s)$, and $\mathbb{E}|\log q(W\mid U,S)|<\infty$. Then
\begin{equation}
\MI(U;W\mid S)\;\ge\;H_\nu(W\mid S)+\mathbb{E}\log q(W\mid U,S).
\label{eq:cba_bound}
\end{equation}

\emph{Proof.} The integrability assumptions imply that, for $P_S$-almost every $s$, both $\mathbb{E}\big[|\log p(W\mid s)|\,\big|\,S=s\big]$ and $\mathbb{E}\big[|\log q(W\mid U,s)|\,\big|\,S=s\big]$ are finite. For each such $s$, apply Lemma~1 to the pair $(U,W)$ under the conditional law $P_{UW\mid S=s}$, with density $p(\cdot\mid s)$ and decoder $q(\cdot\mid\cdot,s)$. This gives
\begin{equation}
\mathrm{KL}\!\left(P_{UW\mid S=s}\,\middle\|\,P_{U\mid S=s}\otimes P_{W\mid S=s}\right)
\ge-\mathbb{E}\big[\log p(W\mid s)\mid S=s\big]+\mathbb{E}\big[\log q(W\mid U,s)\mid S=s\big].
\end{equation}
Integrating both sides against $P_S$ and using \Cref{eq:cmi_def} gives \Cref{eq:cba_bound}. $\square$

\textbf{Lemma 3.} Let $U$, $W$, $S$ be random variables.
\begin{enumerate}
\item[(a)] Let $(s,w)\mapsto\varphi_s(w)$ be jointly measurable. Then $\MI\big(U;\varphi_S(W)\mid S\big)\le\MI(U;W\mid S)$.
\item[(b)] Let $m$ be measurable. Then $\MI\big((U,m(S));W\mid S\big)=\MI(U;W\mid S)$.
\item[(c)] Let $\xi$ be independent of $(U,W,S)$. Then $\MI\big(U;(W,\xi)\mid S\big)=\MI(U;W\mid S)$.
\end{enumerate}
Taking $S$ constant gives the unconditional versions.

\emph{Proof.} Fix $s$ and consider the integrand of \Cref{eq:cmi_def}. For (a), the map $(u,w)\mapsto(u,\varphi_s(w))$ sends $P_{UW\mid S=s}$ to the conditional law of $(U,\varphi_S(W))$ given $S=s$, and sends the product $P_{U\mid S=s}\otimes P_{W\mid S=s}$ to the product of the corresponding marginals. \Cref{eq:dpi} gives the inequality for the integrand. For (b), conditionally on $S=s$ the variable $m(S)$ equals the constant $m(s)$. The map $(u,w)\mapsto\big((u,m(s)),w\big)$ is a bijection onto its image with measurable inverse, so the integrand is unchanged by \Cref{eq:dpi}. For (c), independence gives $P_{(U,W,\xi)\mid S=s}=P_{UW\mid S=s}\otimes P_\xi$ and $P_{U\mid S=s}\otimes P_{(W,\xi)\mid S=s}=P_{U\mid S=s}\otimes P_{W\mid S=s}\otimes P_\xi$. Multiplying both arguments of a divergence by the same factor $P_\xi$ changes neither absolute continuity nor the density ratio, so the integrand is unchanged. Integrating over $P_S$ completes each case. $\square$

\textbf{Lemma 4 (smoothed targets have densities).} Let $y\in\mathbb{R}^{N\times D}$ be a random array with $\mathbb{E}\log(1+\|y^{(i)}\|)<\infty$ for every token $i$, and let $S$ be any random variable. Let $\xi$ be as in \Cref{app:var_setup}, independent of $(y,S)$, and let $\sigma>0$. Then, for $P_S$-almost every $s$, the conditional law of $\hat y_\sigma$ given $S=s$ has a density with respect to $\nu$, and
\begin{equation}
-N\log K_D-(D-1)\sum_{i=1}^{N}\mathbb{E}\log\!\left(1+\frac{\|y^{(i)}\|}{\sigma}\right)
\;\le\;H_\nu(\hat y_\sigma\mid S)\;\le\;N\log\omega(\mathbb{S}^{D-1}),
\label{eq:smooth_entropy}
\end{equation}
where $K_D=(2\pi)^{-(D-1)/2}\,2^{D-2}\max\{1,\mathbb{E}|\gamma|^{D-1}\}$ with $\gamma\sim\mathcal{N}(0,1)$. In particular, $H_\nu(\hat y_\sigma\mid S)$ is finite.

\emph{Proof.} \emph{Step 1: one token with a fixed mean.} Let $m\in\mathbb{R}^D$ and $Y=m+\sigma\zeta$ with $\zeta\sim\mathcal{N}(0,\mathbf{I}_D)$. In polar coordinates $Y=\varrho u$ with $\varrho>0$ and $u\in\mathbb{S}^{D-1}$, Lebesgue measure factorizes as $\varrho^{D-1}d\varrho\,\omega(du)$. Here $\omega$ is unnormalized surface measure, so no additional sphere-area factor is needed. Integrating out the radius gives the direction density
\begin{equation}
f_m(u)=\int_0^\infty(2\pi\sigma^2)^{-D/2}\exp\!\left(-\frac{\|\varrho u-m\|^2}{2\sigma^2}\right)\varrho^{D-1}\,d\varrho
\end{equation}
with respect to $\omega$. Let $b=u^\top m$. The orthogonal decomposition $m=bu+(m-bu)$ gives
\begin{equation}
\|\varrho u-m\|^2=(\varrho-b)^2+\|m-bu\|^2\ge(\varrho-b)^2.
\end{equation}
Dropping the nonnegative second term increases the exponential. Extending the resulting nonnegative integral to the real line, with $|\varrho|^{D-1}$ on the added negative half-line, gives
\begin{align}
f_m(u)&\le(2\pi\sigma^2)^{-D/2}\int_{-\infty}^{\infty}|\varrho|^{D-1}e^{-(\varrho-b)^2/(2\sigma^2)}\,d\varrho \notag\\
&=(2\pi)^{-(D-1)/2}\sigma^{-(D-1)}\,\mathbb{E}|b+\sigma\gamma|^{D-1}.
\end{align}
For $p=D-1\ge1$ and $x,x'\ge0$, convexity gives $(x+x')^p\le2^{p-1}(x^p+x'^p)$. Hence $\mathbb{E}|b+\sigma\gamma|^{D-1}\le2^{D-2}\big(\|m\|^{D-1}+\sigma^{D-1}\mathbb{E}|\gamma|^{D-1}\big)$, and
\begin{equation}
f_m(u)\le(2\pi)^{-(D-1)/2}2^{D-2}\left(\Big(\frac{\|m\|}{\sigma}\Big)^{D-1}+\mathbb{E}|\gamma|^{D-1}\right)\le K_D\left(1+\frac{\|m\|}{\sigma}\right)^{D-1},
\label{eq:pn_bound}
\end{equation}
where the last step uses $x^{p}+1\le(1+x)^{p}$ for $x\ge0$.

\emph{Step 2: the array.} Since $\xi$ is independent of $(y,S)$, conditionally on $(y,S)=(m,s)$ the tokens of $y_\sigma$ are independent with laws $\mathcal{N}(m^{(i)},\sigma^2\mathbf{I}_D)$. By Step~1, $\hat y_\sigma$ then has the conditional density $p(w\mid m,s)=\prod_{i}f_{m^{(i)}}(w^{(i)})$ with respect to $\nu$. By Tonelli's theorem, the mixture $p(w\mid s)=\mathbb{E}\big[p(w\mid y,s)\mid S=s\big]$ is a density of the conditional law of $\hat y_\sigma$ given $S=s$.

\emph{Step 3: upper bound and integrability of the negative part.} Write $M=\nu((\mathbb{S}^{D-1})^N)=\omega(\mathbb{S}^{D-1})^N$. For any density $q$ with respect to $\nu$,
\begin{equation}
\int q(\log q)^-\,d\nu\le M/e,
\qquad (\log q)^-=\max\{-\log q,0\},
\end{equation}
because $-x\log x\le1/e$ on $[0,1]$, with $0\log0=0$. Thus $\int q\log q\,d\nu$ is well defined, though it may be $+\infty$. Nonnegativity of $\mathrm{KL}(q\nu\,\|\,\nu/M)$ gives $-\int q\log q\,d\nu\le\log M$, which yields the upper entropy bound after averaging over $S$.

\emph{Step 4: lower bound.} Put $A(y)=N\log K_D+(D-1)\sum_i\log(1+\|y^{(i)}\|/\sigma)$. By \Cref{eq:pn_bound}, $\log p(w\mid y,S)\le A(y)$, and $\mathbb{E}|A(y)|<\infty$ by assumption. Together with Step~3 this gives absolute integrability of $\log p(\hat y_\sigma\mid y,S)$. Convexity of $x\log x$ and the mixture formula in Step~2 give
\begin{equation}
\int p(w\mid s)\log p(w\mid s)\,\nu(dw)
\le\mathbb{E}\!\left[\int p(w\mid y,s)\log p(w\mid y,s)\,\nu(dw)\,\middle|\,S=s\right].
\end{equation}
Jensen's inequality and the interchange of integrals are valid by applying them to the nonnegative convex function $x\log x+1/e$. Averaging over $S$ bounds the left-hand side by $\mathbb{E}A(y)<\infty$. Step~3 bounds its negative part, so $\mathbb{E}|\log p(\hat y_\sigma\mid S)|<\infty$. Negating the inequality gives the lower bound in \Cref{eq:smooth_entropy}, without assuming the entropy's finiteness in advance. $\square$

\textbf{Lemma 5 (cosine perturbation).} For nonzero $x,y,a\in\mathbb{R}^D$,
\begin{equation}
|\cos(a,x)-\cos(a,y)|\le\|\hat x-\hat y\|\le\min\!\left\{2,\frac{2\|x-y\|}{\|y\|}\right\}.
\end{equation}

\emph{Proof.} Write $\hat x-\hat y=(x-y)/\|y\|+x\,(1/\|x\|-1/\|y\|)$. The norm of the second term is $\big|\|y\|-\|x\|\big|/\|y\|\le\|x-y\|/\|y\|$. Hence $\|\hat x-\hat y\|\le2\|x-y\|/\|y\|$. The distance between two unit vectors is also at most two, and $|\cos(a,x)-\cos(a,y)|=|\hat a^\top(\hat x-\hat y)|\le\|\hat x-\hat y\|$ by the Cauchy--Schwarz inequality. $\square$

\subsection{Raw and Purified Targets}
\label{app:var_targets}

\textbf{Why the targets are smoothed.} Lemmas~1 and~2 require the target to have a density with respect to the reference measure. In our setting the normalized target $\hat z$ has none, for two independent reasons. First, training images are 8-bit and finitely many, so the law of $\hat z$ is atomic. Second, even if images were continuous, the encoder input has $3\cdot224^2=150{,}528$ coordinates for DINOv2 and $3\cdot256^2=196{,}608$ for MAE, whereas $(\mathbb{S}^{D-1})^N$ has dimension $N(D-1)=196{,}352$ for $D=768$ and $261{,}888$ for $D=1024$. The encoder and the normalization are locally Lipschitz, and a locally Lipschitz image of a lower-dimensional set has $\nu$-measure zero. The raw array law is therefore singular with respect to $\nu$, and its conditional laws remain supported on the same null set. We therefore state the bounds for the smoothed targets of \Cref{app:var_setup}, for which Lemma~4 proves the required densities and finite entropies, and we use Lemma~5 to return to the training loss.

For an array $y$ whose tokens are nonzero almost surely, define the target-only smoothing correction
\begin{equation}
\delta_\sigma(y)=\frac1N\sum_{i=1}^{N}\mathbb{E}\big\|\hat y_\sigma^{(i)}-\hat y^{(i)}\big\|\in[0,2].
\label{eq:smoothing_correction}
\end{equation}
It depends on the target law and $\sigma$, not on the denoiser or projector. Lemma~5 bounds the expected change in mean cosine by $\delta_\sigma(y)$, uniformly over all readouts. Moreover, $\hat y_\sigma^{(i)}\to\hat y^{(i)}$ almost surely as $\sigma\downarrow0$, and the distance is bounded by two. Dominated convergence therefore gives $\delta_\sigma(y)\to0$, without an inverse-norm moment assumption. Zero target tokens would require a separate convention for the training cosine; the unsmoothed bounds below assume they occur with probability zero.

\textbf{Proposition 3 (raw target).} Assume $\mathbb{E}\log(1+\|z^{(i)}\|)<\infty$ for every $i$. Then, for every $\sigma>0$, every $\kappa>0$, and every $(\theta,\psi)$,
\begin{equation}
\MI(h;z)\;\ge\;\MI(h;\hat z_\sigma)\;\ge\;H_\nu(\hat z_\sigma)+N\log c_D(\kappa)+\kappa N\,\mathbb{E}\!\left[\frac{1}{N}\sum_{i=1}^{N}\cos\!\left(a^{(i)},z_\sigma^{(i)}\right)\right],
\label{eq:raw_bound}
\end{equation}
where $H_\nu(\hat z_\sigma)$ is finite and does not depend on $(\theta,\psi)$. If moreover the tokens of $z$ are nonzero almost surely, then
\begin{equation}
\MI(h;z)\;\ge\;H_\nu(\hat z_\sigma)+N\log c_D(\kappa)-\kappa N\Big(\mathcal{L}_{\mathrm{align}}^{(t)}(z)+\delta_\sigma(z)\Big),
\label{eq:raw_bound_loss}
\end{equation}
with $\mathcal{L}_{\mathrm{align}}^{(t)}$ the alignment loss at noise time $t$ defined below \Cref{eq:vmf_loglik}.

\emph{Proof.} \emph{Step 1.} Since $\xi$ is independent of $(h,z)$, Lemma~3(c) gives $\MI(h;z)=\MI\big(h;(z,\xi)\big)$. Since $\hat z_\sigma$ is a measurable function of $(z,\xi)$, Lemma~3(a) gives $\MI\big(h;(z,\xi)\big)\ge\MI(h;\hat z_\sigma)$.

\emph{Step 2.} By Lemma~4 with $y=z$ and $S$ constant, $\hat z_\sigma$ has a density with respect to $\nu$ and $H_\nu(\hat z_\sigma)$ is finite. It depends only on the data, the frozen encoder, and $\sigma$. Apply Lemma~1 with $U=h$, $W=\hat z_\sigma$, and the decoder $q_\psi$ of \Cref{eq:vmf_decoder}. Its integrability assumption holds by the bound below \Cref{eq:vmf_loglik}. Substituting \Cref{eq:vmf_loglik} with $y=z_\sigma$ into \Cref{eq:ba_bound} gives \Cref{eq:raw_bound}.

\emph{Step 3.} Lemma~5 with $x=z_\sigma^{(i)}$ and $y=z^{(i)}$ gives $\cos(a^{(i)},z_\sigma^{(i)})\ge\cos(a^{(i)},z^{(i)})-\|\hat z_\sigma^{(i)}-\hat z^{(i)}\|$. Taking expectations and averaging over tokens gives $\mathbb{E}\big[\tfrac1N\sum_i\cos(a^{(i)},z_\sigma^{(i)})\big]\ge-\mathcal{L}_{\mathrm{align}}^{(t)}(z)-\delta_\sigma(z)$. Substituting into \Cref{eq:raw_bound} gives \Cref{eq:raw_bound_loss}. $\square$

\textbf{Averaging over noise times.} The time $t$ is drawn independently of the clean image, VAE posterior draw, and $\xi$, so the target-dependent constants are the same for every $t$. Taking the expectation over $t$ in \Cref{eq:raw_bound_loss} gives $\MI(h_t;z\mid t)\ge H_\nu(\hat z_\sigma)+N\log c_D(\kappa)-\kappa N\big(\mathcal{L}_{\mathrm{align}}+\delta_\sigma(z)\big)$. For the purified bound below, the corresponding left-hand side is $\sum_i\MI(h_t;z^{(i)}\mid S_i,t)$ and the loss is likewise averaged over $t$.

For the purified target, let $r=z-d_\phi(X_p)$ be the residual of \Cref{eq:lap_target}. Locality gives $r^{(i)}=z^{(i)}-d_{\phi,i}(S_i)$ and $r_\sigma^{(i)}=z_\sigma^{(i)}-d_{\phi,i}(S_i)$. The maps $d_{\phi,i}$ are fixed and measurable.

\textbf{Proposition 4 (purified target).} Assume $\mathbb{E}\log(1+\|r^{(i)}\|)<\infty$ for every $i$. Then, for every $\sigma>0$, every $\kappa>0$, and every $(\theta,\psi)$,
\begin{align}
\MI_{\mathrm{loc}}(h;z)
&\ge\sum_{i=1}^{N}\MI\big(h;\hat r_\sigma^{(i)}\mid S_i\big)\notag\\
&\ge\sum_{i=1}^{N}H_\omega\big(\hat r_\sigma^{(i)}\mid S_i\big)+N\log c_D(\kappa)
 +\kappa\sum_{i=1}^{N}\mathbb{E}\cos\!\left(a^{(i)},r_\sigma^{(i)}\right),
\label{eq:lap_bound}
\end{align}
where each conditional entropy is finite and does not depend on $(\theta,\psi)$. If moreover the tokens of $r$ are nonzero almost surely, then
\begin{equation}
\MI_{\mathrm{loc}}(h;z)\;\ge\;\sum_{i=1}^{N}H_\omega\big(\hat r_\sigma^{(i)}\mid S_i\big)+N\log c_D(\kappa)-\kappa N\Big(\mathcal{L}_{\mathrm{align}}^{(t)}\big(y^{\mathrm{pure}}\big)+\delta_\sigma(r)\Big).
\label{eq:lap_bound_loss}
\end{equation}

\emph{Proof.} \emph{Step 1: conditional data processing for one token.} Fix $i$. Since $\xi^{(i)}$ is independent of $(h,z^{(i)},S_i)$, Lemma~3(c) gives $\MI(h;z^{(i)}\mid S_i)=\MI\big(h;(z^{(i)},\xi^{(i)})\mid S_i\big)$.

Define $\varphi_s(w,u)$ as the normalization of $w+\sigma u-d_{\phi,i}(s)$, with an arbitrary value on its null set of zero vectors. It is jointly measurable, and $\hat r_\sigma^{(i)}=\varphi_{S_i}(z^{(i)},\xi^{(i)})$ almost surely. Lemma~3(a) therefore gives $\MI(h;z^{(i)}\mid S_i)\ge\MI(h;\hat r_\sigma^{(i)}\mid S_i)$.

\emph{Step 2: conditional variational bound.} Lemma~4 with $N=1$, $y=r^{(i)}$, and $S=S_i$ supplies a density with respect to $\omega$ and finite $H_\omega(\hat r_\sigma^{(i)}\mid S_i)$. Apply Lemma~2 with $U=h$, $W=\hat r_\sigma^{(i)}$, and the decoder $q_i(w\mid h,s)=\mathrm{vMF}(w;\hat a^{(i)},\kappa)$, which may ignore $s$. Its log-density is bounded and equals $\log c_D(\kappa)+\kappa\cos(a^{(i)},r_\sigma^{(i)})$ at the target. Summing this inequality and Step~1 over $i$ gives \Cref{eq:lap_bound}. This step requires no independence between target tokens or neighborhoods.

\emph{Step 3: the training target.} Lemma~5 and \Cref{eq:smoothing_correction} bound the change in the mean cosine by $\delta_\sigma(r)$. Since $\mathcal{L}_{\mathrm{align}}^{(t)}(r)=\mathcal{L}_{\mathrm{align}}^{(t)}(y^{\mathrm{pure}})$, replacing the smoothed score in \Cref{eq:lap_bound} gives \Cref{eq:lap_bound_loss}. $\square$

\textbf{The constants of \Cref{eq:info_bounds}.} \Cref{eq:raw_bound_loss,eq:lap_bound_loss} are the two inequalities of \Cref{eq:info_bounds}, with
\begin{align}
c_\sigma&=H_\nu(\hat z_\sigma)+N\log c_D(\kappa)-\kappa N\delta_\sigma(z),
\label{eq:main_constants}\\
c'_\sigma&=\sum_{i=1}^{N}H_\omega\big(\hat r_\sigma^{(i)}\mid S_i\big)+N\log c_D(\kappa)-\kappa N\delta_\sigma(r).
\notag
\end{align}
Both are determined by the data, target construction, $\sigma$, and $\kappa$; $c'_\sigma$ also depends on the specified neighborhoods. Neither depends on $(\theta,\psi)$. The loss bounds require $\mathbb{E}\log(1+\|y^{(i)}\|)<\infty$ and almost surely nonzero tokens for the respective target $y\in\{z,r\}$. The main text states the bounds at a fixed noise time; the remark on averaging over noise times above extends them to the time-averaged loss. The raw bound concerns joint information about $z$, whereas the purified bound concerns the sum in \Cref{eq:local_information}; they are not a decomposition of a common information quantity.

\textbf{Optimization direction.} Write either right-hand side of \Cref{eq:info_bounds} as $B_\sigma(y)=c_\sigma(y)-\kappa N\mathcal{L}_{\mathrm{align}}^{(t)}(y)$, where $c_\sigma(y)$ is the appropriate constant above, $c_\sigma(z)=c_\sigma$ and $c_\sigma(r)=c'_\sigma$. Since $\kappa N>0$, minimizing the alignment loss is equivalent to maximizing $B_\sigma(y)$. The sign is also explicit in the combined objective:
\begin{equation}
\mathcal{L}_{\mathrm{flow}}^{(t)}+\lambda_a\mathcal{L}_{\mathrm{align}}^{(t)}(y)
=\mathcal{L}_{\mathrm{flow}}^{(t)}-\frac{\lambda_a}{\kappa N}B_\sigma(y)
 +\frac{\lambda_a}{\kappa N}c_\sigma(y).
\label{eq:alignment_sign}
\end{equation}
For $\lambda_a>0$, minimizing this objective rewards a larger information lower bound alongside a smaller denoising error. No entropy estimate or additional information loss is needed in training. This is an equivalence between objectives, not a guarantee that either the bound or mutual information increases at every optimization step. The constants differ across targets, so alignment losses alone cannot rank their mutual information.

\textbf{The role of $\sigma$.} Both propositions hold for every $\sigma>0$ under their stated assumptions. Although $\delta_\sigma(y)\to0$ for almost surely nonzero targets, the entropy term tends to $-\infty$ when the corresponding unsmoothed target law is singular. The following argument applies to the raw array and separately to each purified token.

For the raw target take $y=z$, $S$ constant, and reference measure $\nu_\star=\nu$; for a purified token take $y=r^{(i)}$, $S=S_i$, and $\nu_\star=\omega$. Assume $y$ has nonzero tokens almost surely and $P_{\hat y\mid S=s}\perp\nu_\star$ for $P_S$-almost every $s$. Set $M_\star=\nu_\star(\operatorname{supp}\nu_\star)$, $Q=\nu_\star/M_\star$, and $P_\sigma=P_{(S,\hat y_\sigma)}$. Using the same $(y,S,\xi)$ for every $\sigma$ gives $\hat y_\sigma\to\hat y$ almost surely, by continuity of normalization at nonzero tokens. Thus $P_\sigma\Rightarrow P_0=P_{(S,\hat y)}$. The KL chain rule gives
\begin{equation}
\mathrm{KL}(P_\sigma\,\|\,P_S\otimes Q)=\log M_\star-H_{\nu_\star}(\hat y_\sigma\mid S).
\end{equation}
The singular conditional limit has $\mathrm{KL}(P_0\,\|\,P_S\otimes Q)=\infty$. Weak lower semicontinuity of KL~\citep[Theorem~3.6]{polyanskiy2016notes} therefore implies, for every sequence $\sigma_n\downarrow0$,
\begin{equation}
\infty=\mathrm{KL}(P_0\,\|\,P_S\otimes Q)
\le\liminf_{n\to\infty}\mathrm{KL}(P_{\sigma_n}\,\|\,P_S\otimes Q).
\end{equation}
Consequently $H_{\nu_\star}(\hat y_\sigma\mid S)\to-\infty$. No monotonicity in $\sigma$ is needed. Raw-array singularity was established above. For LAP with discrete encoder inputs, conditioning on $S_i=s$ leaves an atomic law of $z^{(i)}$; subtracting the fixed vector $d_{\phi,i}(s)$ and normalizing preserves atomicity. Thus each conditional token entropy tends to $-\infty$, and so does their finite sum. This concerns conditional residual laws, not their marginals. With fixed $\kappa$, the bounded cosine term cannot offset the diverging entropy. The bounds therefore become vacuous as $\sigma\downarrow0$, even though the loss correction vanishes. \Cref{app:var_usable} avoids this smoothing limitation.

\textbf{The squared-error form.} The same token-wise construction uses a Gaussian decoder whose mean adds a correction from $h$ to the local prediction. Assume $\mathbb{E}\|r\|^2<\infty$ and $\mathbb{E}\|g_\psi(h)\|^2<\infty$. Conditionally on $S_i$, $z_\sigma^{(i)}$ has a Lebesgue density, since it is a Gaussian convolution. Translation by $d_{\phi,i}(S_i)$ leaves its conditional entropy unchanged. This entropy is at least $\tfrac D2\log(2\pi e\sigma^2)$, the entropy of the added noise, and at most $\tfrac D2\log\big(2\pi e\,\mathbb{E}\|r_\sigma^{(i)}\|^2/D\big)$ by the Gaussian maximum-entropy bound and Jensen's inequality. Both bounds are finite. Lemma~3 gives $\MI(h;z^{(i)}\mid S_i)\ge\MI(h;z_\sigma^{(i)}\mid S_i)$. For $\tau>0$, choose $q_i(z_\sigma^{(i)}\mid h,s)=\mathcal{N}\big(z_\sigma^{(i)};d_{\phi,i}(s)+a^{(i)},\tau^2\mathbf{I}_D\big)$. Applying Lemma~2 to each token and summing, then using independence and the zero mean of $\xi$ to obtain $\mathbb{E}\|r_\sigma-g_\psi(h)\|^2=\mathbb{E}\|r-g_\psi(h)\|^2+\sigma^2ND$, gives
\begin{equation}
\MI_{\mathrm{loc}}(h;z)\;\ge\;\sum_{i=1}^{N}H_{\mathrm{Leb}}\big(z_\sigma^{(i)}\mid S_i\big)-\frac{ND}{2}\log(2\pi\tau^2)-\frac{1}{2\tau^2}\Big(\mathbb{E}\big\|r-g_\psi(h)\big\|^2+\sigma^2ND\Big).
\label{eq:lap_bound_mse}
\end{equation}
Minimizing residual regression error therefore maximizes a lower bound on the sum of token-wise conditional information terms, with each local prediction supplied to its decoder as side information.

\textbf{Batchwise coefficients.} Proposition~4 covers LAP-N and the frozen dataset-level linear predictor of \Cref{app:controls}. For LAP-L, the affine coefficients $\widehat W$ are refitted on each batch and are random. The same token-wise proof applies with $S_i=(X_p^{(i)},\widehat W)$, giving a bound on $\sum_i\MI(h;z^{(i)}\mid X_p^{(i)},\widehat W)$. The coefficients contain batch-level information, so this is not identical to conditioning on the local patch alone. All target statistics use only clean latents and frozen encoder features, so their laws do not depend on $(\theta,\psi)$. Stop-gradient prevents backpropagation through the target construction.

\textbf{What the conditional terms reward.} For a fixed token $i$ and any measurable function $m_i$, Lemma~3(b) gives
\begin{equation}
\MI\big((h,m_i(S_i));\,z^{(i)}\mid S_i\big)=\MI(h;z^{(i)}\mid S_i).
\end{equation}
Thus appending a function of that token's side information cannot increase its conditional information term. This invariance is term-specific: appending functions of other neighborhoods, or of the full clean latent, can increase it. Nor is the sum invariant to appending all $m_j(S_j)$ at once. By contrast, the raw joint information satisfies
\begin{equation}
\MI\big((h,m_i(S_i));\,z\big)=\MI(h;z)+\MI\big(m_i(S_i);z\mid h\big)\;\ge\;\MI(h;z).
\end{equation}
If $\MI(h;z)<\infty$, this inequality is strict whenever $\MI(m_i(S_i);z\mid h)>0$. These are statements about the information quantities on the left-hand sides, not invariances of the residual loss. A raw-target decoder also admits a token-wise conditional bound, so the existence of such a bound alone does not distinguish LAP from raw alignment. The chain rule applied separately at every token gives
\begin{equation}
\sum_{i=1}^{N}\MI\big(h;(z^{(i)},S_i)\big)
=\sum_{i=1}^{N}\MI(h;S_i)+\MI_{\mathrm{loc}}(h;z),
\label{eq:lap_chain}
\end{equation}
which separates each token's local side information from its target information beyond that side information. Proposition~4 bounds the second sum. Overlapping neighborhoods and target tokens can duplicate information across terms, so this is not a decomposition of the raw joint quantity $\MI(h;z)$.

\textbf{Relation to the denoising loss.} In our SiT implementation, every block after block $l$ and the final layer act on the residual stream together with the conditioning vector $e$. There are no other skip connections. Hence $v_\theta(x_t,t)$ is a function of $(h,e)$. The velocity $v^*=\varepsilon-X$ has a smooth Lebesgue density, since $\varepsilon$ is Gaussian and independent of $X$, and its entropy is finite when $X$ has finite second moments. No smoothing is needed here. Write $d$ for the dimension of the latent and assume $\mathbb{E}\|v_\theta(x_t,t)-v^*\|^2<\infty$. Lemma~1 with $U=(h,e)$, $W=v^*$, Lebesgue reference measure, and the decoder $\mathcal{N}(v^*;v_\theta,\tau^2\mathbf{I})$ gives
\begin{equation}
\MI\big((h,e);v^*\big)\;\ge\;H_{\mathrm{Leb}}(v^*)-\frac{d}{2}\log(2\pi\tau^2)-\frac{1}{2\tau^2}\,\mathbb{E}\big\|v_\theta(x_t,t)-v^*\big\|^2 .
\end{equation}
The last expectation is $\mathcal{L}_{\mathrm{flow}}$ evaluated at time $t$. Minimizing it therefore maximizes a lower bound on the information that $(h,e)$ carries about $v^*$. The data-dependent part of the velocity is $-X$, but this does not identify $\mathcal{L}_{\mathrm{flow}}$ with the sum of local-information terms $\sum_i\MI(h;S_i)$ in \Cref{eq:lap_chain}. The connection motivates supervision that complements reconstruction, without proving that the two objectives capture disjoint information.

\textbf{Bounds versus bounded quantities.} Raising a lower bound does not by itself raise the bounded quantity. For the smoothed raw target, the Barber--Agakov step has an expected conditional-KL gap; for the purified bound there is one such gap per token and side-information variable. The bounds on the original targets also include data-processing gaps and cosine-perturbation slack. These gaps need not remain fixed during training. The inequalities do not predict a generation metric or guarantee a gain from purification.

\subsection{An Exact Identity for Usable Information}
\label{app:var_usable}

Shannon information is a blunt description of a deterministic network (\Cref{app:var_scope}). A more direct comparison asks how well a projector can predict the target relative to a constant readout. Following the predictive-score construction of \citet{xu2020vinfo}, this comparison has an exact expression with no density assumption and no smoothing.

Let $w\in(\mathbb{S}^{D-1})^N$ be any random target, such as $\hat z$ or $\hat r$. Consider the predictive families
\begin{equation}
\mathcal{V}_h=\{q_\psi(\cdot\mid h):\psi\in\Psi\},
\qquad
\mathcal{V}_\varnothing=\Big\{\textstyle\prod_{i}\mathrm{vMF}\big(\cdot\,;\mu,\kappa\big):\mu\in\mathbb{S}^{D-1}\Big\},
\end{equation}
where $\mathcal{V}_\varnothing$ uses one constant direction for all tokens and all inputs. Define the predictive entropies $H_{\mathcal{V}}(w\mid h)=\inf_{\psi}\mathbb{E}\big[-\log q_\psi(w\mid h)\big]$ and $H_{\mathcal{V}}(w\mid\varnothing)=\inf_{q\in\mathcal{V}_\varnothing}\mathbb{E}\big[-\log q(w)\big]$, and the usable information $\MI_{\mathcal{V}}(h\to w)=H_{\mathcal{V}}(w\mid\varnothing)-H_{\mathcal{V}}(w\mid h)$.

\textbf{Assumption 1.} For every $\mu\in\mathbb{S}^{D-1}$ there is $\psi\in\Psi$ with $g_\psi(h)^{(i)}=\mu$ for all $h$ and $i$. This holds for the token-wise MLP projector, by setting the weights of its last layer to zero and its bias to $\mu$.

For our shared token-wise projector, write $g_\psi(h)^{(i)}=G_\psi(h^{(i)})$ and let $J$ be an independent uniform token index. The score defined above is $N$ times the usual $\mathcal{V}$-information from $h^{(J)}$ to $w^{(J)}$, with $J$ itself not supplied to the decoder. Assumption~1 supplies every constant vMF direction required by optional ignorance in this token-level formulation. The shared-direction baseline is not the family of all input-independent distributions over whole arrays. Position-dependent constant readouts are excluded from the baseline by design; since $h$ carries positional embeddings, realizing them requires reading $h$.

\textbf{Proposition 5.} Let $\bar m=\mathbb{E}\big[\tfrac1N\sum_{i=1}^{N}w^{(i)}\big]\in\mathbb{R}^D$ be the mean of the normalized target tokens. Then
\begin{equation}
\MI_{\mathcal{V}}(h\to w)=\kappa N\left(\sup_{\psi\in\Psi}\mathbb{E}\!\left[\frac1N\sum_{i=1}^{N}\cos\!\left(g_\psi(h)^{(i)},w^{(i)}\right)\right]-\|\bar m\|\right),
\label{eq:usable_identity}
\end{equation}
and under Assumption~1 this quantity is nonnegative.

\emph{Proof.} \emph{Step 1: the score with access to $h$.} Let $C_\psi=\mathbb{E}[\tfrac1N\sum_i\cos(g_\psi(h)^{(i)},w^{(i)})]$. By \Cref{eq:vmf_loglik}, the predictive entropy is
\begin{align}
H_{\mathcal V}(w\mid h)
&=\inf_{\psi}\big\{-N\log c_D(\kappa)-\kappa N C_\psi\big\} \notag\\
&=-N\log c_D(\kappa)-\kappa N\sup_{\psi}C_\psi.
\end{align}
The infimum becomes a supremum because the coefficient of $C_\psi$ is negative.

\emph{Step 2: the best constant readout.} A shared unit direction $\mu$ earns expected cosine
\begin{equation}
\mathbb{E}\!\left[\frac1N\sum_i\mu^\top w^{(i)}\right]
=\mu^\top\bar m\le\|\bar m\|,
\label{eq:constant_optimum}
\end{equation}
by linearity of expectation and the Cauchy--Schwarz inequality. If $\bar m\neq0$, equality holds at $\mu^*=\bar m/\|\bar m\|$. If $\bar m=0$, every unit direction earns zero. Thus the optimum is $C_0=\|\bar m\|$, and
\begin{equation}
H_{\mathcal V}(w\mid\varnothing)=-N\log c_D(\kappa)-\kappa N C_0.
\end{equation}

\emph{Step 3: subtracting the baseline.} The normalizing constants cancel when the two predictive entropies are subtracted, giving
\begin{equation}
H_{\mathcal V}(w\mid\varnothing)-H_{\mathcal V}(w\mid h)
=\kappa N\big(\sup_\psi C_\psi-C_0\big).
\end{equation}
Under Assumption~1, the projector can realize the optimal constant direction, so $\sup_\psi C_\psi\ge C_0$. This proves both the identity and its nonnegativity. $\square$

In the notation of \Cref{sec:baseline}, $\|\bar m\|=C_0$ and the supremum is $\sup_\psi C_\psi$, so \Cref{eq:usable_identity} reads $\MI_{\mathcal V}(h\to w)=\kappa N(\sup_\psi C_\psi-C_0)$: the usable information is the gain of the best projector over the constant-readout baseline of \Cref{eq:constant_baseline}.

\textbf{Computing the baseline.} Given $n$ target arrays with $N$ tokens each, normalize each token, average the resulting directions, and take the norm:
\begin{equation}
\widehat C_0(y)=\left\|\frac{1}{nN}\sum_{j=1}^{n}\sum_{i=1}^{N}
\frac{y_j^{(i)}}{\|y_j^{(i)}\|}\right\|.
\label{eq:constant_estimator}
\end{equation}
This is the exact constant-readout optimum for the empirical target distribution and an estimate of the population baseline. It requires only target features, with no denoiser or fitted probe. For batchwise purification, we first form the targets in the same batches as the training procedure, then pool their normalized directions (\Cref{app:constant_baseline}). The baseline uses one direction shared across positions and images, as specified above.

\textbf{The constant-readout reference.} The best shared constant direction attains cosine $C_0=\|\bar m\|$ without reading $h$. \Cref{app:constant_baseline} measures $C_0$ for every target. For raw MAE it is $0.93$, compared with $0.957$ for the trained projector. Their difference, about $0.03$, is illustrative because the two estimates use different samples. Since $C_0$ is computed from unit-normalized tokens, even a zero-mean affine residual can have $C_0>0$. The measured difference is not a certified population bound or an upper bound on \Cref{eq:usable_identity}, which optimizes over the entire projector family. LAP-L and LAP-N subtract a batchwise affine prediction and a frozen nonlinear local prediction, respectively, before this comparison is made.

\textbf{Prediction gain over a fixed baseline.} With the Gaussian families $\{\mathcal{N}(z;\,d_\phi(X_p)+b,\,\tau^2\mathbf{I}):b\in\mathbb{R}^D\}$ and $\{\mathcal{N}(z;\,d_\phi(X_p)+g_\psi(h),\,\tau^2\mathbf{I}):\psi\in\Psi\}$, where $b$ is shared across tokens, the analogous log-score improvement is
\begin{equation}
\Delta_{\mathrm{pred}}(h;d_\phi)=\frac{1}{2\tau^2}\left(\inf_{b}\,\mathbb{E}\|r-b\|^2-\inf_{\psi}\,\mathbb{E}\|r-g_\psi(h)\|^2\right).
\end{equation}
Here $r-b$ subtracts $b$ from every token. With a finite residual second moment and a correction family containing every shared constant $b$, this score is finite and nonnegative. It measures improvement over the specified frozen baseline. We do not identify it with conditional $\mathcal{V}$-information in the full sense of \citet{hewitt2021conditional}, which imposes an optional-ignorance condition on the joint predictive family. The fixed baseline and restricted correction family need not satisfy that condition.

\subsection{Scope}
\label{app:var_scope}

\textbf{Deterministic hidden states.} Since $h$ is a function of $(x_t,e)$, \Cref{eq:dpi} gives $\MI(h;z)\le\MI\big((x_t,e);z\big)$, with equality whenever that function is injective. Shannon information therefore does not distinguish representations that differ by a bijection, and it cannot increase beyond what the noisy input contains. Training directly changes the score of the current projector. Proposition~5 characterizes the best score in the family, which a trained projector need not attain. Neither is an identity for Shannon information.

\textbf{Local versus full-latent conditioning.} Proposition~4 conditions each token only on its predictor's local input $S_i$, while $h$ can carry information from outside that neighborhood. Conditioning instead on the full clean latent would remove this interpretation. To see why, let $c$ be the class label and assume the forward noise and label-dropout randomness are independent of the clean data. At fixed $t$, the model construction implies $h\perp z\mid(X_p,c)$ and hence
\begin{equation}
\MI(h;z\mid X_p)\le\MI(c;z\mid X_p)\le H(c\mid X_p).
\end{equation}
Without class conditioning, the left-hand side is exactly zero; it is also zero after conditioning on $(X_p,e)$. This follows from the forward-noising Markov structure, whether or not the VAE latent determines the image. The local sum in \Cref{eq:local_information} avoids conditioning away all image input, but it can still be small when a local neighborhood already predicts its teacher token. It also changes when the neighborhood changes. The finite receptive field and capacity of $d_\phi$ define an operational target transformation, without guaranteeing a semantic residual (\Cref{sec:conclusion}).

\textbf{Residualization is not conditional independence.} A residual can remain predictable from its local input. For example, consider one target token with local input $S$ uniform on $\{-1,0,1\}$ and target $z=(S^2,1)$. Since $S$ and $S^2$ are uncorrelated, the least-squares affine prediction is $(2/3,1)$ and $r=(S^2-2/3,0)$. Reading $h=S$ achieves cosine $1$ through $g(h)=(h^2-2/3,0)$, while the best constant direction attains only $1/3$, even though $\MI(h;z\mid S)=0$. This example illustrates limited predictor capacity and is not a claim about a trained diffusion model. A conditional information bound therefore does not establish that the residual loss ignores local content. LAP's selectivity comes from its predictor class and must be assessed empirically.

\textbf{No account of harm.} Neither the information bounds nor the predictive-score identity establishes that redundant content harms generation. Explaining why the predictable-only target can be worse than no alignment requires an argument about optimization or model capacity, which this analysis does not provide.

\textbf{Relation to information-maximizing diffusion.} InfoDiffusion~\citep{wang2023infodiffusion} learns an auxiliary latent $u$ through an encoder $q_\phi(u\mid x_0)$, written $u$ here to avoid a clash with the teacher features $z$, and conditions the reverse process on it. Its regularized objective has the form
\begin{equation}
\mathbb{E}_{q}\big[\mathcal{L}_{D}(x_0)\big]+\zeta\,\MI_q(x_0;u)-\beta\,\mathrm{D}\big(q_\phi(u)\,\|\,p(u)\big),
\end{equation}
which is maximized for positive weights $\zeta$ and $\beta$. Here $\mathcal L_D$ is the diffusion evidence lower bound, $\MI_q(x_0;u)$ is the data--latent information under the learned joint distribution, and $q_\phi(u)$ is the aggregate posterior. The positive information term encourages informative latents and addresses their being ignored by the diffusion decoder. InfoDiffusion obtains a tractable objective through a decomposition involving the encoder and aggregate-posterior divergences.

Alignment instead starts with a frozen teacher $z=f(I)$. Its purpose is to shape intermediate denoiser features so that a projector can read the chosen teacher content from them. The teacher is a training target and is not supplied to the denoiser at inference. The vMF argument gives a lower bound on $\MI(h;z)$, and \Cref{eq:alignment_sign} shows that the alignment term rewards a larger bound. This differs from learning the auxiliary latent and regularizing its posterior in InfoDiffusion.

LAP further selects the target content through a fixed local predictor. Its residual objective admits the bound of \Cref{eq:lap_bound_loss} on the sum of conditional information terms for teacher tokens given their local predictor inputs. This supplies a formal connection to information beyond those neighborhoods. The bound alone does not establish such a separation in the learned representation: a raw-target decoder also admits a token-wise conditional bound, and residuals can remain locally predictable. The substantive change is the target supplied to the restricted projector. Its benefit is assessed by the interventions and generation results, while the information analysis clarifies the objective and its optimization direction.

\FloatBarrier
\section{Experimental and Diagnostic Protocols}
\label{app:probes}

This appendix defines the measurements used in the main text and shows how they vary with noise level, subspace rank, and intervention type. Unless stated otherwise, diagnostics use seed 0 of a 400K-step ImageNet-1k checkpoint. Generation results use the seed coverage stated in each table.

\subsection{Training and Evaluation Settings}

Our base model is SiT-B/2 with 12 blocks and hidden width 768. The SD-VAE produces a $4\times32\times32$ latent. Patches of size $2\times2$ give 256 tokens, each containing 16 latent values. Alignment is applied after block 8 through a three-layer MLP with hidden width 2048 and SiLU activations. The alignment coefficient is $\lambda_a=0.5$. The main teacher comparison uses MAE ViT-L/16, AIMv2 ViT-L/14, CLIP ViT-L/14~\citep{radford2021clip}, DINOv3 ViT-B/16, and DINOv2 ViT-B/14. Development comparisons also include I-JEPA ViT-H/14~\citep{assran2023ijepa} and MoCo v3 ViT-B/16~\citep{chen2021mocov3}. The original $256\times256$ image is preprocessed separately for each encoder to obtain a $16\times16$ patch grid, as specified in \Cref{tab:teacher_preprocessing}. No interpolation of the extracted feature grid is used.

\begin{table}[H]
\centering\small
\setlength{\tabcolsep}{4pt}
\caption{\textbf{Teacher preprocessing and feature extraction.} Input is the actual backbone resolution after resizing the original $256\times256$ RGB image. All outputs contain 256 patch tokens. IN denotes ImageNet normalization, CLIP denotes OpenAI CLIP normalization, and half denotes mean and standard deviation $0.5$ per channel. CLS and register tokens are excluded. Recorded checkpoint identifiers and normalization constants are in the reproducibility manifest.}
\label{tab:teacher_preprocessing}
\begin{tabular}{lcrrl}
\toprule
Teacher & Input & Norm. & $D$ & Extracted features \\
\midrule
MAE B/L/16 & 256 & IN & 768/1024 & Last block, before final LN \\
DINOv2 S/B/L/14 & 224 & IN & 384/768/1024 & Normalized patch tokens \\
CLIP L/14 & 224 & CLIP & 1024 & Before final LN/projection \\
CLIP B/16 & 256 & CLIP & 768 & Before final LN/projection \\
I-JEPA H/14 & 224 & IN & 1280 & Final normalized tokens \\
MoCo v3 B/16 & 256 & IN & 768 & Backbone patch tokens \\
AIMv2 L/14 & 224 & CLIP & 1024 & Final hidden-state output \\
DINOv3 B/L/H+/16 & 256 & IN & 768/1024/1280 & Backbone patch tokens \\
SigLIP 2 B/16 & 256 & half & 768 & Vision hidden-state output \\
PE-Core B/16, L/14 & 256, 224 & half & 768/1024 & Backbone patch tokens \\
PE-Spatial B/16 & 256 & half & 768 & Backbone patch tokens \\
Web-SSL DINO-300M & 224 & IN & 1024 & Final hidden-state output \\
EVA-02 B/14 & 224 & CLIP & 768 & Backbone patch tokens \\
\bottomrule
\end{tabular}
\end{table}

RGB values are divided by 255. ImageNet normalization uses mean $(0.485,0.456,0.406)$ and standard deviation $(0.229,0.224,0.225)$; CLIP uses $(0.48145466,0.4578275,0.40821073)$ and $(0.26862954,0.26130258,0.27577711)$. Tensor resizing is bicubic with \texttt{align\_corners=False} and \texttt{antialias=False}. For CLIP L/14 resizing precedes normalization; for the other 224-pixel backbones it follows normalization. Absolute positional embeddings are resampled when the loading configuration changes the patch grid, including MAE B/L and CLIP B/16 at 256 pixels. MAE and CLIP omit the final feature LayerNorm in their alignment targets; the other wrappers retain the backbone's returned features.

The reproducibility manifest accompanying the code records checkpoint identifiers, source-file hashes, normalization and extraction rules, data counts, and validation results.

Dev-200 retains original class IDs $0,\ldots,199$ from the training split of \path{evanarlian/imagenet_1k_resized_256}, in source streaming order, with no random class selection or label remapping. It contains 255,224 images. The processed ImageNet-1k corpus contains 1,272,453 images across all 1,000 classes, from the training shards of \path{ILSVRC/imagenet-1k}. These counts match unique stored image stems and their cached latents; they are not counts of flip-augmented records. For full-scale preprocessing, each RGB image is resized bicubically so that its shorter side is 256 pixels and then center-cropped to $256\times256$. Dev-200 uses the source's resized images, resizing to $256\times256$ if needed. Class counts and hashes of both latent-label manifests are provided in the reproducibility manifest.

Base runs use batch 64, AdamW with constant learning rate $10^{-4}$, $(\beta_1,\beta_2)=(0.9,0.999)$, zero weight decay, $\epsilon=10^{-8}$, and gradient-norm clipping at 1.0. We use fp16 with TF32 enabled, uniformly sampled noise times, and EMA decay $0.9999$ per optimizer update. Class labels are dropped once inside the model with probability $0.1$. The image loader applies no random crop or horizontal flip. A fresh posterior sample is drawn from the cached SD-VAE moments whenever an image is used, with latent scale $0.18215$. Target comparisons on dev-200 train for 50K steps. The DINOv2 auxiliary-prediction sweep and its projection controls train for 100K steps. The full ImageNet-1k comparisons train for 400K steps. The larger-batch and larger-model settings are listed separately in \Cref{app:regime}.

The development metric, FID-DINOv2, uses the final normalized CLS feature of DINOv2 ViT-B/14 (dimension 768). It compares 10K EMA-generated images with a fresh shuffled sample of 10K training images. Sampling uses 50 Euler ODE steps without guidance and independent uniformly sampled class labels. Generated images are clamped in floating point before feature extraction. This evaluator does not fix an evaluation RNG seed or reuse a cached real-feature reference, so repeated evaluations can vary independently of the training seed. Standard FID-50K uses a different feature space and sampling protocol, described in \Cref{app:adm}. The two metrics should be compared only within their respective settings.

\subsection{Linear Probes and Subspace Interventions}

The main hidden-state diagnostics use 1,024 images for fitting and 256 disjoint images for evaluation. These images are drawn at random from the ImageNet-1k training data. They are held out from probe fitting, not from diffusion training. Every checkpoint receives the same images, VAE posterior samples, noise, and times, with $t\sim\mathcal U[0.3,0.7)$. Sampling 80 patch tokens per image gives 81,920 fit tokens and 20,480 evaluation tokens. The component probes use the separate sampling rule below.

For a probe from features $H$ to responses $Y$, we center both matrices on the fit split and solve
\begin{equation}
W=(H_c^\top H_c+\lambda I)^{-1}H_c^\top Y_c,
\qquad
\lambda=10^{-3}\frac{\operatorname{tr}(H_c^\top H_c)}{d}.
\end{equation}
Here $d$ is the feature dimension of $H$. The intercept restores the fit-split means. We report evaluation-split $R^2$, computed as one minus total squared prediction error divided by response variation around the evaluation-split mean. Negative values indicate worse prediction than this mean baseline. The relative penalty scales with feature magnitude. The image-disjoint evaluation uses FP32 inference and FP64 ridge solves, without tuning on evaluation scores.

\textbf{What each probe measures.} Presence probes the hidden state for target features. Prediction utility probes the target for clean-latent patches. Velocity sufficiency probes the aligned representation for velocity patches. These quantities have different inputs and outputs and should not be treated as interchangeable.

\textbf{Which directions are removed.} We form the aligned subspace from the top-$r$ left singular vectors of the probe matrix for $h\to a_t$. A target-probe control instead uses the matrix for $h\to y$. The main-text subspaces use only the fitting images and rank $r=256$. Supplementary rank sweeps are reported separately below.

For a projector $\Pi$ and bin-conditional hidden-state mean $\bar h_b$, mean filling replaces the hidden state by
\begin{equation}
h^{\mathrm{ablate}}=(I-\Pi)h+\Pi\bar h_b.
\end{equation}
The remaining network then runs normally. For the main-text results, $\bar h_b$ is estimated from all tokens of the fitting images. Baseline, full-layer, and subspace effects use identical evaluation images and noise draws. We divide the increase in velocity MSE by the full-layer increase. The random reference averages five fixed rank-256 subspaces. A supplementary resampling control replaces $\Pi h$ using a random permutation of images in the batch, preserving patch positions.

These are interventions on one layer, not measurements of every pathway by which alignment affects training. Downstream computation can compensate for a local ablation~\citep{mcgrath2023hydra}. Compensation may differ between subspace and full-layer interventions, so the normalized scores are most useful for matched comparisons.

\subsection{Image-Disjoint Results and Split Sensitivity}
\label{app:image_disjoint}

\Cref{tab:image_disjoint} reports the main-text diagnostics and compares presence under three evaluation protocols. The paired token split uses exactly the same feature pool and fit/evaluation token counts as the image split, but allows tokens from every image on both sides. The dense token-split protocol uses a separate dense-token collection. Differences between this protocol and the image split therefore reflect both sampling and the split, and cannot all be attributed to image overlap.

\begin{table}[t]
\centering\small
\setlength{\tabcolsep}{4pt}
\caption{\textbf{Image-disjoint diagnostics.} ImageNet-1k/400K, mid-noise, seed 0 unless marked. Presence always probes raw DINOv2. ``Token (dense)'' and ``Token (paired)'' denote dense and matched-pool token splits, respectively. Velocity $R^2$ and use are evaluated on disjoint images. Use is a percentage of full-layer mean-fill ablation.}
\label{tab:image_disjoint}
\begin{tabular}{lrrrrr}
\toprule
 & \multicolumn{3}{c}{DINOv2 presence $R^2$} & Velocity & Use \\
\cmidrule(lr){2-4}
Model & Token (dense) & Token (paired) & Image-split & $R^2$ & (\%) \\
\midrule
Unaligned & 0.196 & 0.166 & 0.130 & 0.451 & 14.03 \\
LAP-L (MAE) & 0.268 & 0.226 & 0.172 & 0.114 & 15.38 \\
LAP-N (MAE) & 0.290 & 0.240 & 0.178 & 0.004 & 10.61 \\
REPA & 0.343 & 0.286 & 0.207 & 0.115 & 12.21 \\
LAP-L (DINOv2) & 0.342 & 0.289 & 0.213 & $-0.016$ & 6.67 \\
LAP-N (DINOv2) & 0.361 & 0.297 & 0.213 & $-0.066$ & 4.18 \\
REPA + prediction loss & 0.297 & 0.255 & 0.197 & 0.534 & 18.86 \\
\midrule
REPA (seed 1) & 0.348 & 0.291 & 0.208 & 0.112 & 10.64 \\
REPA + prediction loss (seed 1) & 0.300 & 0.256 & 0.197 & 0.533 & 11.51 \\
\bottomrule
\end{tabular}
\end{table}

The main comparisons persist across images. LAP-N (MAE) closes $62.8\%$ of the unaligned-to-REPA presence gap (95\% interval: $60.5$--$65.0\%$). For the auxiliary loss, velocity $R^2$ rises from $0.115/0.112$ to $0.534/0.533$ in seeds 0/1. Aligned-subspace use rises from $12.2/10.6\%$ to $18.9/11.5\%$, with a smaller change in seed 1. Under the same protocol, the five rank-256 random subspaces of the seed-0 REPA model give $22.2$--$23.0\%$, with mean $22.7\%$ (95\% interval: $21.7$--$23.7\%$). This is the random reference in \Cref{fig:causal_tests}(a). For DINOv2 purification, presence is retained while velocity prediction and normalized use decrease. We show absolute probe $R^2$ in \Cref{fig:inside_model}(b), including values below zero, and display use on a separate axis.

The image-bootstrap intervals in \Cref{fig:causal_tests,fig:content_diagnostics,fig:inside_model} use 2,000 resamples of evaluation images, with identical resamples for paired comparisons. Each fitted probe remains fixed, and each bootstrap $R^2$ uses the response mean of that resample. Component intervals also condition on the fitted batch decompositions. These intervals quantify evaluation-image variability. They do not cover probe-refitting uncertainty, additional training seeds, or new random-subspace draws.

\subsection{Representation Changes after Purification}
\label{app:inside}

\begin{figure}[H]
\centering
\setlength{\parskip}{0pt}
\setlength{\abovecaptionskip}{5pt}
\includegraphics[width=\linewidth]{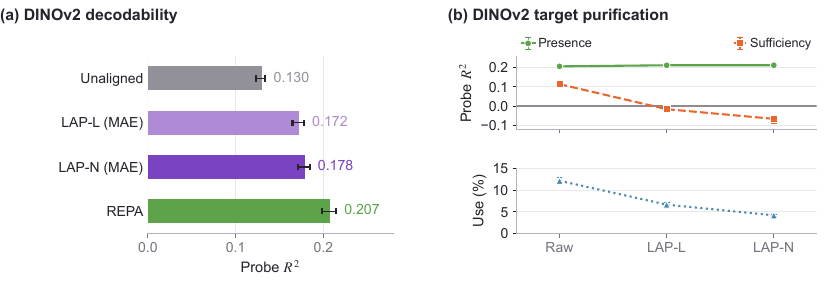}
\vspace{-25pt}
\caption{\textbf{Purification preserves readable features while reducing prediction coupling.}
(a) DINOv2 decodability after MAE-based alignment.
(b) Absolute probe $R^2$ (top) and normalized ablation effect (bottom) for DINOv2 targets. Negative $R^2$ is worse than the evaluation-mean baseline. Presence always probes raw DINOv2. Whiskers give 95\% image-bootstrap intervals. Seed 0, mid-noise, 400K steps, disjoint fit and evaluation images (\Cref{app:image_disjoint}).}
\vspace{-5pt}
\label{fig:inside_model}
\end{figure}

\textbf{Purified MAE yields features that are readable as DINOv2.} Although DINOv2 is never used in its training, a model aligned to purified MAE becomes more DINOv2-decodable. LAP-N raises the held-out probe score from $0.130$ in the unaligned model to $0.178$, closing about $63\%$ of the gap to explicit DINOv2 alignment at $0.207$ (\Cref{fig:inside_model}(a)). The probes are evaluated on images held out from probe fitting, not from diffusion training.

\textbf{Purification reduces coupling to velocity prediction.} Purifying the DINOv2 target keeps the teacher decodable from the hidden state, while linear velocity prediction from the aligned representation and the denoiser's reliance on the aligned subspace both decrease (\Cref{fig:inside_model}(b)). Negative velocity $\Rtwo$ reflects poor generalization of the linear probe, not the absence of reconstruction information. Generation quality is preserved while this coupling decreases, consistent with \Cref{sec:f2}, where increasing it worsened generation. Together, these results support choosing a target by the content it asks the model to match, rather than by its reconstruction or class predictability alone.

\subsection{Diagnostics Across Noise Levels}
\label{app:noise_diagnostics}

The supplementary noise, rank, and resampling checks use the dense token-split protocol. It collects approximately $10^5$ dense tokens from shuffled image batches and randomly splits them 80/20. Images can appear on both sides, so these scores measure held-out-token predictability. The conditional mean, full-layer effect, and subspace effects use separate sampling passes. A nonlinear REPA probe under this protocol gives $R^2=0.387$, compared with $0.343$ for its linear counterpart.

\Cref{tab:perbin} reports this protocol across high noise $[0.7,1]$, mid-noise $[0.3,0.7)$, and low noise $[0,0.3)$. Presence uses DINOv2 as the probe target, except for raw MAE, where it uses MAE. The cosine column also compares against DINOv2, except for raw MAE. MAE-residual projectors output MAE-dimensional features, so their DINOv2 cosine is undefined.

REPA's aligned-subspace use is lowest at high noise. The predictable-only target becomes more velocity-predictive toward low noise, whereas the purified targets show the opposite trend. These are observations under the stated probes. They do not establish that the aligned subspace is the only route through which target information can affect the denoiser.

\Needspace{14\baselineskip}
\begingroup\small\setlength{\LTcapwidth}{\textwidth}\renewcommand{\tablename}{\normalsize Table}
\begin{longtable}{@{}llrrrr@{}}
\caption{\normalsize\textbf{Diagnostics across noise levels.} ImageNet-1k/400K, seed 0, dense token-split protocol. Use is the rank-256 aligned-subspace effect as a percentage of the full-layer effect. Probe targets are specified in the text. A dash marks a cosine that is undefined because the projector outputs MAE-dimensional features.}
\label{tab:perbin}\\
\toprule
Model & Noise & Presence $R^2$ & Cosine & Velocity $R^2$ & Use (\%) \\
\midrule
\endfirsthead
\multicolumn{6}{l}{Table \thetable\ continued}\\
\toprule
Model & Noise & Presence $R^2$ & Cosine & Velocity $R^2$ & Use (\%) \\
\midrule
\endhead
\bottomrule
\endfoot

Unaligned & High & 0.177 & 0.005 & 0.582 & 8.5 \\*

 & Mid & 0.196 & 0.008 & 0.451 & 13.9 \\*

 & Low & 0.196 & 0.006 & 0.333 & 15.0 \\

\addlinespace[3pt]

Raw MAE & High & 0.255 & 0.947 & 0.329 & 3.8 \\*

 & Mid & 0.386 & 0.960 & 0.275 & 17.7 \\*

 & Low & 0.485 & 0.970 & 0.269 & 22.7 \\

\addlinespace[3pt]

REPA & High & 0.293 & 0.513 & 0.126 & 2.1 \\*

 & Mid & 0.343 & 0.630 & 0.164 & 10.9 \\*

 & Low & 0.386 & 0.714 & 0.164 & 13.4 \\

\addlinespace[3pt]

LAP-L (DINOv2) & High & 0.290 & 0.482 & 0.112 & 1.4 \\*

 & Mid & 0.342 & 0.606 & 0.089 & 6.5 \\*

 & Low & 0.385 & 0.692 & 0.060 & 8.0 \\

\addlinespace[3pt]

LAP-N (DINOv2) & High & 0.301 & 0.445 & 0.081 & 0.8 \\*

 & Mid & 0.361 & 0.565 & 0.079 & 4.5 \\*

 & Low & 0.403 & 0.644 & 0.066 & 5.5 \\

\addlinespace[3pt]

LAP-L (MAE) & High & 0.247 & --- & 0.190 & 3.0 \\*

 & Mid & 0.268 & --- & 0.149 & 15.5 \\*

 & Low & 0.283 & --- & 0.063 & 18.8 \\

\addlinespace[3pt]

LAP-N (MAE) & High & 0.258 & --- & 0.129 & 2.2 \\*

 & Mid & 0.290 & --- & 0.074 & 11.0 \\*

 & Low & 0.302 & --- & 0.056 & 13.4 \\

\addlinespace[3pt]

Predictable-only (DINOv2) & High & 0.204 & 0.208 & 0.163 & 8.8 \\*

 & Mid & 0.210 & 0.219 & 0.184 & 11.3 \\*

 & Low & 0.217 & 0.220 & 0.270 & 16.5 \\

\addlinespace[3pt]

REPA + prediction loss & High & 0.259 & 0.501 & 0.628 & 2.8 \\*

 & Mid & 0.297 & 0.607 & 0.541 & 16.0 \\*

 & Low & 0.324 & 0.679 & 0.452 & 25.6 \\

\end{longtable}
\endgroup

\subsection{Reference Subspaces, Resampling, and Rank}

The controls in this subsection use the dense token-split protocol. A random subspace gives a reference for how much disruption an intervention of the same rank causes. For REPA, the rank-256 random-subspace effect in \Cref{tab:random_subspace} is larger than the aligned-subspace effect in every noise bin. The unaligned model's untrained readout provides a second reference. It does not define the same directions as REPA's learned readout.

\begin{table}[H]
\centering\small
\setlength{\tabcolsep}{5pt}
\caption{\textbf{Random-subspace references.} Dense token-split protocol. Mean-fill use as a percentage of full-layer ablation. REPA brackets give the observed ranges over random subspace draws. These are not training-seed intervals.}
\label{tab:random_subspace}
\begin{tabular}{lccc}
\toprule
Checkpoint & High & Mid & Low \\
\midrule
REPA (400K) & 13.8 [13.6, 14.0] & 22.4 [21.0, 23.9] & 30.1 [28.8, 31.5] \\
Unaligned (400K) & 13.0 & 15.9 & 20.7 \\
\bottomrule
\end{tabular}
\end{table}

Resampling preserves the marginal distribution of the replaced component more closely than mean filling. \Cref{tab:resampling} compares the two interventions at mid-noise. REPA and the purified-target models have smaller normalized effects under resampling, but the exact ordering of all methods is not preserved. For example, raw MAE and the predictable-only model change order. The robust observation is the small resampling effect for REPA and purified targets, not equivalence of the two estimators.

\begin{table}[H]
\centering\small
\setlength{\tabcolsep}{5pt}
\caption{\textbf{Intervention sensitivity.} Dense token-split protocol, ImageNet-1k/400K, seed 0, mid-noise, rank 256. Each estimator uses its corresponding full-layer normalization.}
\label{tab:resampling}
\begin{tabular}{lrr}
\toprule
Model & Mean fill (\%) & Resampling (\%) \\
\midrule
Unaligned & 13.9 & 15.0 \\
Raw MAE & 17.7 & 6.5 \\
LAP-L (MAE) & 15.5 & 4.9 \\
LAP-N (MAE) & 11.0 & 1.4 \\
REPA & 10.9 & 2.0 \\
LAP-L (DINOv2) & 6.5 & 1.6 \\
LAP-N (DINOv2) & 4.5 & 1.1 \\
Predictable-only (DINOv2) & 11.3 & 10.6 \\
REPA + prediction loss & 16.0 & 14.6 \\
\bottomrule
\end{tabular}
\end{table}

The rank sweep in \Cref{tab:rank_sweep} tests whether the conclusion depends on removing 256 directions. Increasing rank generally removes more useful information. Comparisons should therefore be made at the same rank. REPA remains close to the unaligned reference at ranks 32 and 128, and below it at rank 256. The decreasing DINOv2 sequence holds at all three ranks. The MAE ordering does not: at rank 32, LAP-N (MAE) has the largest aligned-subspace effect among the MAE targets.

\begin{table}[H]
\centering\small
\setlength{\tabcolsep}{5pt}
\caption{\textbf{Rank sensitivity of mean-fill use.} Dense token-split protocol, mid-noise, 400K, seed 0. Values are percentages of full-layer ablation. Parentheses show the target-probe subspace control.}
\label{tab:rank_sweep}
\begin{tabular}{lccc}
\toprule
Model & $r{=}32$ & $r{=}128$ & $r{=}256$ \\
\midrule
Unaligned & 0.8 (0.2) & 4.4 (0.8) & 13.9 (2.8) \\
Raw MAE & 2.0 (1.0) & 8.6 (7.1) & 17.7 (16.4) \\
REPA & 0.8 (0.3) & 4.8 (1.8) & 10.9 (8.7) \\
LAP-L (DINOv2) & 0.6 (0.3) & 2.9 (1.6) & 6.5 (6.5) \\
LAP-N (DINOv2) & 0.2 (0.3) & 2.1 (1.9) & 4.5 (8.8) \\
LAP-L (MAE) & 3.0 (0.3) & 10.4 (1.7) & 15.5 (5.6) \\
LAP-N (MAE) & 5.8 (0.5) & 8.7 (1.9) & 11.0 (8.7) \\
Predictable-only (DINOv2) & 1.6 (0.1) & 4.5 (0.5) & 11.3 (2.2) \\
REPA + prediction loss & 1.7 (1.2) & 4.3 (3.7) & 16.0 (7.3) \\
\bottomrule
\end{tabular}
\end{table}

\subsection{Component Readout and Easy-to-Match Content}

The component diagnostic in \Cref{fig:content_diagnostics}(b) uses the same 1,024 fit and 256 evaluation images as the other main-text probes. We fit the affine clean-latent-to-DINOv2 decomposition within batches of 64 images. No decomposition batch crosses the split boundary. We select 48 tokens per image for the component probes, giving 49,152 fit and 12,288 evaluation tokens. Readout cosines use all 256 tokens of each evaluation image. Thus, the targets remain batch-defined components, while the probes are tested on disjoint images.

The image-predictable component has higher presence than the residual ($0.256$ vs.\ $0.188$), while the projector has higher cosine similarity to the residual ($0.581$ vs.\ $0.313$). The paired differences have 95\% image-bootstrap intervals of $[0.047,0.086]$ for predictable-minus-residual presence and $[0.254,0.283]$ for residual-minus-predictable cosine. These measurements distinguish availability from readout matching. They do not prove that the projector causally selects semantic content, because component geometry and scale can also affect cosine.

For comparison, a random-token split of the same component collection gives predictable/residual presence $0.443/0.263$. The dense token-split component protocol samples 60K tokens across 20 batches and gives approximately $0.40/0.25$, with readout cosines $0.31/0.58$. Both protocols retain the ordering, but token-held-out scores are higher. The development-scale token-split residual/predictable cosine pair is $0.52/0.42$.

\subsection{Constant-Readout Evaluation}
\label{app:constant_baseline}
The constant-readout baseline $C_0$ of \Cref{sec:baseline} measures how much alignment cosine a single fixed direction obtains without image-specific computation. We estimate it as the norm of the sample mean of unit-normalized tokens. Features come from $2{,}048$ training images drawn with a fixed seed, processed in consecutive batches of $64$ so that the batchwise LAP-L target is formed exactly as in training, with one fixed VAE posterior sample per image shared by all targets. \Cref{tab:constant_baseline_full} reports estimates for dev-200 and ImageNet-1k with the corresponding purifiers. Raw MAE has a baseline of $0.933$ on dev-200 and $0.929$ on ImageNet-1k. Its logged training cosine of $0.957$ is an illustrative comparison from different samples (\Cref{app:var_usable}). For predictable-only DINOv2, the baseline is $0.802$ on dev-200 and $0.733$ on ImageNet-1k. Both permit high alignment scores without reading the hidden state. Under LAP-N, the smallest token norms across these samples are approximately $24.3$ for MAE, $20.6$ for DINOv2, and $8.9$ for CLIP.

\textbf{Per-position constant readouts.} The shared-direction baseline is the strictly input-free one. A readout that is allowed one fixed direction per token position still reads nothing about the image, but it can exploit position-dependent structure in the target; its best expected cosine is $C_0^{\mathrm{pos}}=\frac1N\sum_i\|\mathbb{E}[\hat y^{(i)}]\|$, the position-wise analogue of \Cref{eq:constant_baseline}, and $C_0^{\mathrm{pos}}\ge C_0$ by the triangle inequality. \Cref{tab:constant_baseline_full} reports both. For the raw targets the two nearly coincide for MAE and CLIP, whereas raw DINOv2 tokens carry a position-dependent mean that a shared direction misses ($0.16$ against $0.31$ on ImageNet-1k). The evaluated purified targets retain position-dependent mean structure: a per-position readout still reaches about $0.33$ on LAP-N MAE, $0.21$ on LAP-N DINOv2, and $0.10$ on LAP-N CLIP. For the shared token-wise projector, positional information is available through the hidden state. The main text uses $C_0$ to isolate a shared direction, while the per-position baseline quantifies what a readout with access to position alone can achieve without image content. The tested target transformations do not explicitly subtract a position-specific mean across images. The iREPA-style operation standardizes each image across its positions (\Cref{app:spatial_interaction}); this also does not generally force the position-wise directional mean to vanish. Subtracting a frozen per-position mean is a further member of the purification family, with the token position as the predictor's only input; \Cref{app:lapp} evaluates it as LAP-P.

\begin{table}[H]
\centering\small
\setlength{\tabcolsep}{5pt}
\caption{\textbf{Constant-readout baselines by dataset, encoder, and target.} Empirical estimates from $2{,}048$ images per dataset. $C_0$ is the best expected cosine of one direction shared by all tokens (\Cref{eq:constant_baseline}); $C_0^{\mathrm{pos}}=\frac1N\sum_i\|\mathbb{E}[\hat y^{(i)}]\|\ge C_0$ is the best expected cosine of a readout with one fixed direction per token position, which uses the position but not the image. LAP-N uses the purifier trained on the same dataset ($k=5$ for MAE, AIMv2, and DINOv3, $k=3$ for DINOv2 and CLIP); CLIP $k=5$ on dev-200 gives $0.062$ and $0.094$. The predictable-only DINOv2 control, the batchwise affine prediction of DINOv2 from the clean latent, has $C_0=0.802$ and $0.733$ and $C_0^{\mathrm{pos}}=0.804$ and $0.735$ on dev-200 and ImageNet-1k; LAP is not applied to it.}
\label{tab:constant_baseline_full}
\setlength{\tabcolsep}{2.5pt}
\begin{tabular}{llrrrrrr}
\toprule
 & & \multicolumn{3}{c}{$C_0$ (shared direction)} & \multicolumn{3}{c}{$C_0^{\mathrm{pos}}$ (per position)} \\
\cmidrule(lr){3-5}\cmidrule(lr){6-8}
Data & Encoder & Raw & LAP-L & LAP-N & Raw & LAP-L & LAP-N \\
\midrule
\multirow{5}{*}{Dev-200} & MAE & 0.933 & 0.020 & 0.011 & 0.945 & 0.437 & 0.329 \\
 & CLIP & 0.607 & 0.088 & 0.067 & 0.620 & 0.169 & 0.110 \\
 & DINOv2 & 0.221 & 0.007 & 0.011 & 0.347 & 0.272 & 0.207 \\
 & AIMv2 & 0.663 & 0.041 & 0.036 & 0.701 & 0.221 & 0.130 \\
 & DINOv3 & 0.525 & 0.018 & 0.017 & 0.587 & 0.310 & 0.168 \\
\midrule
\multirow{5}{*}{ImageNet-1k} & MAE & 0.929 & 0.017 & 0.010 & 0.941 & 0.420 & 0.326 \\
 & CLIP & 0.585 & 0.082 & 0.066 & 0.593 & 0.140 & 0.098 \\
 & DINOv2 & 0.162 & 0.005 & 0.011 & 0.306 & 0.263 & 0.205 \\
 & AIMv2 & 0.663 & 0.042 & 0.035 & 0.695 & 0.188 & 0.110 \\
 & DINOv3 & 0.509 & 0.012 & 0.019 & 0.568 & 0.290 & 0.164 \\
\bottomrule
\end{tabular}
\end{table}

The predictable-only DINOv2 target is also easy to match. Its two full-scale runs start near cosine $0.82/0.81$ at step 500 and finish near $0.83/0.85$. Raw DINOv2 starts near $0.32$ and finishes near $0.62/0.63$. Training-log cosines and held-out diagnostic cosines use different samples and should not be equated. In particular, the DINOv2 cosine in \Cref{tab:perbin} for the predictable-only model is measured against raw DINOv2, rather than that model's own training target.

\subsection{Position-Dependent Purification}
\label{app:lapp}

The per-position baseline above identifies shared target structure that LAP-L and LAP-N can retain: a mean token that depends on position $i$ but not on the image. A position-dependent predictor subtracts $m_i=\mathbb{E}[z^{(i)}]$, estimated once on $8{,}192$ training images and then frozen. We call this LAP-P. The population position means average to the global mean token. For MAE, LAP-P gives $C_0=0.019$ and $C_0^{\mathrm{pos}}=0.027$, compared with $0.933$ and $0.945$ for the raw target (\Cref{tab:lapp}). Position-dependent centering can also be applied after LAP-N by subtracting the per-position mean of its residual (LAP-N$+$P).

\begin{table}[H]
\centering\small
\setlength{\tabcolsep}{5pt}
\caption{\textbf{Position-dependent purification at development scale.} $C_0$ and $C_0^{\mathrm{pos}}$ from $2{,}048$ dev-200 images (\Cref{app:constant_baseline}) and dev-200/50K FID-DINOv2 for training seeds 0 and 1 and their mean. The DINOv2 purifier uses $k{=}3$. Shaded rows use LAP; bold marks the lowest FID per encoder and column.}
\label{tab:lapp}
\begin{tabular}{llrrrrr}
\toprule
Encoder & Target & $C_0$ & $C_0^{\mathrm{pos}}$ & Seed 0 & Seed 1 & Mean \\
\midrule
\multirow{5}{*}{MAE} & Raw & 0.933 & 0.945 & 1641.4 & 1643.2 & 1642.3 \\
 & \ours LAP-L & \ours 0.020 & \ours 0.437 & \ours 1504.3 & \ours 1512.3 & \ours 1508.3 \\
 & \ours LAP-N & \ours 0.011 & \ours 0.329 & \ours 1486.2 & \ours 1482.6 & \ours 1484.4 \\
 & \ours LAP-P & \ours 0.019 & \ours 0.027 & \ours 1495.9 & \ours 1502.8 & \ours 1499.4 \\
 & \ours LAP-N$+$P & \ours 0.012 & \ours 0.025 & \ours \textbf{1471.3} & \ours \textbf{1436.4} & \ours \textbf{1453.8} \\
\midrule
\multirow{5}{*}{DINOv2} & Raw & 0.221 & 0.347 & 1551.2 & 1518.6 & 1534.9 \\
 & \ours LAP-L & \ours 0.007 & \ours 0.272 & \ours 1546.0 & \ours \textbf{1487.8} & \ours 1516.9 \\
 & \ours LAP-N & \ours 0.011 & \ours 0.207 & \ours \textbf{1508.4} & \ours 1536.2 & \ours 1522.3 \\
 & \ours LAP-P & \ours 0.010 & \ours 0.019 & \ours 1518.9 & \ours 1507.3 & \ours 1513.1 \\
 & \ours LAP-N$+$P & \ours 0.010 & \ours 0.020 & \ours 1512.1 & \ours 1504.0 & \ours \textbf{1508.1} \\
\bottomrule
\end{tabular}
\end{table}

\textbf{Development scale.} On MAE, LAP-P alone reaches $1499.4$, better than LAP-L ($1508.3$) and close to LAP-N ($1484.4$), although its predictor uses no image content; LAP-N$+$P is the best development-scale MAE target ($1453.8$, $30.6$ below LAP-N). On DINOv2 the four purified targets lie within $15$ points of one another.

\begin{table}[H]
\centering\small
\setlength{\tabcolsep}{6pt}
\caption{\textbf{Position-dependent purification at full scale.} MAE ViT-L teacher, ImageNet-1k/400K FID-50K for training seeds 0 and 1 and their mean; full-scale LAP-P runs were made for MAE only. Shaded rows use LAP; bold marks the lowest FID per column.}
\label{tab:lapp_full}
\begin{tabular}{lrrr}
\toprule
Target & Seed 0 & Seed 1 & Mean \\
\midrule
Raw & 53.68 & 52.41 & 53.05 \\
\ours LAP-L & \ours 44.81 & \ours 45.48 & \ours 45.15 \\
\ours LAP-N & \ours 43.06 & \ours \textbf{41.53} & \ours 42.30 \\
\ours LAP-P & \ours 44.02 & \ours 44.62 & \ours 44.32 \\
\ours LAP-N$+$P & \ours \textbf{42.33} & \ours 41.80 & \ours \textbf{42.07} \\
\bottomrule
\end{tabular}
\end{table}

\textbf{Full scale.} At full scale, the additional benefit of position-dependent centering on top of LAP-N is not resolved by the two seeds (\Cref{tab:lapp_full}). LAP-P reaches $44.32$ FID-50K, $0.83$ below LAP-L but $2.0$ above LAP-N. LAP-N$+$P reaches $42.07$ against $42.30$ for LAP-N, a difference within seed variation that improves only one of the two matched seeds. Thus, position-dependent centering improves on LAP-L in this comparison, while its additional benefit on top of LAP-N is not resolved by these two seeds. We retain LAP-P as an ablation. Per-image spatial standardization in iREPA instead centers each image across its own positions, so it does not generally remove a position-dependent mean across images (\Cref{app:spatial_interaction}).

\subsection{Clean-Latent Prediction Across Noise Levels}

The raw-target utility scores in \Cref{tab:ladder} use 30 shuffled batches of 32 images (245,760 tokens). Probes fit the first 80\% of tokens and evaluate on the remaining 20\%, without a second token shuffle. This boundary falls between images.

To ask what target features add beyond the noisy input, we compare two token-wise clean-latent probes. Here $x_{t,p}^{(i)}\in\mathbb R^{16}$ is the noisy latent patch at position $i$, $X_p^{(i)}\in\mathbb R^{16}$ is its clean counterpart, and $[x_{t,p}^{(i)},y^{(i)}]$ concatenates that patch with its corresponding target feature along the feature dimension:
\begin{equation}
\Delta\Rtwo=\Rtwo([x_{t,p}^{(i)},y^{(i)}]\!\to\!X_p^{(i)})-\Rtwo(x_{t,p}^{(i)}\!\to\!X_p^{(i)}).
\end{equation}
Each probe shares its coefficients across tokens and is fit separately within each noise bin. Neither the exact time $t$ nor the token position is supplied as an additional feature.
\Cref{tab:incremental_utility} shows that MAE adds more linear reconstruction information than DINOv2 or CLIP, including at high noise. This does not translate into the largest alignment benefit. At low noise, the hidden states of both unaligned and REPA models already predict clean-latent patches with $R^2\approx0.95$. Incremental probes use 16 batches of 32 images and the same contiguous 80/20 token split. This boundary can split one image. Separate posterior and noise draws give slightly different noisy-input baselines for each encoder. Small negative increments can arise from estimation and regularization.

\begin{table}[H]
\centering\small
\setlength{\tabcolsep}{5pt}
\caption{\textbf{Clean-latent prediction.} Incremental utility uses dev-200 data. Hidden-state scores use the dense token-split protocol on 400K ImageNet-1k checkpoints. The baseline row shows the range across the three encoder evaluations. High, mid, and low denote the noise bins defined above.}
\label{tab:incremental_utility}
\begin{tabular}{llccc}
\toprule
 & & High & Mid & Low \\
\midrule
Base range & $\Rtwo(x_{t,p}^{(i)}\!\to\!X_p^{(i)})$ & 0.037--0.051 & 0.472--0.501 & 0.926--0.934 \\
\midrule
\multirow{3}{*}{$\Delta\Rtwo$} & MAE & 0.655 & 0.266 & 0.013 \\
 & DINOv2 & 0.268 & 0.052 & $-0.001$ \\
 & CLIP & 0.114 & 0.026 & $-0.001$ \\
\midrule
\multirow{2}{*}{$\Rtwo(h_t^{(i)}\!\to\!X_p^{(i)})$} & Unaligned & 0.264 & 0.676 & 0.949 \\
 & REPA & 0.248 & 0.662 & 0.946 \\
\bottomrule
\end{tabular}
\end{table}

\subsection{Seed Coverage and Uncertainty}

Core generation comparisons use two independent training seeds. Slash-separated pairs give seeds 0 and 1. A lone generation value is a single evaluated run unless identified as a mean. Diagnostic coverage is specified separately. The range between two seeds describes the observed variation and is not a confidence interval.

The dev-200 REPA pair, $1551.2/1518.6$, gives a sample standard deviation $s=|1551.2-1518.6|/\sqrt{2}\approx23.0$. Adding the repeat at $1553.3$ gives $s\approx19.5$. Even this one additional run changes the estimate. Variation can also depend on the configuration.

Under an independent, equal-variance approximation, the standard error of the difference between averages of $n$ and $m$ runs is $s\sqrt{1/n+1/m}$. With only two runs per configuration, this approximation does not justify precise significance claims or equivalence tests. We therefore report individual values, compare consistent directions across seeds and metrics, and avoid ranking closely performing methods.

Evaluating another metric on the same generated images checks sensitivity to the feature space. Drawing new samples checks sampling variation. Evaluating intermediate checkpoints shows the training trajectory. None of these replaces independent training seeds. The Inception-space check and complete trajectories are in \Cref{app:adm}.

\FloatBarrier
\section{Controls for Predictive Supervision and Target Content}
\label{app:controls}

The controls address two possible confounds: the architecture used for auxiliary prediction, and the geometry or instability introduced by decomposing a target. All development values in this appendix are FID-DINOv2. Standard FID is named explicitly.

\subsection{Auxiliary Prediction Controls}

The velocity-prediction head $r_\omega$ is a timestep-conditioned MLP with one hidden layer of width 256 and SiLU activation. It receives the aligned representation $a_t$. The loss is uniformly weighted over noise times, and gradients pass through the head into the shared hidden state. Setting $\lambda_s=0$ recovers REPA.

We also apply the same loss to a fixed random projection of $h$ and to a separately learned projection. These controls test whether the degradation requires supervision of the aligned representation itself. At dev-200/100K with $\lambda_s=2$, the seed-0 scores are $1284.5$ and $1277.4$, compared with $1242.0$ for supervision through the aligned representation and $1157.3$ for REPA. Both controls reduce aligned-subspace use from $8.3\%$ to about $4.1$--$4.3\%$, while leaving target presence and aligned-readout sufficiency near the REPA reference.

Thus, degradation is not unique to the aligned projection. The results are consistent with interference through shared parameters, but they do not isolate a particular gradient-level mechanism. A separate MAE sweep at dev-200/50K gives $1641.4$, $1605.4$, and $1649.7$ for $\lambda_s=0,0.5,2$. This single-run check has no monotonic benefit.

The main negative result replicates across two seeds. At dev-200/100K, auxiliary prediction gives $1242.0/1252.3$ against REPA's $1157.3/1161.2$. At full scale, it gives $879.5/877.1$ against $825.1/821.1$. Standard FID-50K likewise increases from $41.75$ to $46.79$ at seed means. The full sweep is in \Cref{tab:f2}, and the standard metrics of this control are in \Cref{tab:mech_full}.

\subsection{The Affine Target Decomposition}

Let $X_p\in\mathbb{R}^{M\times16}$ contain the clean-latent patches from one batch, with $M=BN$ tokens. Append a column of ones to obtain $\widetilde X_p$. For encoder tokens $Z\in\mathbb{R}^{M\times D}$, ordinary least squares defines
\begin{equation}
\widehat W=\widetilde X_p^+Z,\qquad
Z_{\mathrm{pred}}=\widetilde X_p\widehat W,\qquad
Z_{\mathrm{res}}=Z-Z_{\mathrm{pred}}.
\end{equation}
Aligning to $Z_{\mathrm{res}}$ defines LAP-L. It is the affine variant of the residual-target construction in \Cref{sec:purifier}, with coefficients fitted to the current batch rather than frozen before training. The fitted component has rank at most 17 across the batch tokens. With an intercept, the residual has zero batch mean and is orthogonal to the design matrix, up to numerical precision. These in-batch properties do not imply independence from local appearance or semantic purity.

To quantify the split, we use centered explained variance,
\[
\rho_{\mathrm{lin}}
=1-\frac{\|Z_{\mathrm{res}}\|_F^2}{\|Z-\mathbf{1}\bar z^\top\|_F^2},
\]
where $\bar z$ is the mean token in the fitted batch. The recorded per-batch residual norms give approximately $2.7\%$ on dev-200 and $2.1\%$ at full scale. The batch mean is also removed, but does not count toward this centered score. A dataset-level fit gives $1.9\%$ and $1.3\%$, respectively, measured on its fitting collection rather than held-out tokens.

Refitting the coefficients each batch also changes the target. Across batch recompositions, per-token target-direction cosine is about $0.85$ on dev-200 and $0.76$ at full scale. The per-batch and dataset-level coefficient matrices have cosine about $0.90$ and $0.85$. These checks quantify the variation rather than establish that it is negligible.

The shared mean token carries $4.9\%$ of uncentered target energy on dev-200 and $2.6\%$ at full scale. These energy fractions use a different denominator from centered explained variance. The distinction separates the contribution of the affine intercept from explained centered variation.

For \Cref{fig:motivating}(a), the affine fit uses all 64 images in the first batch of the full-scale purifier-validation permutation. Two examples are selected for legibility. After the split, each component is centered within each image only for display. Maps show cosine to patch $(7,7)$ on the zero-indexed token grid with a shared $[-1,1]$ scale. These maps are illustrative, and the FID values summarize separate dataset-level generation evaluations.

\subsection{Separating Rank, Instability, and Predictable Content}

\begin{figure}[H]
\centering
\setlength{\parskip}{0pt}
\setlength{\abovecaptionskip}{5pt}
\includegraphics[width=\linewidth]{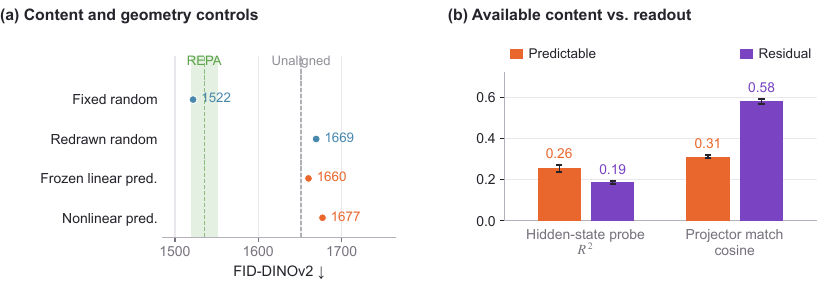}
\vspace{-12pt}
\caption{\textbf{Reconstruction content is easy to recover but ineffective for alignment}.
(a) Dev-200/50K controls (single runs). Baseline lines and bands show two-seed means and ranges.
(b) Predictable content is more decodable, while the projector matches the residual more closely (95\% image-bootstrap intervals). One 400K REPA checkpoint, mid-noise, disjoint fit/evaluation images (\Cref{app:probes,app:controls}).}
\vspace{-5pt}
\label{fig:content_diagnostics}
\end{figure}

\Cref{fig:content_diagnostics}(a) summarizes four single-run development controls. A fixed random rank-17 projection of DINOv2 gives $1521.7$, close to the REPA seed range. A rank-matched projection can therefore remain effective. Because it is a projection of image features, this control retains image-dependent information.

Redrawing the projection each batch gives $1669.4$, close to the unaligned baseline. Target instability can therefore remove the benefit. However, the frozen dataset-level reconstruction predictor gives $1660.2$, and the nonlinear predicted component $d_\phi(X_p)$ gives $1677.0$. Both reconstruction-based targets remain ineffective without per-batch coefficient fitting. These controls weaken explanations based solely on rank or refitting noise.

At full scale, the redrawn random control gives $937.9$ in one run, close to the unaligned $945.7/947.4$ pair. Its 100K--400K trajectory is $1273.8/1099.2/995.3/937.9$. Unlike the predictable-only target, it does not show the same worse-than-unaligned endpoint.

\subsection{Where the Predictable-Only Target Becomes Harmful}

At development scale, the predictable-only target loses the REPA benefit but remains near the unaligned model. At full scale, both predictable-only runs ($1000.6/995.2$) are worse than both unaligned runs ($945.7/947.4$). The mean proxy gap is about $51$ points. Standard FID-50K agrees: $56.71$ versus $53.92$. The predictable-only target also has the lowest precision and recall in \Cref{tab:mech_full}.

The guided check gives the same direction ($29.92$ vs.\ $27.61$, seed 0). The claim of harm below the unaligned baseline is therefore scoped to the full-scale evaluations. The development controls establish loss of alignment benefit, not harm in every training setting.

\FloatBarrier
\addtocontents{toc}{\protect\newpage}
\section{Purifier Training, Validation, and Ablations}
\label{app:purifier}

LAP-L uses the batchwise affine fit in \Cref{app:controls}. This appendix describes LAP-N's frozen nonlinear predictor, its validation and selection, and its interaction with other alignment transformations. Both variants subtract local predictions, but their fitting protocols and predictor inputs differ.

\subsection{Architecture and Validation Criteria}

The predictor operates on the $16\times16$ patch grid. It consists of a $k\times k$ convolution from 16 to 512 channels, GELU, a $1\times1$ convolution from 512 to 512 channels, GELU, and a final $1\times1$ convolution to the encoder dimension $D$. We fit it with mean-squared error for 20K steps using batch 256, AdamW at learning rate $10^{-3}$, and a cosine learning-rate schedule. We use weight decay $10^{-4}$ and 500 warmup steps. A seed-0 permutation reserves 4,096 images for validation and fits the predictor on the remaining images. Fitting takes approximately two A100-hours per predictor. Locality refers to the latent-token grid, not a strict receptive-field bound in image pixels.

On held-out data, we define the explained-variance score
\begin{equation}
\rho=1-\frac{\mathbb{E}\|z-d_\phi(X_p)\|^2}
{\mathbb{E}\|z-\bar z\|^2}.
\end{equation}
This score measures prediction error relative to a constant mean-token predictor. It is an operational measure of removable local predictability, not a semantic-content estimate.

We assess class-information retention with linear softmax probes fit on 21,000 images and evaluated on the 4,096 images held out from purifier fitting. Using the same seed-0 permutation, probe fitting takes indices $8192{:}29192$ and evaluation takes $0{:}4096$; their image IDs are disjoint. We test mean-pooled features and token features, sampling eight positions per image with replacement and averaging their logits at evaluation. Features are centered by their own fitting-set mean and scaled by the raw target's fitting-set per-dimension standard deviation plus $10^{-6}$. Probes use full-batch Adam with learning rate $10^{-2}$ and weight decay $10^{-4}$ for at most 2,500 iterations; an improvement below $10^{-4}$ in cross-entropy between 500-iteration checks triggers early stopping. The linear-residual reference is formed in extraction batches of 128 images. These batches can straddle probe subsets, so its affine coefficients are not restricted to the probe-fit images; the class-probe fitting and evaluation image IDs remain disjoint.

The implemented selection rule compares each residual to the linear-residual target and to the component removed by the predictor. Let $a_{\mathrm{lin}}$ be the pooled linear-residual accuracy and $n_v=4096$. The tolerance is $\delta=2\sqrt{a_{\mathrm{lin}}(1-a_{\mathrm{lin}})/n_v}$. A receptive field is feasible when, for both pooled and token probes, its residual accuracy is at least the corresponding linear-residual accuracy minus $\delta$, and exceeds the predicted-component accuracy by more than $\delta$. For MAE, AIMv2, and DINOv3, we choose the largest feasible tested field. For DINOv2 and CLIP, the base comparisons use a fixed $k=3$ rather than the largest feasible field. Larger fields are evaluated separately as feasibility checks. This tolerance is a selection heuristic, not a significance test, and the probes do not certify complete semantic preservation.

We separately monitor the training-to-validation gap in $\rho$ to check prediction generalization. \Cref{tab:purifier_validation} reports the pooled accuracies and these gaps.

\begin{table}[H]
\centering\small
\setlength{\tabcolsep}{5pt}
\caption{\textbf{Purifier validation on dev-200.} Class accuracy is held-out pooled top-1 accuracy over 200 classes. Chance accuracy is $0.005$. The last column is the training-to-validation gap in $\rho$. Bold marks the higher classification accuracy between the raw and purified targets in each row.}
\label{tab:purifier_validation}
\begin{tabular}{llrrrr}
\toprule
Encoder & $k$ & $\rho$ & Raw accuracy & Purified accuracy & $\rho$ gap \\
\midrule
MAE ViT-L & 1 & 0.178 & 0.661 & \textbf{0.693} & $\leq0.003$ \\
 & 3 & 0.314 & 0.661 & \textbf{0.688} & $\leq0.003$ \\
 & 5 & 0.352 & 0.661 & \textbf{0.687} & $\leq0.003$ \\
\midrule
DINOv2 ViT-B & 3 & 0.125 & \textbf{0.798} & 0.794 & $\leq0.003$ \\
\midrule
CLIP ViT-L & 3 & 0.074 & 0.787 & \textbf{0.791} & $\leq0.014$ \\
 & 5 & 0.084 & 0.787 & \textbf{0.790} & $\leq0.014$ \\
\bottomrule
\end{tabular}
\end{table}

All listed predictors pass the class-probe feasibility rule, and their training-to-validation gaps in $\rho$ are below $0.014$. At full ImageNet-1k scale, $\rho$ is $0.344$ for MAE at $k=5$, $0.100$ for DINOv2 at $k=3$, and $0.063/0.079$ for CLIP at $k=3/5$. MAE has the largest locally predictable fraction at both scales.

\Cref{tab:purifier_feasibility} gives the residual-versus-prediction accuracies for all evaluated candidates. The pooled/token linear-residual accuracies and tolerance $\delta$ are respectively MAE $0.6848/0.5420$, $0.01452$; DINOv2 $0.7942/0.7986$, $0.01263$; CLIP $0.7886/0.7683$, $0.01276$ on dev-200, and AIMv2 $0.7830/0.6663$, $0.01288$; DINOv3 $0.7139/0.7048$, $0.01412$ on ImageNet-1k. All listed candidates pass both checks. AIMv2 and DINOv3 use $k=5$, the largest of their tested $\{1,3,5\}$ fields, selected directly on ImageNet-1k; their held-out $\rho$ values are $0.1265$ and $0.1755$, and training-minus-validation gaps are $-0.00074$ and $-0.000001$. These 1,000-class accuracies should not be compared directly with the 200-class development accuracies.

\begin{table}[H]
\centering\small
\setlength{\tabcolsep}{5pt}
\caption{\textbf{Purifier feasibility checks.} Each accuracy pair is residual/predicted-component top-1 accuracy on the same 4,096 held-out images. ``Pass'' requires both inequalities in both pooled and token probes, using the linear references and tolerances above. DINOv2 $k=5,7$ and CLIP $k=5$ are feasibility controls; the base comparisons use fixed $k=3$ checkpoints.}
\label{tab:purifier_feasibility}
\begin{tabular}{llrccc}
\toprule
Dataset & Encoder & $k$ & Pooled res./pred. & Token res./pred. & Both \\
\midrule
dev-200 & MAE-L & 1 & $0.6926/0.1492$ & $0.5464/0.0603$ & Pass \\
 &  & 3 & $0.6882/0.2053$ & $0.5449/0.0901$ & Pass \\
 &  & 5 & $0.6868/0.2170$ & $0.5452/0.0996$ & Pass \\
\midrule
dev-200 & DINOv2-B & 3 & $0.7942/0.2703$ & $0.7986/0.1311$ & Pass \\
 &  & 5 & $0.7937/0.2795$ & $0.7981/0.1533$ & Pass \\
 &  & 7 & $0.7935/0.2856$ & $0.7976/0.1672$ & Pass \\
\midrule
dev-200 & CLIP-L & 3 & $0.7908/0.2366$ & $0.7698/0.1101$ & Pass \\
 &  & 5 & $0.7898/0.2410$ & $0.7720/0.1133$ & Pass \\
\midrule
ImageNet-1k & AIMv2-L & 1 & $0.7810/0.0657$ & $0.6582/0.0212$ & Pass \\
 &  & 3 & $0.7805/0.1099$ & $0.6738/0.0439$ & Pass \\
 &  & 5 & $0.7812/0.1165$ & $0.6748/0.0581$ & Pass \\
\midrule
ImageNet-1k & DINOv3-B & 1 & $0.7175/0.0718$ & $0.7124/0.0273$ & Pass \\
 &  & 3 & $0.7168/0.1272$ & $0.7061/0.0669$ & Pass \\
 &  & 5 & $0.7166/0.1306$ & $0.7058/0.0796$ & Pass \\
\bottomrule
\end{tabular}
\end{table}

\subsection{Visualizing the MAE Target Transformation}
\label{app:target_geometry}

\begin{figure}[H]
\centering
\setlength{\parskip}{0pt}
\setlength{\abovecaptionskip}{5pt}
\includegraphics[width=\linewidth]{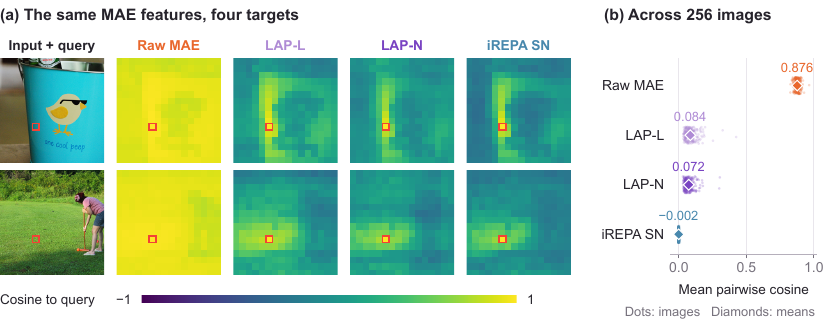}
\vspace{-15pt}
\caption{\textbf{How purification changes the MAE target.}
(a) Cosine similarity to the fixed red-boxed query patch for the raw features, LAP-L and LAP-N, and the iREPA-style spatially normalized target (SN), on a shared scale.
(b) Mean off-diagonal patch cosine over 256 held-out images. Both LAP-L and LAP-N substantially reduce MAE's high common similarity, with a lower mean for LAP-N; per-image spatial normalization gives a near-zero value in this evaluation.}
\label{fig:lap_targets}
\vspace{-5pt}
\end{figure}

\Cref{fig:lap_targets} uses the frozen full-scale MAE encoder and its $k=5$ purifier. We take the first $256$ images from the purifier's held-out images, without selecting images by feature quality. The two displayed examples are chosen for image legibility and because the two LAP targets visibly differ on them; their query is at row 10, column 5 of the $16\times16$ token grid, using zero-based indices, one of five fixed query positions recorded for every image. All aggregate measurements include the full $256$-image subset and do not depend on this choice.

We compare raw tokens $z$ with two LAP targets, each formed as in training: LAP-L, the least-squares affine residual of $z$ on the patchified clean latent within consecutive batches of $64$ images; and LAP-N, $z-d_\phi(X_p)$. LAP-L and LAP-N use a sampled clean VAE latent. Each heatmap shows cosine similarity between the query token and every patch token. We use a common $[-1,1]$ color scale and do not subtract per-image means, since doing so would conceal the shared component being examined.

For each image, we also summarize similarity over all distinct token pairs:
\[
 s(y)=\frac{1}{N(N-1)}\sum_{i\ne j}
 \frac{\langle y_i,y_j\rangle}{\|y_i\|\,\|y_j\|},\qquad N=256.
\]
The image-averaged scores are $0.876$ for raw MAE, $0.084$ for LAP-L, and $0.072$ for LAP-N. Both local predictors substantially reduce common patch similarity, with a further reduction for the nonlinear predictor. This statistic describes target geometry and is not a semantic-quality or generation metric. The iREPA-style spatially normalized target, which standardizes each image across its own tokens per channel (\Cref{app:spatial_interaction}), scores $-0.002$. Spatial centering makes the unnormalized token mean zero, but does not in general make the mean of unit-normalized tokens zero. Thus, the near-zero pairwise cosine is an empirical observation, and this statistic alone does not identify which target transformation best supports generation. The two operations differ in what they remove, a per-image mean and scale versus locally predictable content, and their composition is evaluated in \Cref{tab:compose}.

As a separate check, we fit a constant mean-token readout on a disjoint $512$-image subset of the purifier's training split. Its evaluation cosine is $0.929$ against raw MAE and approximately zero against LAP-L ($-0.002$) and LAP-N ($-0.005$). These values corroborate the loss of a strong shared direction after purification. This evaluation uses its own image subset and should not be conflated with the trained projector's cosine or with the constant-readout protocol in \Cref{app:constant_baseline}. All measurements use frozen checkpoints.

\begin{figure}[H]
\centering
\setlength{\parskip}{0pt}
\setlength{\abovecaptionskip}{5pt}
\includegraphics[width=\linewidth]{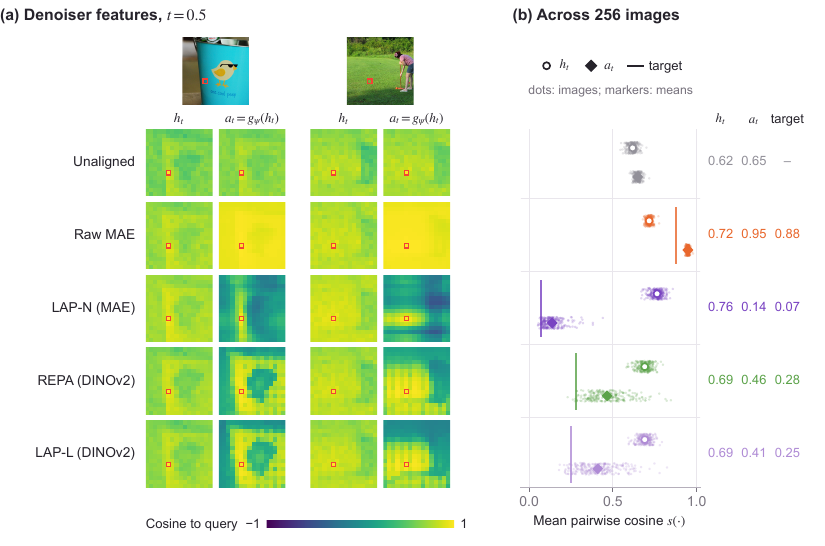}
\vspace{-12pt}
\caption{\textbf{The denoiser's features under raw and purified targets.} (a) Cosine to the red-boxed query patch for the block-8 hidden state $h_t$ and the projector output $a_t$ at $t=0.5$, two held-out images, five 400K seed-0 models. (b) Mean pairwise cosine of $h_t$ (circles), $a_t$ (diamonds), and the model's own target (bars) over 256 images. Raw MAE has the highest projector-output similarity.}
\label{fig:denoiser_maps}
\vspace{-5pt}
\end{figure}

\textbf{Where the shared direction lives in the trained model.} \Cref{fig:denoiser_maps} applies the same query-patch visualization to the denoiser itself: the block-8 hidden state $h_t$ and the projector output $a_t=g_\psi(h_t)$ at $t=0.5$, one shared noise draw, for the 400K seed-0 checkpoints of the unaligned model, raw MAE, LAP-N (MAE), REPA (DINOv2), and LAP-L (DINOv2), on the same 256 held-out images. \Cref{tab:denoiser_maps} gives the image-averaged pairwise cosine $s(\cdot)$ of $h_t$, of $a_t$, and of each model's own training target, together with the cosine between $a_t$ and the projector's bias vector. The hidden states of all five models have similar statistics ($0.62$--$0.76$), including the unaligned model. Raw MAE produces high projector-output similarity ($0.95$) and a cosine of $0.74$ between $a_t$ and the projector's bias. After LAP-N, the bias cosine is $-0.004$ and projector-output similarity is much lower ($0.14$). These observations are consistent with a substantial shared component in the raw-MAE readout, as suggested by \Cref{sec:baseline}, but do not establish that it is a constant function or that the bias causes the generation gap.

\begin{table}[H]
\centering\small
\setlength{\tabcolsep}{5pt}
\caption{\textbf{Patch similarity inside the trained models.} Image-averaged mean pairwise cosine $s(\cdot)$ over 256 held-out images at $t=0.5$ (block 8, 400K, seed 0), and the cosine between the projector output and the projector's bias vector.}
\label{tab:denoiser_maps}
\begin{tabular}{lrrrr}
\toprule
Model & $s(h_t)$ & $s(a_t)$ & $s(\text{target})$ & $\cos(a_t,\text{bias})$ \\
\midrule
No alignment & 0.618 & 0.648 & -- & untrained \\
Raw MAE & 0.716 & 0.948 & 0.876 & 0.737 \\
LAP-N (MAE) & 0.764 & 0.137 & 0.072 & $-0.004$ \\
REPA (DINOv2) & 0.690 & 0.465 & 0.281 & 0.192 \\
LAP-L (DINOv2) & 0.690 & 0.410 & 0.251 & $-0.007$ \\
\bottomrule
\end{tabular}
\end{table}

\subsection{Receptive-Field Selection and the CLIP Check}

For MAE, the selection rule chooses the largest feasible tested $k$, without using diffusion-generation scores. Among $k\in\{1,3,5\}$, it selects $k=5$. DINOv2 uses fixed $k=3$ in the base comparisons. Feasibility controls at $k=5,7$ give $\rho=0.151,0.168$, with class accuracy near $0.794$ against raw accuracy $0.798$. Both pass the feasibility rule, so the base DINOv2 setting is not a selection over the complete grid.

The DINOv2 comparison gives mean generation scores $1532.4$ and $1547.5$ for $k=3$ and $k=5$, using four seeds each. The available runs do not establish a benefit from increasing the receptive field. CLIP uses the cheaper fixed $k=3$ in the base comparisons, although $k=5$ also passes the criteria. \Cref{tab:clip_ladder} records both choices.

\begin{table}[H]
\centering\small
\caption{\textbf{CLIP target transformations.} Dev-200/50K FID-DINOv2. Lower is better. Means are computed before rounding the individual seed values. Bold marks the lowest FID in each column.}
\label{tab:clip_ladder}
\begin{tabular}{lrrr}
\toprule
Target & Seed 0 & Seed 1 & Mean \\
\midrule
Raw CLIP & 1548.9 & 1553.0 & 1550.9 \\
LAP-L & \textbf{1517.1} & 1513.9 & \textbf{1515.5} \\
LAP-N, $k=3$ & 1537.6 & \textbf{1498.1} & 1517.9 \\
LAP-N, $k=5$ & 1527.7 & 1516.2 & 1522.0 \\
\bottomrule
\end{tabular}
\end{table}

The three purified CLIP means lie within $6.5$ points of one another. Each improves on raw CLIP by about $29$--$35$ points, much less than the approximately $158$-point MAE improvement. The two-seed results do not resolve which CLIP transformation is best. CLIP also illustrates why $\rho$ is not a universal ranking rule: it has a smaller predictable fraction than DINOv2 but a slightly larger observed development-scale gain.

\subsection{Interaction with iREPA-Style Operations}
\label{app:spatial_interaction}

We test whether LAP-N and operations inspired by iREPA~\citep{singh2025irepa} provide complementary benefits. Our control replaces the MLP's first layer with a $3\times3$ convolution and its remaining two layers with $1\times1$ convolutions, preserving depth, intermediate width, and nonlinearities. It also standardizes target features across spatial tokens within each image and feature dimension. The original iREPA head is a single $3\times3$ convolution. Our depth-matched control therefore studies spatial mixing and normalization under our recipe, rather than reproducing the original iREPA architecture.

The design crosses raw versus LAP-N targets with an MLP versus the combined convolutional-head and spatial-normalization operations. \Cref{tab:irepa_interaction} reports the individual values. Each cell uses two dev-200/50K runs. We also apply the operations to the predictable-only DINOv2 target.

\begin{table}[H]
\centering\small
\setlength{\tabcolsep}{5pt}
\caption{\textbf{LAP-N and iREPA-inspired operations.} FID-DINOv2 for seeds 0 and 1. The right column bundles a three-layer convolutional head and spatial normalization. Bold marks the lowest FID for each encoder, head, and training seed.}
\label{tab:irepa_interaction}
\begin{tabular}{lcc}
\toprule
Target & MLP head & Conv head $+$ spatial norm \\
\midrule
Raw MAE & 1641.4 / 1643.2 & \textbf{1508.0} / 1496.9 \\
LAP-N (MAE) ($k{=}5$) & \textbf{1486.2} / \textbf{1482.6} & 1510.7 / \textbf{1484.4} \\
Raw DINOv2 ($=$ REPA) & 1551.2 / \textbf{1518.6} & \textbf{1476.6} / \textbf{1504.3} \\
LAP-N (DINOv2) ($k{=}3$) & \textbf{1508.4} / 1536.2 & 1514.3 / 1523.7 \\
Predictable-only (DINOv2) & 1679.9 / 1641.6 & 1648.6 / 1653.8 \\
\bottomrule
\end{tabular}
\end{table}

The combined operations improve raw MAE by about $140$ points, but increase the LAP-N (MAE) mean by about $13$ points. For DINOv2, they improve the raw and LAP-N means by about $44$ and $3$ points. Defining interaction as the raw-target improvement minus the LAP-N-target improvement gives about $153$ points for MAE and $41$ for DINOv2. The MAE result is the clearer indication of overlapping benefits. The smaller DINOv2 differences remain sensitive to seed variation.

The combined operations improve raw MAE by about $140$ points, compared with about $158$ for LAP-N. These comparisons are consistent with shared effects on easy-to-match content. They do not identify which operation causes the remaining improvement or prove a common mechanism.

On the predictable-only target, the combined operations give a mean of $1651.2$. Spatial normalization alone gives $1665.4/1650.1$, or $1657.8$ on average. Both remain close to the unaligned mean of $1651.2$. Thus, these operations do not recover a clear benefit from the predictable-only target in this development setting.

A four-seed comparison gives $1506.0$ for raw DINOv2 with the combined operations and $1532.4$ for LAP-N (DINOv2) alone. This comparison does not resolve their ordering. We consequently treat the MAE interaction as the stronger observation and do not claim that LAP-N supersedes iREPA. These are dev-200/50K comparisons. At full scale, adding the iREPA-style operations to LAP-N (MAE) improves mean FID-50K from $42.30$ to $41.32$ (\Cref{tab:compose}), so the interaction depends on the training setting.

\subsection{A Check on Relational Structure}

We additionally compare pairwise patch-token cosine matrices on 512 development images. For each image, center the tokens of the raw target, purified target, and clean latent separately, then form their cosine matrices. Let $s_z$, $s_y$, and $s_X$ contain the upper-triangular off-diagonal entries of these matrices. Regress both $s_z$ and $s_y$ on $[\mathbf{1},s_X]$ and compute the Pearson correlation between their residuals. We average that correlation across images. This removes linear dependence on the latent-patch similarity vector, not all appearance information.

The images are the last 512 entries in dataset order and were not explicitly excluded from purifier fitting. This is a descriptive relational-retention diagnostic, not held-out validation or iREPA's spatial-structure metric. The recorded DINOv2 retention scores are $0.97/0.96/0.96$ for $k=3/5/7$. MAE scores range from $0.83$ to $0.93$ across its grid.

The larger structural change occurs for MAE, which also benefits more from purification. Under this diagnostic, structural change therefore does not track generation harm across the tested targets. This does not rule out a role for spatial structure. The measurement and the intervention probe different aspects of the representation.

\FloatBarrier
\section{Robustness to Batch Size, Capacity, and Guidance}
\label{app:regime}

The base result is specific to SiT-B/2 trained with batch 64 for 400K steps. We test larger batches, larger models, and a longer training budget to determine which conclusions extend beyond that setting. Each row of \Cref{tab:regime} changes the model size, the batch size, or the number of training steps of the base recipe, which is the first row, and reports FID-50K without and with guidance. All entries are two-seed means. The last column adds LAP-L to the DINOv2 teacher.

\begin{table}[H]
\centering\small
\setlength{\tabcolsep}{4.5pt}
\renewcommand{\arraystretch}{1.08}
\caption{\textbf{Generation quality across model size, batch size, training budget, and guidance.} FID-50K, two-seed means; lower is better. Within a row all methods share the configuration, and rows are not comparable to each other. The first row ($^\ast$) is the base setting of \Cref{tab:adm}. REPA aligns to the raw DINOv2-B target and the last column to its LAP-L target. Guidance uses CFG scale $1.5$. Bold marks the lowest FID within each row.}
\label{tab:regime}
\begin{tabular}{lrr@{\hspace{10pt}}rrrrr}
\toprule
Model & Batch & Steps & Unaligned & Raw MAE & REPA & LAP-N (MAE) & LAP-L (DINOv2) \\
\midrule
\multicolumn{8}{l}{\emph{Without guidance}} \\
SiT-B/2 & 64 & 400K$^\ast$ & 53.92 & 53.05 & 41.75 & 42.30 & \textbf{41.47} \\
SiT-B/2 & 64 & 800K & 43.48 & 42.60 & 33.44 & 33.95 & \textbf{33.24} \\
SiT-B/2 & 256 & 400K & 36.54 & 34.72 & 27.67 & 28.57 & \textbf{27.58} \\
SiT-L/2 & 64 & 400K & 37.93 & 35.07 & 19.28 & 23.77 & \textbf{18.60} \\
SiT-L/2 & 256 & 400K & 20.14 & 16.05 & 10.01 & 11.97 & \textbf{9.78} \\
SiT-XL/2 & 256 & 400K & 17.86 & 13.65 & 8.10 & 9.67 & \textbf{7.76} \\
\midrule
\multicolumn{8}{l}{\emph{With guidance (CFG 1.5)}} \\
SiT-B/2 & 64 & 400K$^\ast$ & 27.70 & 26.37 & 17.82 & 17.90 & \textbf{17.60} \\
SiT-B/2 & 64 & 800K & 18.12 & 17.08 & 11.37 & 11.72 & \textbf{11.26} \\
SiT-B/2 & 256 & 400K & 12.94 & 11.62 & 7.89 & 8.40 & \textbf{7.79} \\
SiT-L/2 & 64 & 400K & 15.64 & 13.53 & 5.14 & 7.44 & \textbf{4.96} \\
SiT-L/2 & 256 & 400K & 5.37 & 4.19 & 2.95 & 3.60 & \textbf{2.89} \\
SiT-XL/2 & 256 & 400K & 4.71 & 3.81 & \textbf{2.87} & 3.42 & 2.89 \\
\bottomrule
\end{tabular}
\end{table}

\subsection{Larger Batch and Model}

\textbf{SiT-B/2, batch 256.} LAP-N (MAE) remains close to REPA: FID-50K is $28.57$ versus $27.67$ over two seeds ($28.98/28.16$ against $27.96/27.37$), a gap of $0.90$. Raw MAE also improves over the unaligned model in this setting. Its lack of benefit in the base configuration should not be generalized to every training recipe.

\textbf{SiT-L/2.} LAP-N improves MAE at both tested batch sizes, but a gap to REPA remains. At batch 64, the standard FID values are $37.93$ for unaligned, $35.07$ for raw MAE, $23.77$ for LAP-N (MAE), and $19.28$ for REPA (two-seed means). At batch 256, they are $20.14$, $16.05$, $11.97$, and $10.01$.

For a common baseline, define the fraction of the REPA gain recovered by LAP-N as
\begin{equation}
G=\frac{\mathrm{FID}_{\mathrm{unaligned}}-\mathrm{FID}_{\mathrm{LAP-N}}}
{\mathrm{FID}_{\mathrm{unaligned}}-\mathrm{FID}_{\mathrm{REPA}}}.
\end{equation}
LAP-N (MAE) recovers about $76\%$ of the standard-FID gain on SiT-L/2 at batch 64 and $81\%$ at batch 256. These are descriptive summaries of two seeds, not confidence bounds.

The larger-model results also show a quality tradeoff. REPA has higher Inception Score than LAP-N (MAE) at both batch sizes ($70.3$ vs.\ $56.8$, and $110.5$ vs.\ $96.2$). We therefore describe the result as substantial recovery, rather than parity across metrics.

\textbf{SiT-XL/2, batch 256.} At the largest tested scale, FID-50K is $17.86$ for unaligned, $13.65$ for raw MAE, $8.10$ for REPA, and $9.67$ for LAP-N (MAE), two-seed means; our REPA value is close to the $7.9$ reported by \citet{yu2025repa} at the same budget. LAP-N improves raw MAE by $4.0$ FID and recovers about $84\%$ of the REPA gain, trailing REPA by $1.6$ FID.

\textbf{Twice the training budget.} Continuing the four base SiT-B/2, batch-64 runs from their 400K checkpoints to 800K steps with the same recipe gives FID-50K $43.48$ (unaligned), $42.60$ (raw MAE), $33.44$ (REPA), and $33.95$ (LAP-N, MAE), two-seed means (\Cref{tab:regime}). The ordering of the base setting is unchanged: raw MAE stays within $0.9$ FID of no alignment, LAP-N improves it by $8.7$ FID ($10.8$ at 400K), and purified MAE trails DINOv2-based REPA by about $0.52$ FID on average ($34.69/33.21$ against $33.48/33.39$). Under guidance at CFG scale $1.5$, the 800K models give $18.12$, $17.08$, $11.37$, and $11.72$, so LAP-N (MAE) stays within $0.4$ FID of REPA. For DINOv2, LAP-L stays close to REPA across all six configurations in \Cref{tab:regime}, including the base 400K setting. It has the lower mean FID-50K in all six unguided comparisons and in five of the six guided comparisons. Every absolute difference between REPA and LAP-L is below $0.7$ FID, small next to the MAE gains.

\subsection{Alignment Depth}
\label{app:depth}

The base SiT-B/2 recipe aligns at block 8. For SiT-B/2, \citet{singh2025irepa} report better REPA results at block 4, and REG and sREPA use depth 4 in their original settings. We compare block 4 with block 8 on ImageNet-1k at batch 64, using two seeds, the raw target and the best LAP variant from \Cref{tab:adm}, and REG with and without LAP-L. \Cref{tab:depth} reports Inception FID-10K at 100K steps and FID-50K at 400K steps. Block 4 improves every raw target at 400K: MAE $53.1\to50.5$, AIMv2 $48.0\to42.8$, CLIP $44.7\to40.6$, DINOv2 $41.9\to36.2$, DINOv3 $42.7\to37.4$, and REG $29.2\to26.2$. The LAP gains persist and grow at block 4, in both seeds: MAE $11.7$ versus $10.8$ FID, AIMv2 $4.6$ versus $3.7$, and CLIP $3.7$ versus $2.5$; at 100K, AIMv2 and CLIP likewise show larger gains at block 4 ($8.0$ and $7.0$ versus $4.3$ and $4.5$ FID-10K). For DINOv2, DINOv3, and REG, the block-4 LAP changes are small ($-0.5$, $-0.4$, and $-0.1$ FID-50K). The gap between purified MAE and DINOv2-based REPA is $0.5$ at block 8 and $2.6$ at block 4. At 100K, their block-4 FIDs are close ($71.5$ versus $72.0$).

\begin{table}[H]
\centering\small
\setlength{\tabcolsep}{5pt}
\caption{\textbf{Alignment at block 4 versus block 8.} (a) Inception FID-10K of the 100K-step checkpoints; (b) FID-50K at 400K steps. Two training seeds with their mean in parentheses. Target is the alignment target; all rows use REPA except the last pair, which uses REG. The DINOv2 rows and the block-4 REG rows use two training seeds separate from those of \Cref{tab:adm,tab:compose}, so their block-8 DINOv2 values differ slightly from \Cref{tab:adm}; the block-8 REG values come from the training seeds of \Cref{tab:compose}. Bold marks the lowest reported mean per teacher and evaluation budget.}
\label{tab:depth}
\begin{tabular}{llrr}
\toprule
\multicolumn{4}{c}{\textbf{(a) 100K steps: FID-10K}} \\
\midrule
Teacher & Target & Block 4 & Block 8 \\
\midrule
\multirow{2}{*}{MAE} & Raw & 87.26 / 87.09 (87.18) & 87.83 / 86.54 (87.19) \\
 & \ours LAP-N & \ours 71.73 / 71.20 (\textbf{71.46}) & \ours 72.91 / 72.63 (72.77) \\
\multirow{2}{*}{AIMv2} & Raw & 80.95 / 80.94 (80.95) & 82.41 / 82.52 (82.47) \\
 & \ours LAP-L & \ours 72.51 / 73.43 (\textbf{72.97}) & \ours 78.98 / 77.28 (78.13) \\
\multirow{2}{*}{CLIP} & Raw & 80.04 / 80.17 (80.11) & 79.33 / 79.93 (79.63) \\
 & \ours LAP-N & \ours 72.65 / 73.59 (\textbf{73.12}) & \ours 75.67 / 74.63 (75.15) \\
\multirow{2}{*}{DINOv3} & Raw & 73.24 / 73.74 (73.49) & 75.16 / 77.61 (76.38) \\
 & \ours LAP-L & \ours 70.32 / 71.93 (\textbf{71.13}) & \ours 76.88 / 72.60 (74.74) \\
\multirow{2}{*}{DINOv2} & Raw & 71.68 / 72.26 (71.97) & 74.00 / 75.58 (74.79) \\
 & \ours LAP-L & \ours 70.72 / 71.33 (\textbf{71.03}) & \ours 74.00 / 74.27 (74.13) \\
\multirow{2}{*}{DINOv2, REG} & Raw & 63.04 / 61.76 (62.40) & 63.32 / 66.00 (64.66) \\
 & \ours LAP-L & \ours 61.71 / 62.45 (\textbf{62.08}) & \ours 63.70 / 64.90 (64.30) \\
\bottomrule
\end{tabular}

\medskip
\begin{tabular}{llrr}
\toprule
\multicolumn{4}{c}{\textbf{(b) 400K steps: FID-50K}} \\
\midrule
Teacher & Target & Block 4 & Block 8 \\
\midrule
\multirow{2}{*}{MAE} & Raw & 50.54 / 50.49 (50.51) & 53.68 / 52.41 (53.05) \\
 & \ours LAP-N & \ours 38.88 / 38.72 (\textbf{38.80}) & \ours 43.06 / 41.53 (42.30) \\
\multirow{2}{*}{AIMv2} & Raw & 42.79 / 42.84 (42.82) & 48.08 / 47.87 (47.97) \\
 & \ours LAP-L & \ours 37.86 / 38.55 (\textbf{38.20}) & \ours 44.40 / 44.22 (44.31) \\
\multirow{2}{*}{CLIP} & Raw & 40.28 / 40.88 (40.58) & 44.73 / 44.74 (44.74) \\
 & \ours LAP-N & \ours 36.95 / 36.83 (\textbf{36.89}) & \ours 42.32 / 42.11 (42.21) \\
\multirow{2}{*}{DINOv3} & Raw & 37.02 / 37.83 (37.43) & 42.12 / 43.34 (42.73) \\
 & \ours LAP-L & \ours 36.46 / 37.62 (\textbf{37.04}) & \ours 43.23 / 40.92 (42.07) \\
\multirow{2}{*}{DINOv2} & Raw & 35.50 / 36.83 (36.17) & 41.56 / 42.14 (41.85) \\
 & \ours LAP-L & \ours 35.09 / 36.32 (\textbf{35.70}) & \ours 41.04 / 40.91 (40.98) \\
\multirow{2}{*}{DINOv2, REG} & Raw & 26.01 / 26.42 (26.22) & 28.66 / 29.82 (29.24) \\
 & \ours LAP-L & \ours 25.98 / 26.29 (\textbf{26.13}) & \ours 28.48 / 28.85 (28.67) \\
\bottomrule
\end{tabular}
\end{table}

\subsection{Classifier-Free Guidance}

The guidance check evaluates the 400K checkpoints of the base SiT-B/2, batch-64 recipe with CFG scale $1.5$ applied at every sampling step (SDE-250, 50K samples, ADM evaluator), for both training seeds (\Cref{tab:cfg}). The MAE, CLIP, and AIMv2 gains persist under guidance in both seeds ($-8.5$, $-1.6$, and $-2.6$ FID on average), and LAP-N (MAE) recovers $99\%$ of the REPA gain relative to the unaligned model ($27.70\to17.90$ against $17.82$). For DINOv2, DINOv3, and REG the guided difference changes sign between the two seeds and the means lie within $0.2$ FID of each other ($-0.2$, $-0.2$, and $+0.1$), consistent with the small unguided differences. Mean recall is higher with LAP for these three pairs. The same check on the MAE tokenizer rows of \Cref{tab:tokenizers} and on the block-4 pairs of \Cref{app:depth} gives the same picture: under guidance, LAP-N improves MAE by $3.4$ FID on the REPA-E VAE ($13.31\to9.93$), by $9.5$ on EQ-VAE ($26.09\to16.58$), and by $9.0$ at block 4 ($24.84\to15.80$), in both seeds, whereas DINOv2 at block 4 changes little ($13.73$ vs.\ $13.41$). In the larger settings of \Cref{tab:regime}, guidance keeps the same ordering: LAP-N (MAE) reaches $8.40$, $7.44$, $3.60$, and $3.42$ FID against REPA's $7.89$, $5.14$, $2.95$, and $2.87$ at SiT-B/2 batch 256, SiT-L/2 batch 64, SiT-L/2 batch 256, and SiT-XL/2 batch 256.

\begin{table}[H]
\centering\small
\setlength{\tabcolsep}{5pt}
\caption{\textbf{Guided sampling.} FID-50K, Inception Score, precision, and recall at CFG scale $1.5$ for the 400K checkpoints; two-seed means (the predictable-only control has one seed). The first block is the base recipe of \Cref{tab:adm}; each later block changes one component: the REG recipe (\Cref{tab:compose}), the frozen tokenizer (\Cref{tab:tokenizers}), or the alignment depth (\Cref{tab:depth}, with its training seeds). Target is the alignment target. Shaded rows use LAP; bold marks the better value within each raw/LAP pair.}
\label{tab:cfg}
\begin{tabular}{llrrrr}
\toprule
Teacher & Target & FID $\downarrow$ & IS $\uparrow$ & Prec. $\uparrow$ & Rec. $\uparrow$ \\
\midrule
\multicolumn{6}{l}{\emph{Base recipe: REPA, SD-VAE, block 8}} \\
-- & No alignment & 27.70 & 53.78 & 0.603 & 0.530 \\
\multirow{2}{*}{MAE} & Raw & 26.37 & 57.46 & 0.606 & \textbf{0.532} \\
 & \ours LAP-N & \ours \textbf{17.90} & \ours \textbf{79.89} & \ours \textbf{0.676} & \ours 0.529 \\
\multirow{2}{*}{AIMv2} & Raw & 22.32 & 68.58 & 0.638 & 0.528 \\
 & \ours LAP-L & \ours \textbf{19.68} & \ours \textbf{77.35} & \ours \textbf{0.654} & \ours \textbf{0.545} \\
\multirow{2}{*}{CLIP} & Raw & 19.75 & 76.23 & 0.654 & 0.536 \\
 & \ours LAP-N & \ours \textbf{18.12} & \ours \textbf{82.83} & \ours \textbf{0.661} & \ours \textbf{0.542} \\
\multirow{2}{*}{DINOv3} & Raw & 18.42 & 82.29 & \textbf{0.656} & 0.536 \\
 & \ours LAP-L & \ours \textbf{18.22} & \ours \textbf{84.34} & \ours \textbf{0.656} & \ours \textbf{0.553} \\
\multirow{3}{*}{DINOv2} & Raw & 17.82 & 85.97 & 0.665 & 0.538 \\
 & \ours LAP-L & \ours \textbf{17.60} & \ours \textbf{86.07} & \ours \textbf{0.666} & \ours \textbf{0.544} \\
 & Predictable-only & 29.92 & 51.78 & 0.574 & 0.531 \\
\midrule
\multicolumn{6}{l}{\emph{REG recipe}} \\
\multirow{2}{*}{DINOv2} & Raw & \textbf{15.19} & \textbf{110.55} & 0.662 & 0.539 \\
 & \ours LAP-L & \ours 15.31 & \ours 109.02 & \ours \textbf{0.663} & \ours \textbf{0.549} \\
\midrule
\multicolumn{6}{l}{\emph{Frozen REPA-E VAE}} \\
\multirow{2}{*}{MAE} & Raw & 13.31 & 97.91 & 0.751 & \textbf{0.501} \\
 & \ours LAP-N & \ours \textbf{9.93} & \ours \textbf{118.63} & \ours \textbf{0.780} & \ours 0.497 \\
\midrule
\multicolumn{6}{l}{\emph{Frozen EQ-VAE}} \\
\multirow{2}{*}{MAE} & Raw & 26.09 & 57.65 & 0.624 & 0.511 \\
 & \ours LAP-N & \ours \textbf{16.58} & \ours \textbf{86.69} & \ours \textbf{0.701} & \ours \textbf{0.512} \\
\midrule
\multicolumn{6}{l}{\emph{Alignment at block 4}} \\
\multirow{2}{*}{MAE} & Raw & 24.84 & 59.79 & 0.622 & \textbf{0.532} \\
 & \ours LAP-N & \ours \textbf{15.80} & \ours \textbf{86.53} & \ours \textbf{0.697} & \ours 0.529 \\
\multirow{2}{*}{DINOv2} & Raw & 13.73 & 101.30 & 0.705 & 0.533 \\
 & \ours LAP-L & \ours \textbf{13.41} & \ours \textbf{102.90} & \ours \textbf{0.706} & \ours \textbf{0.536} \\
\bottomrule
\end{tabular}
\end{table}

\FloatBarrier
\section{An Encoder-Free Alignment Control}
\label{app:arc}

This experiment asks whether a residual target from the diffusion model's own teacher can provide guidance comparable to a pretrained encoder. We call the residual variant alignment to residual context (ARC) and compare it with an otherwise identical raw self-teacher target.

\subsection{Construction and Evaluation}

The student's block-8 readout is aligned to an EMA teacher's layer-10 hidden state. The teacher processes an independently noised copy of the same image at $t=0.25$. For ARC, we remove the part of the teacher state predicted by local clean and noised patches using an affine fit. ARC-raw skips this removal. Both variants then standardize each feature dimension across batch tokens. Residualization is the manipulated variable.

Both variants use dev-200 and batch 64. The alignment coefficient increases linearly from zero to $0.5$ over the first 10K steps. We evaluate the pre-specified 25K, 50K, 75K, and 100K checkpoints. A diagnostic on unaligned checkpoints gives residual variance fractions of $0.53$--$0.78$ and DINOv2-probe enrichment factors of $1.67$ at 10K and $1.15$ at 50K. These are relative diagnostics of the self-teacher, not evidence that it contains as much useful guidance as DINOv2.

\begin{table}[H]
\centering\small
\caption{\textbf{Encoder-free target comparison.} Dev-200 FID-DINOv2, single runs; the unaligned value at 25K comes from a second training seed. Bold marks the lowest FID at each training checkpoint.}
\label{tab:arc_curves}
\begin{tabular}{lrrrr}
\toprule
Target & 25K & 50K & 75K & 100K \\
\midrule
Unaligned & 2564.6 & 1652.4 & 1454.3 & 1355.1 \\
Raw self-teacher & 2555.0 & \textbf{1622.5} & \textbf{1425.3} & \textbf{1307.5} \\
Residual self-teacher & \textbf{2537.3} & 1623.6 & 1429.5 & 1313.1 \\
\bottomrule
\end{tabular}
\end{table}

\subsection{What the Comparison Establishes}

In \Cref{tab:arc_curves}, the residual-minus-raw difference ranges from $-17.7$ to $+5.6$ across the trajectory, without a consistent advantage for residualization. At 100K, the raw and residual variants improve on the unaligned model by $47.6$ and $42.0$ proxy points. These gains are only about $21$--$24\%$ of the matched REPA gain.

This control provides no clear benefit from residualizing the self-teacher under the tested recipe. It does not establish that pretrained encoders are necessary: the self-teacher may simply provide too little useful structure. Other encoder-free regularizers can improve diffusion training~\citep{wang2025dispersive}. A stronger test would match the quality of the encoder-free target to the pretrained target before comparing residualization.

\FloatBarrier
\section{Generation Evaluation and Additional Results}
\label{app:adm}

\subsection{Standard Evaluation and Mechanism Controls}

Standard evaluation generates 50K images from each 400K EMA checkpoint using the Euler--Maruyama SDE sampler with 250 steps and no classifier-free guidance. We use the ADM/guided-diffusion evaluation suite and its ImageNet-256 reference statistics to report FID, Inception Score, precision, and recall. Base comparisons evaluate both training seeds. Means and sample SDs use two decimals for FID/IS and three for precision/recall; a displayed zero SD can reflect rounding. Larger-model and guidance checks use the coverage stated in \Cref{app:regime}.

\Cref{tab:mech_full} gives the standard metrics of the two interventions of \Cref{sec:mechanism}. The auxiliary prediction loss improves velocity prediction but worsens generation relative to REPA on FID, Inception Score, precision, and recall. The predictable-only target is worse than the unaligned baseline on standard FID and has the lowest precision and recall of all base-setting runs.

\begin{table}[H]
\centering\small
\setlength{\tabcolsep}{4.5pt}
\caption{\textbf{Standard metrics of the interventions of \Cref{sec:mechanism}.} DINOv2 ViT-B teacher, protocol of \Cref{tab:adm}, mean $\pm$ sample SD over two training seeds. The first two rows repeat \Cref{tab:adm} for reference. Bold marks the best mean per metric.}
\label{tab:mech_full}
\begin{tabular}{lrrrr}
\toprule
Target & FID $\downarrow$ & IS $\uparrow$ & Prec. $\uparrow$ & Rec. $\uparrow$ \\
\midrule
No alignment & 53.92\sdev{0.03} & 25.34\sdev{0.16} & 0.435\sdev{0.002} & 0.602\sdev{0.006} \\
REPA (raw DINOv2) & \textbf{41.75}\sdev{0.56} & \textbf{35.50}\sdev{0.62} & \textbf{0.492}\sdev{0.001} & \textbf{0.630}\sdev{0.005} \\
REPA $+$ prediction loss (\Cref{sec:f2}) & 46.79\sdev{0.16} & 29.99\sdev{0.27} & 0.473\sdev{0.001} & 0.617\sdev{0.007} \\
Predictable-only target (\Cref{sec:dissociation}) & 56.71\sdev{0.42} & 24.37\sdev{0.00} & 0.415\sdev{0.001} & 0.591\sdev{0.003} \\
\bottomrule
\end{tabular}
\end{table}

\textbf{Additional teachers.} Six encoders were screened by $C_0$ on dev-200 (2{,}048 images, the protocol of \Cref{app:constant_baseline}) before any training: AIMv2-L~\citep{fini2024aimv2} $0.66$, SigLIP~2-B~\citep{tschannen2025siglip2} $0.57$, DINOv3-B~\citep{simeoni2025dinov3} $0.52$, EVA-02-B~\citep{fang2023eva02} $0.44$, PE-Core-B~\citep{bolya2025pe} $0.44$, and Web-SSL DINO-300M~\citep{fan2025webssl} $0.37$. Each yields $256$ patch tokens after the original $256\times256$ image undergoes the encoder-specific resizing and normalization of \Cref{tab:teacher_preprocessing}; class and register tokens are then excluded. By a rule fixed before the runs, DINOv3 and the highest-$C_0$ candidate, AIMv2, were trained at full scale with the raw target and with LAP-L, two seeds each (\Cref{tab:adm}); LAP-L needs no per-teacher purifier. The same table includes LAP-N, with purifiers selected by the rule of \Cref{app:purifier}. DINOv3 weights are the timm port of the official LVD-1689M checkpoint; AIMv2 is released under a research-only license. Seeds: DINOv3 $42.12/43.34$, with LAP-L $43.23/40.92$; AIMv2 $48.08/47.87$, with LAP-L $44.40/44.22$.

\textbf{REPA-family baselines.} The iREPA-style, VA-REPA, and CLIP rows of \Cref{tab:adm,tab:compose} use the base recipe with the component settings specified below. The iREPA-style rows use the depth-matched control of \Cref{app:spatial_interaction}: spatial standardization of the target within each image and a three-layer convolutional head in place of the MLP; they are not the original single-convolution iREPA head. VA-REPA~\citep{varepa2026} reweights the alignment loss by $w(t)=\sigma\big(k(\tau-t)\big)$ with the authors' defaults $\tau=0.7$, $k=20$, normalized by the batch sum of $w$, and the same $\lambda_a=0.5$; we do not include its companion Stable Velocity Matching objective, so the row isolates the alignment schedule. REPA-E~\citep{leng2025repae} tunes the VAE end to end, which our budget does not cover; its frozen end-to-end-tuned VAE is evaluated in \Cref{tab:tokenizers}, described next. The recipe and evaluation are identical to the base rows. Individual FIDs are in \Cref{tab:seeds_adm,tab:seeds_adm2}.
The composition rows apply those operations or that schedule to a LAP target with no other change. \Cref{tab:compose} reports the evaluated variant with the lowest two-seed mean FID within each recipe. \Cref{tab:tokenizers} uses LAP-N for MAE with every tokenizer. \Cref{tab:adm_full} reports both LAP variants for each teacher and recipe. Individual seeds: LAP-N (MAE) with the VA-REPA schedule $42.24/40.95$, with the iREPA-style operations $41.12/41.52$; raw MAE with the VA-REPA schedule $53.46/52.19$; LAP-L (DINOv2) with the iREPA-style operations $40.28/40.83$ and with the VA-REPA schedule $40.37/40.40$.

\textbf{REG and sREPA.} REG~\citep{reg2025} prepends a noised DINOv2 CLS token to the latent token sequence, denoises it jointly with the image tokens (loss weight $\beta=0.03$, the authors' default), and aligns the resulting 257-token sequence; the CLS token is projected by a linear layer followed by an affine LayerNorm and is discarded at sampling time. Our REG implementation follows the official code. sREPA~\citep{structra2026} adds the mean squared error between the off-diagonal entries of the token self-similarity matrices of the $\ell_2$-normalized projected features and of the target, with the authors' defaults $\lambda_{\mathrm{proj}}=1.0$ and $\lambda_{\mathrm{struc}}=2.0$; no official code was available, and our implementation was checked numerically against the paper's Algorithm~1. In the LAP arms the structural loss is computed on the LAP target; the CLS row of REG is not purified, because it has no co-located latent patch. For MAE, whose forward pass drops the class token, the REG target uses MAE's own post-LayerNorm CLS feature; unlike the DINOv2 CLS token it is nearly constant across images (mean pairwise cosine $0.98$ vs.\ $0.03$), which is consistent with REG giving raw MAE no gain. Both recipes use our alignment depth (block 8) and batch size rather than the authors' (depth 4, batch 256), so absolute FIDs differ from the published ones. Individual seeds: DINOv2 with REG $28.66/29.82$, with LAP-L $28.48/28.85$; DINOv2 with sREPA $38.14/38.13$, with the selected LAP-N variant $37.21/36.62$; MAE with REG $53.29/52.54$, with LAP-N $41.70/41.61$; MAE with sREPA $50.06/50.80$, with LAP-N $38.90/38.02$.

\begin{table}[H]
\centering\small
\setlength{\tabcolsep}{4.5pt}
\renewcommand{\arraystretch}{1.0}
\caption{\textbf{Linear and nonlinear purification under the REPA-family recipes of \Cref{tab:compose}.} Protocol of \Cref{tab:adm}, mean $\pm$ sample SD over two training seeds. MAE under REG and sREPA was trained with LAP-N only. Shaded rows use LAP. Bold marks the best mean within each teacher--recipe group and metric, including ties at displayed precision.}
\label{tab:adm_full}
\begin{tabular}{llrrrr}
\toprule
Recipe & Target & FID $\downarrow$ & IS $\uparrow$ & Prec. $\uparrow$ & Rec. $\uparrow$ \\
\midrule
\multicolumn{6}{l}{\emph{MAE ViT-L teacher}} \\
\multirow{3}{*}{iREPA} & Raw & 43.06\sdev{0.62} & 33.23\sdev{0.66} & 0.491\sdev{0.004} & 0.617\sdev{0.006} \\
 & \ours LAP-L & \ours 43.10\sdev{0.40} & \ours 33.07\sdev{0.62} & \ours 0.490\sdev{0.003} & \ours \textbf{0.627}\sdev{0.005} \\
 & \ours LAP-N & \ours \textbf{41.32}\sdev{0.28} & \ours \textbf{34.70}\sdev{0.37} & \ours \textbf{0.502}\sdev{0.002} & \ours 0.626\sdev{0.001} \\
\cmidrule(lr){1-6}
\multirow{3}{*}{VA-REPA} & Raw & 52.82\sdev{0.90} & 26.32\sdev{0.58} & 0.437\sdev{0.001} & 0.606\sdev{0.006} \\
 & \ours LAP-L & \ours 43.43\sdev{0.63} & \ours 32.65\sdev{0.63} & \ours 0.492\sdev{0.003} & \ours 0.620\sdev{0.001} \\
 & \ours LAP-N & \ours \textbf{41.60}\sdev{0.91} & \ours \textbf{34.06}\sdev{0.53} & \ours \textbf{0.503}\sdev{0.004} & \ours \textbf{0.622}\sdev{0.004} \\
\cmidrule(lr){1-6}
\multirow{2}{*}{REG} & Raw & 52.92\sdev{0.53} & 26.33\sdev{0.41} & 0.434\sdev{0.002} & 0.608\sdev{0.002} \\
 & \ours LAP-N & \ours \textbf{41.65}\sdev{0.06} & \ours \textbf{34.31}\sdev{0.16} & \ours \textbf{0.500}\sdev{0.003} & \ours \textbf{0.626}\sdev{0.002} \\
\cmidrule(lr){1-6}
\multirow{2}{*}{sREPA} & Raw & 50.43\sdev{0.53} & 27.41\sdev{0.22} & 0.454\sdev{0.002} & 0.606\sdev{0.009} \\
 & \ours LAP-N & \ours \textbf{38.46}\sdev{0.62} & \ours \textbf{36.89}\sdev{0.30} & \ours \textbf{0.519}\sdev{0.003} & \ours \textbf{0.622}\sdev{0.002} \\
\midrule
\multicolumn{6}{l}{\emph{DINOv2 ViT-B teacher}} \\
\multirow{3}{*}{iREPA} & Raw & 41.32\sdev{0.19} & 35.65\sdev{0.01} & 0.496\sdev{0.004} & \textbf{0.635}\sdev{0.003} \\
 & \ours LAP-L & \ours \textbf{40.55}\sdev{0.39} & \ours \textbf{36.29}\sdev{0.16} & \ours \textbf{0.503}\sdev{0.003} & \ours 0.632\sdev{0.004} \\
 & \ours LAP-N & \ours 41.03\sdev{0.65} & \ours 35.70\sdev{0.66} & \ours 0.500\sdev{0.002} & \ours 0.633\sdev{0.004} \\
\cmidrule(lr){1-6}
\multirow{3}{*}{VA-REPA} & Raw & 40.71\sdev{0.14} & 36.27\sdev{0.24} & 0.500\sdev{0.002} & 0.635\sdev{0.001} \\
 & \ours LAP-L & \ours \textbf{40.39}\sdev{0.02} & \ours \textbf{36.54}\sdev{0.04} & \ours \textbf{0.502}\sdev{0.001} & \ours \textbf{0.637}\sdev{0.006} \\
 & \ours LAP-N & \ours 41.89\sdev{0.62} & \ours 35.45\sdev{0.72} & \ours 0.496\sdev{0.003} & \ours 0.628\sdev{0.004} \\
\cmidrule(lr){1-6}
\multirow{3}{*}{REG} & Raw & 29.24\sdev{0.82} & 57.08\sdev{2.10} & 0.559\sdev{0.003} & 0.610\sdev{0.001} \\
 & \ours LAP-L & \ours \textbf{28.67}\sdev{0.26} & \ours 57.49\sdev{0.07} & \ours \textbf{0.567}\sdev{0.005} & \ours \textbf{0.615}\sdev{0.004} \\
 & \ours LAP-N & \ours 29.51\sdev{0.19} & \ours \textbf{57.64}\sdev{0.94} & \ours 0.559\sdev{0.003} & \ours 0.607\sdev{0.008} \\
\cmidrule(lr){1-6}
\multirow{3}{*}{sREPA} & Raw & 38.13\sdev{0.01} & 39.55\sdev{0.01} & 0.515\sdev{0.005} & 0.635\sdev{0.004} \\
 & \ours LAP-L & \ours 37.63\sdev{0.86} & \ours 39.88\sdev{1.11} & \ours 0.518\sdev{0.003} & \ours \textbf{0.636}\sdev{0.001} \\
 & \ours LAP-N & \ours \textbf{36.92}\sdev{0.42} & \ours \textbf{41.11}\sdev{0.46} & \ours \textbf{0.522}\sdev{0.002} & \ours 0.631\sdev{0.005} \\
\bottomrule
\end{tabular}
\end{table}

\textbf{Frozen REPA-E VAE.} The REPA-E VAE rows of \Cref{tab:tokenizers,tab:tok_full} replace the SD-VAE by the end-to-end-tuned VAE released with REPA-E~\citep{leng2025repae}, used frozen: latents are encoded online during training and normalized with the released per-channel statistics, and samples are decoded with the same VAE. Everything else follows the base recipe ($\lambda_a=0.5$, 400K steps, ADM evaluation). With the DINOv2-B target (\Cref{tab:tok_full}), the unaligned seeds are $32.27/32.33$, the REPA seeds $25.88/26.33$, and the selected LAP-N seeds $25.03/24.85$, giving lower FID than raw-target REPA in both seeds. The LAP-L seeds are $25.98/25.39$, yielding a smaller mean improvement (\Cref{tab:tok_full,tab:seeds_adm2}). With the MAE ViT-L target on the same frozen VAE (\Cref{tab:tokenizers}) (purifier refitted on its latents, $k=5$, held-out $\rho=0.357$), the raw-target seeds are $33.42/34.06$ and the LAP-N seeds $27.84/27.28$.

\textbf{Frozen EQ-VAE.} The EQ-VAE rows of \Cref{tab:tokenizers} replace the SD-VAE by the equivariance-regularized fine-tune of \citet{kouzelis2025eqvae}, frozen, using the 5-epoch OpenImages checkpoint on which that paper's SiT-B/2 results are based. Per-channel latent means and standard deviations were measured on $8{,}192$ training images (the released configuration's scalar scaling factor is the SD-VAE default and does not match the measured statistics); latents are encoded online and samples decoded with the same VAE; the released weights carry no declared license. With the MAE ViT-L target the LAP-N purifier was refitted on EQ-VAE latents ($k=5$, held-out $\rho=0.343$). Seeds: raw target $51.44/53.38$, LAP-N $40.19/39.83$; the unaligned seeds are $52.47/52.53$.

\begin{table}[H]
\centering\small
\setlength{\tabcolsep}{4.5pt}
\caption{\textbf{Linear and nonlinear purification with the DINOv2 teacher on the frozen REPA-E VAE.} Protocol of \Cref{tab:adm}, mean $\pm$ sample SD over two training seeds. The MAE results on this tokenizer and on EQ-VAE are in \Cref{tab:tokenizers}. Shaded rows use LAP; bold marks the best mean per metric.}
\label{tab:tok_full}
\begin{tabular}{lrrrr}
\toprule
Target & FID $\downarrow$ & IS $\uparrow$ & Prec. $\uparrow$ & Rec. $\uparrow$ \\
\midrule
No alignment & 32.30\sdev{0.04} & 40.61\sdev{0.09} & 0.584\sdev{0.002} & 0.604\sdev{0.005} \\
Raw & 26.10\sdev{0.32} & 52.41\sdev{0.72} & 0.611\sdev{0.004} & \textbf{0.620}\sdev{0.001} \\
\ours LAP-L & \ours 25.69\sdev{0.41} & \ours 53.11\sdev{0.25} & \ours \textbf{0.615}\sdev{0.003} & \ours 0.614\sdev{0.000} \\
\ours LAP-N & \ours \textbf{24.94}\sdev{0.12} & \ours \textbf{55.13}\sdev{0.41} & \ours 0.614\sdev{0.003} & \ours 0.617\sdev{0.005} \\
\bottomrule
\end{tabular}
\end{table}

\subsection{Pre-Specified Encoder Sweep and an External Check of \texorpdfstring{$C_0$}{C0}}
\label{app:sweep}

\textbf{Motivation.} The five teachers of \Cref{tab:adm} order the LAP-L gain by raw $C_0$, but $C_0$ and raw-target quality are correlated under the paper's own hypothesis, since a large shared direction is what makes a raw target easy to match without image-specific computation. Five points cannot separate the two. We therefore (i) test $C_0$ against the published results of \citet{singh2025irepa} and (ii) run a pre-specified sweep of eight further encoders.

\textbf{External check.} We measured $C_0$ (dev-200, protocol of \Cref{app:constant_baseline}) for $37$ encoders, including every encoder of \citet{singh2025irepa} that we could load (\Cref{tab:c0_sweep} lists those used below; the full screen is released with the code). Against their published SiT-XL/2 results at 100K steps ($25$ encoders; their appendix bar-chart labels, two of which differ slightly from their tables), raw $C_0$ has Spearman correlation $0.81$ with REPA FID and $0.70$ with the gain from REPA to iREPA (both $p<10^{-3}$), but the partial rank correlation of the gain with $C_0$ given REPA FID is $-0.13$ ($p=0.55$). Their SiT-B/2 400K table ($16$ encoders) gives $0.54$, $0.49$, and $0.23$ ($p=0.03$, $0.06$, $0.38$), and their SiT-L/2 table ($8$ encoders) $0.25$, $0.24$, and $0.10$. The shared direction thus predicts which teachers align poorly and gain most from their spatial normalization, but in published data it cannot be separated from teacher quality. Their gain also includes a convolutional projector, which their ablation credits with $0.5$--$7.4$ FID on its own.

\textbf{Sweep design.} Eight encoders were selected by a rule specified before their generation results were available (the design and amendment accompany the code): candidates were stratified by raw $C_0$ into four bins with two encoders each, preferring, within a bin, encoders off the $C_0$-versus-quality diagonal (using iREPA's REPA FID as the quality proxy) and then size variants of families already in \Cref{tab:adm}; encoders above one billion parameters were excluded for cost. Each is trained with the raw target and with LAP-L, which needs no fitted purifier, for 100K steps on ImageNet-1k with the base recipe and two seeds, and evaluated with the Inception FID-10K protocol of \Cref{fig:training_rescue}. The primary test concerns these eight added encoders. An amendment recorded before their results existed also specifies a combined comparison with the five teachers of \Cref{tab:adm}, evaluated at their matched 100K checkpoints. We report both sets separately below. Pre-specified predictions: the LAP-L gain increases with raw $C_0$ (above $3$ FID for $C_0>0.7$, below $1$ FID for $C_0<0.35$); as a secondary predictor, the gain follows $\Delta C_0=C_0(\mathrm{raw})-C_0(\mathrm{LAP\text{-}L})$, the shared direction that LAP-L actually removes, which matters for CLIP-B/16, whose LAP-L target retains $C_0=0.25$; and PE-Core-L, the designated off-diagonal encoder (low $C_0$, poor published REPA FID), gains less than the median encoder despite its poor raw FID. All 32 sweep runs were continued from their 100K checkpoints to 400K steps and evaluated with FID-50K under the protocol of \Cref{tab:adm}, so the 400K columns of \Cref{tab:c0_sweep} compare the eight encoders with the five teachers at their standard endpoint.

\textbf{Eight added encoders.} These encoders contribute 32 runs (\Cref{tab:c0_sweep}). On this primary test set alone, the LAP-L gain has Spearman correlation $0.810$ with raw $C_0$ at 100K steps (exact one-sided permutation $p=0.0109$) and $0.619$ at the 400K endpoint ($p=0.0575$). The corresponding partial rank correlations given raw-target FID are $0.756$ and $0.577$. The association is positive at both endpoints, but the 400K result does not reach the $0.05$ significance level. Evidence on these eight encoders is weaker than in the combined comparison that includes the five teachers motivating the hypothesis.

\textbf{Combined 100K comparison.} Across the 13 encoders, the LAP-L gain has Spearman correlation $0.94$ with raw $C_0$ (one-sided permutation $p\approx3.0\times10^{-6}$), $0.85$ with $\Delta C_0$, and $0.80$ with the raw-target FID; $C_0$ and raw FID are themselves correlated ($0.90$). The partial rank correlation of the gain with $C_0$ given raw FID is $0.83$, whereas that of the gain with raw FID given $C_0$ is $-0.26$: the association with the shared direction persists after controlling for raw-target FID. In this combined set, the pre-specified thresholds held for five of six encoders: the three encoders with $C_0>0.7$ gain $11.5$, $10.7$, and $4.5$ FID, and two of the three with $C_0<0.35$ gain $0.7$ and $-0.2$, while DINOv2-L gains $1.6$. The off-diagonal test came out as predicted: PE-Core-L has a poor raw-target FID ($82.0$) yet no gain ($0.0$; $+1.7$ and $-1.7$ in the two seeds) and ranks 12th of 13. The same order holds within families: DINOv3-H+, -B, and -L ($C_0$ of $0.68$, $0.53$, and $0.48$) gain $3.2$, $1.6$, and $0.7$; the three DINOv2 sizes ($0.22$--$0.25$) gain between $-0.2$ and $1.6$; MAE-B and MAE-L ($0.94$ and $0.93$) gain $11.5$ and $10.7$. CLIP-B/16 retains $C_0=0.25$ after LAP-L yet gains $4.5$, so the secondary predictor $\Delta C_0$ did not improve on raw $C_0$. Seed-to-seed differences in the gain reach $6.7$ FID for DINOv3-B and $3.4$ for PE-Core-L, so individual gains below about $2$ FID are within noise, and the evidence is the ordering, not any single encoder.

\begin{figure}[H]
\centering
\includegraphics[width=\textwidth]{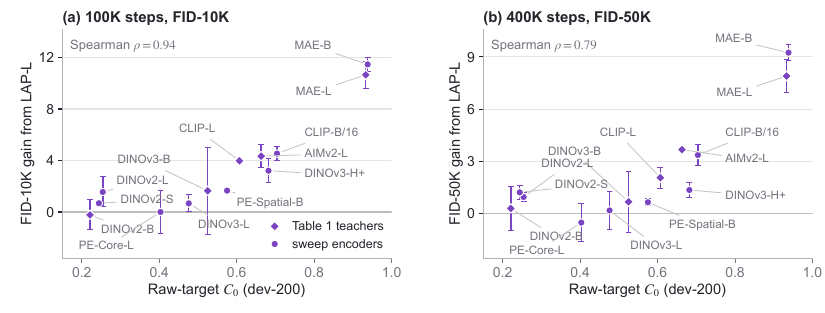}
\caption{\textbf{$C_0$ predicts the purification gain at 100K and 400K steps.} FID gain from LAP-L against raw-target $C_0$ (dev-200) for 13 encoders, two-seed means with seed ranges: (a) Inception FID-10K at 100K steps, (b) FID-50K at 400K steps. Diamonds are the five teachers of \Cref{tab:adm}; values in \Cref{tab:c0_sweep}.}
\label{fig:sweep_400k}
\end{figure}

\textbf{Combined 400K comparison.} At 400K steps (FID-50K, right half of \Cref{tab:c0_sweep} and \Cref{fig:sweep_400k}b), the same ordering holds with more noise: the LAP-L gain has Spearman correlation $0.79$ with raw $C_0$ (one-sided permutation $p\approx0.0010$) and $0.55$ with the raw-target FID, the partial rank correlations are $0.69$ with $C_0$ given raw FID and $-0.14$ with raw FID given $C_0$, and the 100K and 400K gains agree with rank correlation $0.92$. The three encoders with $C_0>0.7$ gain $9.3$, $7.9$, and $3.4$ FID, PE-Core-L again gains nothing ($-0.5$) despite the third-worst raw FID among all 13 encoders, and the three DINOv2 sizes gain $0.3$--$1.2$, slightly above the pre-specified $1$ FID for only one of them. Seed-to-seed differences in the 400K gain reach $3.5$ FID (DINOv3-B), so single gains below about $2$ FID remain within noise.

For both encoder sets, the permutation-test alternative is the pre-specified positive association. We hold the $C_0$ ranks fixed and permute gain ranks across encoder labels. For the eight-encoder primary set, we enumerate all $8!=40{,}320$ permutations at each endpoint and report the fraction with Spearman correlation at least as large as observed. For the combined 13-encoder set, we sample $B=2{,}000{,}000$ uniform permutations at each endpoint and report $p=(b+1)/(B+1)$, where $b$ counts correlations at least as large as observed. The combined-set 400K two-sided value, counting absolute correlations, is approximately $0.0020$. The accompanying verification script reproduces both tests and both encoder-set summaries from the per-seed result files.
\begin{table}[H]
\centering\small
\setlength{\tabcolsep}{4.5pt}
\caption{\textbf{Encoder sweep at 100K and 400K steps.} Dev-200 $C_0$ of the raw and LAP-L targets; Inception FID-10K of the 100K-step checkpoints and FID-50K of the 400K-step checkpoints with the raw and the LAP-L target, two-seed means, with the gain (raw minus LAP-L) at each step count. Rows are ordered by raw $C_0$; $^\dagger$ marks the five teachers of \Cref{tab:adm}, whose 400K values are those of \Cref{tab:adm,tab:seeds_adm}. Bold marks the lower FID between raw and LAP-L targets within each row and step count, including ties at displayed precision. Seed-level gains are given in the text.}
\label{tab:c0_sweep}
\begin{tabular}{lrrrrrrrr}
\toprule
 & & & \multicolumn{3}{c}{100K steps, FID-10K} & \multicolumn{3}{c}{400K steps, FID-50K} \\
\cmidrule(lr){4-6}\cmidrule(lr){7-9}
Encoder & $C_0$ raw & $C_0$ LAP-L & Raw & \ours LAP-L & Gain & Raw & \ours LAP-L & Gain \\
\midrule
MAE ViT-B & 0.938 & 0.022 & 86.94 & \ours \textbf{75.48} & 11.47 & 53.22 & \ours \textbf{43.97} & 9.25 \\
MAE ViT-L$^\dagger$ & 0.933 & 0.020 & 87.19 & \ours \textbf{76.53} & 10.66 & 53.05 & \ours \textbf{45.15} & 7.90 \\
CLIP ViT-B/16 & 0.704 & 0.246 & 83.46 & \ours \textbf{78.92} & 4.53 & 49.05 & \ours \textbf{45.70} & 3.35 \\
DINOv3 ViT-H+ & 0.682 & 0.033 & 80.47 & \ours \textbf{77.27} & 3.20 & 45.59 & \ours \textbf{44.25} & 1.34 \\
AIMv2 ViT-L$^\dagger$ & 0.663 & 0.041 & 82.47 & \ours \textbf{78.13} & 4.34 & 47.97 & \ours \textbf{44.31} & 3.66 \\
CLIP ViT-L$^\dagger$ & 0.607 & 0.088 & 79.63 & \ours \textbf{75.65} & 3.98 & 44.74 & \ours \textbf{42.69} & 2.05 \\
PE-Spatial ViT-B & 0.575 & 0.017 & 76.79 & \ours \textbf{75.13} & 1.66 & 45.94 & \ours \textbf{45.31} & 0.63 \\
DINOv3 ViT-B$^\dagger$ & 0.525 & 0.018 & 76.38 & \ours \textbf{74.74} & 1.64 & 42.73 & \ours \textbf{42.07} & 0.66 \\
DINOv3 ViT-L & 0.476 & 0.028 & 76.30 & \ours \textbf{75.63} & 0.67 & 42.92 & \ours \textbf{42.76} & 0.16 \\
PE-Core ViT-L & 0.403 & 0.030 & \textbf{82.03} & \ours \textbf{82.03} & 0.00 & \textbf{49.25} & \ours 49.79 & $-0.54$ \\
DINOv2 ViT-L & 0.254 & 0.009 & 76.05 & \ours \textbf{74.50} & 1.56 & 42.54 & \ours \textbf{41.61} & 0.93 \\
DINOv2 ViT-S & 0.244 & 0.009 & 74.83 & \ours \textbf{74.15} & 0.68 & 43.19 & \ours \textbf{41.99} & 1.20 \\
DINOv2 ViT-B$^\dagger$ & 0.221 & 0.007 & \textbf{75.03} & \ours 75.25 & $-0.22$ & 41.75 & \ours \textbf{41.47} & 0.28 \\
\bottomrule
\end{tabular}
\end{table}

\textbf{Visualization.}
\Cref{fig:sweep_400k} plots the LAP-L gain of each encoder against its raw-target $C_0$ at 100K steps (FID-10K, panel a, which \Cref{fig:sweep} in the main text shows alone) and at 400K steps (FID-50K, panel b), as two-seed means with seed ranges; within each encoder, the raw and LAP-L targets share the same training and evaluation protocol. The Spearman and partial rank correlations quoted in the main text ($0.94$ and $0.83$ at 100K, $0.79$ and $0.69$ at 400K) are computed from this matched fixed-LAP-L comparison across all 13 encoders in \Cref{tab:c0_sweep}.

\subsection{Checking the Development Metric}

We also evaluate the development samples in InceptionV3 pool3 feature space. This check uses the same 10K generated images and development reference set as the proxy evaluation. It is therefore distinct from standard ImageNet-1k FID-50K.

Across the 11 configurations in \Cref{tab:neutral_metric}, the two feature spaces have Spearman rank correlation $0.93$. Both show a large improvement from purifying MAE, while raw MAE remains close to the unaligned model. They need not agree on the ordering of close methods: three REPA runs span $109.9$--$112.3$ in Inception space and $1518.6$--$1553.3$ in DINOv2 space. The predictable-only target remains near the unaligned baseline at this scale.

This check addresses feature-space sensitivity, not independent training replication. Across independent sample draws, individual DINOv2-proxy scores vary by about $1$--$15$ points (median $7$), so small proxy differences are within evaluation noise.

\begin{table}[H]
\centering\small
\setlength{\tabcolsep}{5pt}
\caption{\textbf{Development results in two feature spaces.} Dev-200/50K, training seed 0; both metrics are computed on the same generated samples within each row. Shaded rows use LAP; bold marks the lowest FID in each feature space.}
\label{tab:neutral_metric}
\begin{tabular}{llrr}
\toprule
Teacher & Target & FID-Inception & FID-DINOv2 \\
\midrule
-- & No alignment & 124.72 & 1652.4 \\
\midrule
\multirow{3}{*}{MAE} & Raw & 122.44 & 1641.4 \\
 & \ours LAP-L & \ours 107.55 & \ours 1504.3 \\
 & \ours LAP-N & \ours \textbf{106.98} & \ours \textbf{1486.2} \\
\midrule
\multirow{3}{*}{CLIP} & Raw & 114.64 & 1548.9 \\
 & \ours LAP-L & \ours 110.72 & \ours 1517.1 \\
 & \ours LAP-N & \ours 110.42 & \ours 1537.6 \\
\midrule
\multirow{4}{*}{DINOv2} & Raw & 110.61 & 1551.2 \\
 & \ours LAP-L & \ours 111.87 & \ours 1546.0 \\
 & \ours LAP-N & \ours 110.09 & \ours 1508.4 \\
 & Predictable-only & 125.01 & 1679.9 \\
\bottomrule
\end{tabular}
\end{table}

\subsection{Complete Training Trajectories}
\label{app:trajectories}

The Inception FID-10K trajectories in \Cref{fig:training_rescue,tab:curves_fid10k} use the ADM sampler and reference statistics of \Cref{tab:adm}, with 10K generated samples per checkpoint. The scores use no guidance and are not numerically interchangeable with FID-50K. At 400K steps, they lie about $2.5$--$3.4$ points above FID-50K and preserve the main ordering (\Cref{tab:curves_fid10k,tab:adm}).

\textbf{Inception FID across training.} LAP-L improves MAE, AIMv2, and CLIP from the first evaluated checkpoint. At 100K steps, their two-seed mean FID-10K gains are $10.7$, $4.3$, and $4.0$, respectively. LAP-N with MAE is below DINOv2-based REPA at 100K ($72.8$ vs.\ $75.0$) and matches it at 400K ($45.1$ vs.\ $45.1$), whereas raw MAE tracks the unaligned model throughout. These are gains per training step, not measured savings in total compute, which also includes fitting and evaluating the purifier (\Cref{app:compute}).

\begin{table}[H]
\centering\small
\setlength{\tabcolsep}{4pt}
\caption{\textbf{Complete Inception FID-10K trajectories.} ImageNet-1k, SiT-B/2, batch 64, no guidance. Mean $\pm$ sample SD over two training seeds at every checkpoint. Each teacher shows the raw target and the LAP variant selected by the final FID-50K in \Cref{tab:adm}, held fixed across checkpoints, adding DINOv2 to the four teachers of \Cref{fig:training_rescue}. Shaded rows use LAP; bold marks the better value within each raw/LAP pair.}
\label{tab:curves_fid10k}
\begin{tabular}{llrrrr}
\toprule
Teacher & Target & 100K & 200K & 300K & 400K \\
\midrule
-- & No alignment & 88.47\sdev{0.60} & 70.79\sdev{1.56} & 62.01\sdev{1.02} & 56.61\sdev{0.26} \\
\midrule
\multirow{2}{*}{MAE ViT-L} & Raw & 87.19\sdev{0.91} & 69.10\sdev{1.03} & 60.87\sdev{1.34} & 55.54\sdev{1.20} \\
 & \ours LAP-N & \ours \textbf{72.77}\sdev{0.19} & \ours \textbf{56.55}\sdev{0.51} & \ours \textbf{49.18}\sdev{0.47} & \ours \textbf{45.05}\sdev{1.32} \\
\midrule
\multirow{2}{*}{AIMv2 ViT-L} & Raw & 82.47\sdev{0.08} & 64.33\sdev{0.17} & 56.03\sdev{0.08} & 50.73\sdev{0.27} \\
 & \ours LAP-L & \ours \textbf{78.13}\sdev{1.20} & \ours \textbf{60.28}\sdev{0.12} & \ours \textbf{52.41}\sdev{0.19} & \ours \textbf{47.19}\sdev{0.34} \\
\midrule
\multirow{2}{*}{CLIP ViT-L} & Raw & 79.63\sdev{0.42} & 60.94\sdev{0.41} & 52.36\sdev{0.09} & 47.43\sdev{0.09} \\
 & \ours LAP-N & \ours \textbf{75.15}\sdev{0.74} & \ours \textbf{57.81}\sdev{0.57} & \ours \textbf{50.05}\sdev{0.30} & \ours \textbf{44.85}\sdev{0.57} \\
\midrule
\multirow{2}{*}{DINOv3 ViT-B} & Raw & 76.38\sdev{1.73} & 58.37\sdev{0.57} & 49.98\sdev{0.57} & 45.70\sdev{0.96} \\
 & \ours LAP-L & \ours \textbf{74.74}\sdev{3.03} & \ours \textbf{57.67}\sdev{1.74} & \ours \textbf{49.35}\sdev{1.63} & \ours \textbf{44.85}\sdev{1.67} \\
\midrule
\multirow{2}{*}{DINOv2 ViT-B} & Raw & \textbf{75.03}\sdev{0.41} & 57.90\sdev{0.37} & \textbf{49.18}\sdev{0.32} & 45.05\sdev{0.62} \\
 & \ours LAP-L & \ours 75.25\sdev{1.24} & \ours \textbf{57.63}\sdev{1.50} & \ours 49.20\sdev{1.46} & \ours \textbf{44.82}\sdev{0.95} \\
\bottomrule
\end{tabular}
\end{table}

\Cref{fig:training_rescue} compares REPA with one LAP variant for MAE, AIMv2, CLIP, and DINOv3. LAP-N is used for MAE and CLIP and LAP-L for AIMv2 and DINOv3, selected by the final FID in \Cref{tab:adm} and held fixed across checkpoints. \Cref{tab:curves_fid10k} adds DINOv2, with LAP-L, and the unaligned reference.

\subsection{Qualitative Samples}
\label{app:qual}

\begin{figure}[p]
\centering
\includegraphics[width=\linewidth]{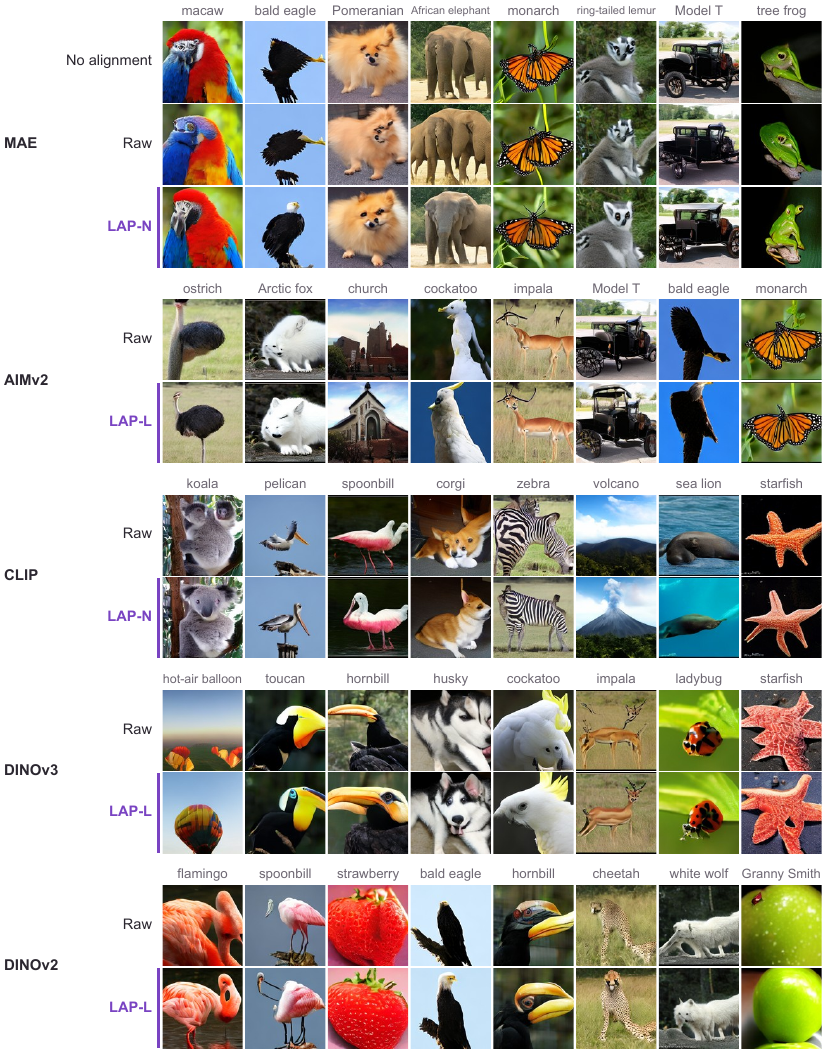}
\caption{\textbf{Samples across teachers} (\Cref{tab:adm}). SiT-B/2, batch 64, SD-VAE, REPA at block 8. Each teacher shows the raw target and the LAP variant of \Cref{tab:adm}. Guidance scale $4.0$; within each comparison, a column shares class and sampling noise across rows (\Cref{app:qual}). Purple labels mark LAP targets.}
\label{fig:qual_teachers}
\end{figure}

\begin{figure}[p]
\centering
\includegraphics[width=\linewidth]{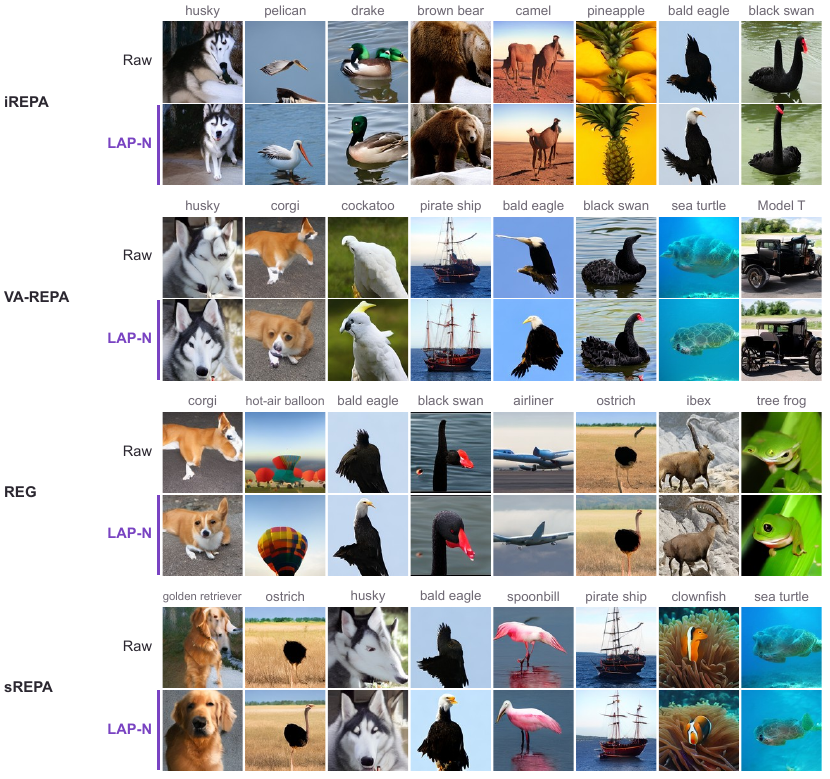}
\caption{\textbf{Samples across REPA-family recipes} (\Cref{tab:compose}). SiT-B/2, batch 64, SD-VAE, MAE ViT-L teacher with the raw target or LAP-N. Guidance scale $4.0$; within each comparison, a column shares class and sampling noise across rows (\Cref{app:qual}). Purple labels mark LAP targets.}
\label{fig:qual_recipes}
\end{figure}

\begin{figure}[p]
\centering
\includegraphics[width=\linewidth]{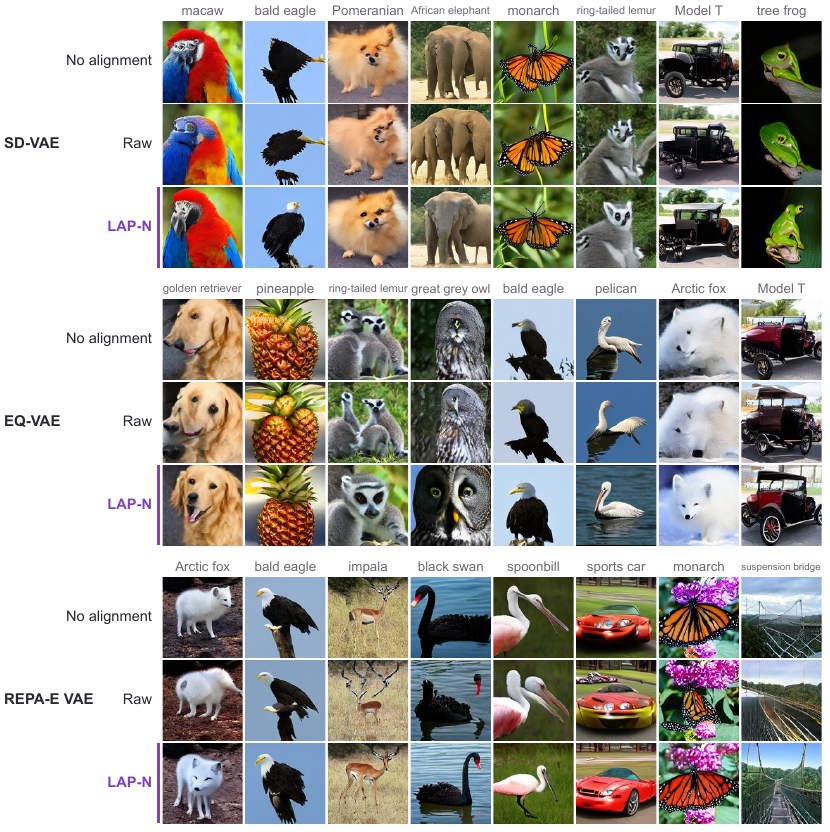}
\caption{\textbf{Samples across frozen tokenizers} (\Cref{tab:tokenizers}). SiT-B/2, batch 64, MAE ViT-L teacher. Guidance scale $4.0$; within each comparison, a column shares class and sampling noise across rows (\Cref{app:qual}). Purple labels mark LAP targets.}
\label{fig:qual_tokenizers}
\end{figure}

\begin{figure}[p]
\centering
\includegraphics[width=\linewidth]{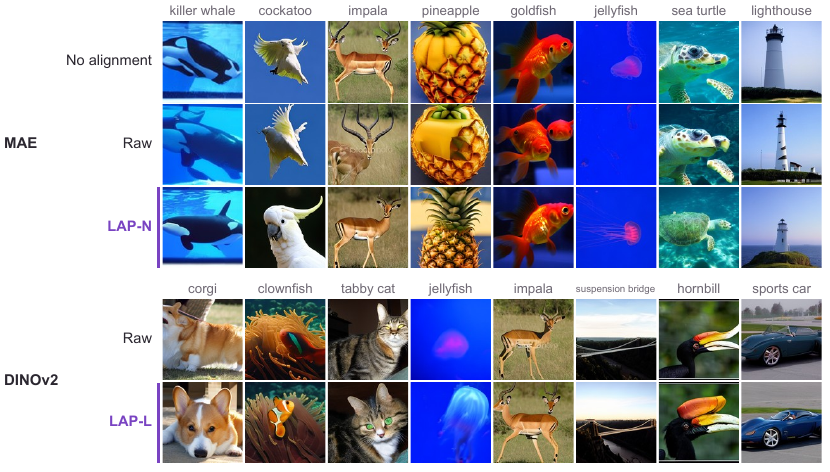}
\caption{\textbf{Samples at SiT-XL/2, batch 256} (\Cref{tab:regime}). MAE ViT-L and DINOv2 ViT-B teachers with the raw target and the LAP variant of \Cref{tab:regime}. Guidance scale $4.0$; within each comparison, a column shares class and sampling noise across rows (\Cref{app:qual}). Purple labels mark LAP targets.}
\label{fig:qual_scale}
\end{figure}

\Cref{fig:qual_teachers,fig:qual_recipes,fig:qual_tokenizers,fig:qual_scale} show samples for the four main comparisons: the five teachers of \Cref{tab:adm}, the REPA-family recipes of \Cref{tab:compose} with the MAE teacher, the frozen tokenizers of \Cref{tab:tokenizers}, and SiT-XL/2 at batch 256 (\Cref{tab:regime}). All models are the 400K EMA checkpoints of training seed 0, sampled with the SDE sampler (250 steps). We use classifier-free guidance at scale $4.0$ for visual clarity; the quantitative results use no guidance or scale $1.5$. Within each comparison, a column shares the class label and all sampling noise across rows, so the models start from the same image layout and differ in shape and detail. These samples illustrate the evaluated models and do not replace the distribution-level metrics.

\FloatBarrier
\section{Computational Resources}
\label{app:compute}

Runs use one A100 unless noted; the REPA-family baseline, CLIP, and tokenizer runs used H100 or A100 GPUs. In the base batch-64 recipe, a 400K-step ImageNet-1k run takes approximately 18--36 hours, depending on the encoder. A 100K-step development run takes about nine hours, and a 50K-step run takes about three to four and a half hours. Fitting one purifier takes about two hours. During diffusion training, LAP-N also evaluates the frozen predictor once per batch in FP32 without gradients. The predictor is discarded for sampling. The experiments do not include matched wall-clock or peak-memory benchmarks, so per-step convergence gains do not establish an advantage in total compute.

Standard 50K-image sampling takes approximately eight to ten A100-hours per model at sampling batch 64. The subsequent ADM metric calculation runs on CPU. Image-disjoint validation, including target extraction and probe fitting, averaged about 40 seconds per frozen checkpoint on one H100 over nine checkpoints.

Batch sizes denote images per optimizer update. For the raw-target and LAP-N SiT-L/2, batch-256 runs with MAE and DINOv2, seed 0 uses one GPU with four accumulation steps of 64 images, whereas seed 1 uses four GPUs with 64 images each and no accumulation. The larger-regime LAP-L runs at effective batch 256 (SiT-B/2, L/2, and XL/2) use four GPUs with 64 images each; the development-scale batch-256 LAP-L runs instead use one GPU with four accumulation steps of 64. LAP-L forms targets on the current local batch, without gathering features across GPUs or accumulation steps. In every reported LAP-L training run, each affine fit therefore uses 64 images, or $64\times256$ token rows. Optimizer steps, EMA updates, and checkpoint counts advance after the effective-batch update.

\FloatBarrier
\section{Two Aerial-Scene Datasets}
\label{app:domains}

The introduction motivates purification by domains in which reconstruction-based encoders are the ones available. We tested this motivation at development scale on two aerial-scene datasets, comparing the ImageNet encoders only; the outcome is consistent with the motivation but does not establish it, and we report it here so that the claim is not read as established.

\textbf{Setup.} NWPU-RESISC45~\citep{cheng2017resisc45}, all 31,500 images in 45 classes at $256\times256$, and AID~\citep{xia2017aid}, 10,000 images in 30 classes downsampled from $600\times600$. Targets: DINOv2 as the off-domain reference, and ImageNet MAE raw and purified by LAP-N with a $k{=}5$ predictor fitted on the dataset's own latents. The recipe is the dev-200 protocol (SiT-B/2, batch 64, 50K steps, $\lambda_a=0.5$), with four training seeds on RESISC45 and two on AID. We report Inception FID between 10K samples (SDE-250, no guidance) and 10K training images of the same dataset; we use Inception features to avoid evaluating with one of the alignment teachers. The same ADM evaluations also provide IS, precision, and recall, reported in \Cref{tab:domains_main}. Means and sample SDs are computed from the unrounded per-seed metrics. 50K steps correspond to about 100 epochs of RESISC45 and, because AID is small, 320 epochs of AID.

\begin{table}[H]
\centering\small
\setlength{\tabcolsep}{5pt}
\caption{\textbf{Development-scale alignment on two aerial-scene datasets.} FID-10K against 10K training images of the same dataset; mean $\pm$ standard deviation over $n$ training seeds. Lower is better. Shaded rows are LAP targets. Bold marks the lowest mean FID within each dataset.}
\label{tab:domains}
\begin{tabular}{llcr}
\toprule
Dataset & Alignment target & $n$ & FID-10K \\
\midrule
\multirow{4}{*}{RESISC45 (45 classes)} & No alignment & 4 & $68.1\pm5.1$ \\
 & DINOv2 & 4 & $66.3\pm3.7$ \\
 & ImageNet MAE, raw & 4 & $64.6\pm4.4$ \\
 & \ours LAP-N (ImageNet MAE) & \ours 4 & \ours $\mathbf{61.5}\pm2.3$ \\
\midrule
\multirow{4}{*}{AID (30 classes)} & No alignment & 2 & $62.4\pm2.1$ \\
 & DINOv2 & 2 & $63.7\pm1.3$ \\
 & ImageNet MAE, raw & 2 & $66.1\pm1.5$ \\
 & \ours LAP-N (ImageNet MAE) & \ours 2 & \ours $\mathbf{60.1}\pm3.3$ \\
\bottomrule
\end{tabular}
\end{table}

\textbf{Outcome.} \Cref{tab:domains} summarizes the FID results. On RESISC45, purified ImageNet MAE has the best mean ($61.5\pm2.3$), about $6.6$ below the unaligned model ($68.1\pm5.1$; Welch $t=2.4$), $4.9$ below off-domain DINOv2 ($66.3\pm3.7$; $t=2.2$), and $3.2$ below raw ImageNet MAE ($t=1.3$). On AID, purification improves ImageNet MAE from $66.1\pm1.5$ to $60.1\pm3.3$ ($t=2.4$), and purified ImageNet MAE is again the best target, but only $2.3$ below the unaligned model ($t=0.8$); DINOv2 does not improve on no alignment there. On both datasets, purified ImageNet MAE is the best aligned target and off-domain DINOv2 is not, which is in the direction of the introduction's motivation. LAP also achieves the highest mean IS and precision on both datasets, and the highest recall on RESISC45. On AID, LAP has slightly lower recall than no alignment ($0.214$ vs.\ $0.222$). The margins over no alignment are at most about two standard deviations over two to four seeds, and the runs are short, so we report the motivation as suggestive rather than established, and say so in the limitations.

\FloatBarrier
\section{Related Work}
\label{app:extrelated}

\subsection{Where and When Representations Guide Generation}

REPA aligns intermediate diffusion features with pretrained visual representations~\citep{yu2025repa}. Related approaches change where that guidance enters the pipeline. REPA-E jointly tunes the VAE and diffusion model~\citep{leng2025repae}, while representation autoencoders use pretrained encoder features as the generative representation space~\citep{zheng2025rae}. SRA and LayerSync derive guidance from the model's own representations or layers~\citep{jiang2025sra,layersync2025}. U-REPA extends alignment to U-Net architectures~\citep{urepa2025}; its multi-resolution matching has no counterpart in the SiT backbone used here, so we do not compare against it.

Other methods change the timing or strength of supervision. HASTE combines early-stopped alignment with attention-map supervision~\citep{wang2025haste}, and VA-REPA adapts alignment using the variance structure of flow matching~\citep{varepa2026}. TARO studies timestep-adaptive alignment for video-to-audio generation~\citep{taro2025}. These approaches motivate studying when alignment helps, whereas our interventions ask which content within a fixed target is useful.

\subsection{Transforming the Target or Matching Objective}

The most directly related work is iREPA~\citep{singh2025irepa}. It compares 27 visual encoders and argues that spatial structure is more predictive of generation performance than global classification accuracy. It also modifies target normalization and the projection head. Our question concerns locally predictable content within a fixed encoder. The interaction experiments in \Cref{app:purifier} test whether the corresponding operations and LAP-N offer additional benefits when combined. They do not establish that spatial structure and local predictability are mutually exclusive explanations.

PixelREPA diagnoses a regression shortcut in pixel-space image transformers and introduces a masked transformer adapter~\citep{pixelrepa2026}. Spectrum Matching studies spectral properties and applies a difference-of-Gaussians transformation to REPA targets~\citep{repadog2026}. SARA and structural representation alignment modify how representation structure is matched~\citep{sara2025,structra2026}. These works show that target processing and the matching objective can matter beyond encoder identity.

Guidance can also come from the generative pipeline. AHPA uses timestep-dependent hierarchical VAE priors~\citep{ahpa2026}, and SRA 2 uses VAE representations~\citep{vaerepa2026}. Representation entanglement introduces representation guidance within the generative architecture~\citep{reg2025}. Our method keeps the pretrained encoder fixed and changes the target supplied to the existing alignment loss.

\subsection{Mechanistic Analysis and Distillation}

AG-REPA uses causal layer interventions in audio flow matching and distinguishes stored information from contribution to the velocity prediction~\citep{agrepa2026}. This is an important precedent for separating decodability from functional contribution. Our study focuses on target content and an auxiliary prediction intervention in image generation, rather than selecting alignment layers.

REED develops a theoretical framework for representation guidance and training curricula~\citep{wang2025reed}. Our linear-Gaussian analysis has a narrower purpose: to illustrate why decodability and predictive contribution can differ. It does not replace a theory of diffusion optimization.

Knowledge-distillation methods such as task-aware layer-wise distillation and variational information distillation adapt what a student learns from a teacher~\citep{liang2023ted,ahn2019vid}. Their objectives are not equivalent to the clean-latent prediction utility measured here. Likewise, linear concept-erasure methods remove specified decodable information under explicit linear criteria~\citep{ravfogel2020inlp,belrose2023leace}. LAP shares the idea of selective removal. Its affine and nonlinear local predictors define operational residuals, without guaranteeing removal of all reconstruction information.

Finally, representation regularization can help without an external encoder~\citep{wang2025dispersive}. The limited self-teacher result in \Cref{app:arc} should therefore be read as a result for that construction, not a general argument against encoder-free guidance.

\FloatBarrier
\Needspace{24\baselineskip}
\section{Individual Runs for the Main Comparisons}
\label{app:seed_results}

This appendix keeps the numerical results behind the main comparisons in one place. Generation columns show individual training seeds. Diagnostic means are identified separately. Two runs describe replication but do not provide a well-calibrated confidence interval. A dash denotes an unavailable measurement.

\subsection{Auxiliary Prediction Sweep}

\begin{figure}[H]
\centering
\setlength{\parskip}{0pt}
\setlength{\abovecaptionskip}{5pt}
\includegraphics[width=\linewidth]{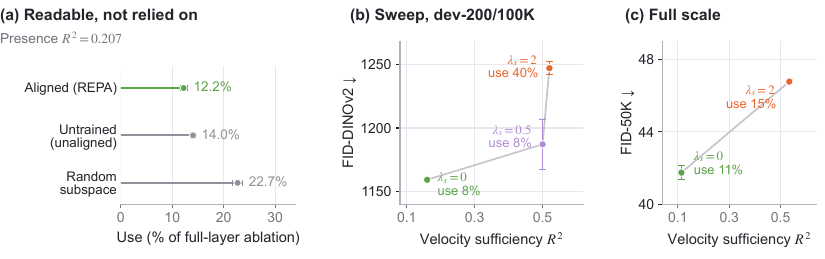}
\vspace{-15pt}
\caption{\textbf{Readability, reliance, and the auxiliary prediction loss.} (a) Rank-256 mean-fill ablations of the REPA model as a percentage of the full-layer effect (ImageNet-1k/400K, seed 0, mid-noise, disjoint fit/evaluation images, 95\% image-bootstrap intervals). (b) The auxiliary prediction loss on dev-200/100K: velocity sufficiency saturates by $\lambda_s=0.5$ while FID-DINOv2 keeps worsening, and reliance on the aligned subspace (``use'') rises only at $\lambda_s=2$. These diagnostics use the token-split protocol. (c) The same loss at full scale, with standard FID-50K and disjoint-image diagnostics. Points are two-seed means and bars span the seed values. Values are in \Cref{tab:f2}.}
\label{fig:causal_tests}
\end{figure}

\Cref{tab:f2} lists the auxiliary prediction-loss runs of \Cref{sec:f2}, and \Cref{fig:causal_tests} plots them. On dev-200, velocity sufficiency saturates by $\lambda_s=0.5$ ($0.50$, against $0.52$ at $\lambda_s=2$), yet mean FID-DINOv2 keeps worsening, from $1159$ to $1187$ and then $1247$. Reliance on the aligned subspace is unchanged up to $\lambda_s=0.5$ ($8.3\%$ and $7.6\%$ of the full-layer effect) and rises to $40\%$ only at $\lambda_s=2$. At full scale it rises from $11\%$ to $15\%$, with seeds at $18.9\%$ and $11.5\%$, while FID-50K worsens in both seeds. The degradation therefore does not require the denoiser to rely more on the supervised directions. For reference, the unmodified REPA model's aligned subspace is decodable (presence $\Rtwo=0.207$), but ablating it increases velocity error by $12.2\%$ of the full-layer effect, less than rank-matched random subspaces at $22.7\%$ (\Cref{fig:causal_tests}(a)).

\begin{table}[H]
\centering\small
\setlength{\tabcolsep}{4.5pt}
\renewcommand{\arraystretch}{1.08}
\caption{\textbf{Auxiliary prediction-loss sweep.} Diagnostics are two-seed means. Dev-200 uses the token-split protocol, while full-scale diagnostics use the image-split protocol. Proxy columns give individual FID-DINOv2 scores. At $\lambda_s=2$, use is $37.6/42.8\%$ on dev-200 and $18.9/11.5\%$ at full scale (seeds 0/1). Standard FID-50K values are in \Cref{tab:seeds_adm}. Bold marks the lowest generation proxy per scale and training seed. Diagnostic columns are not ranked as generation performance.}
\label{tab:f2}
\begin{tabular}{lrrrrr}
\toprule
$\lambda_s$ & Velocity $R^2$ & Align. cosine & Use (\%) & Proxy seed 0 & Proxy seed 1 \\
\midrule
\multicolumn{6}{l}{\emph{Dev-200, 100K steps}} \\
0 & 0.16 & 0.65 & 8.3 & \textbf{1157.3} & \textbf{1161.2} \\
0.5 & 0.50 & 0.63 & 7.6 & 1167.2 & 1207.1 \\
2 & 0.52 & 0.61 & 40.2 & 1242.0 & 1252.3 \\
\midrule
\multicolumn{6}{l}{\emph{ImageNet-1k, 400K steps}} \\
0 & 0.11 & 0.64 & 11.4 & \textbf{825.1} & \textbf{821.1} \\
2 & 0.53 & 0.61 & 15.2 & 879.5 & 877.1 \\
\bottomrule
\end{tabular}
\end{table}

\Needspace{29\baselineskip}
\subsection{Encoder Targets and Purification}

\Cref{tab:ladder} pairs the prediction utility of the five raw encoders with their development-scale generation, and \Cref{tab:dev_ladder} gives the development-scale target ladder.

\begin{table}[H]
\centering\small
\setlength{\tabcolsep}{6pt}
\caption{\textbf{Prediction utility and generation of five raw encoders.} Utility is the clean-latent probe $R^2$ of the raw target. Generation is dev-200/50K FID-DINOv2 with REPA for training seeds 0 and 1 and their mean; these are the points of \Cref{fig:encoder_utility}. The unaligned model scores 1651.2. Bold marks the lowest FID in each column.}
\label{tab:ladder}
\begin{tabular}{lrrrr}
\toprule
Encoder & Utility $R^2$ & Seed 0 & Seed 1 & Mean \\
\midrule
CLIP & 0.18 & 1548.9 & 1553.0 & 1551.0 \\
I-JEPA & 0.29 & \textbf{1529.4} & \textbf{1515.7} & \textbf{1522.6} \\
DINOv2 & 0.33 & 1551.2 & 1518.6 & 1534.9 \\
MoCo v3 & 0.44 & 1584.2 & 1598.0 & 1591.1 \\
MAE & 0.72 & 1641.4 & 1643.2 & 1642.3 \\
\bottomrule
\end{tabular}
\end{table}

\begin{table}[H]
\centering\small
\setlength{\tabcolsep}{6pt}
\caption{\textbf{Development-scale target ladder.} Dev-200/50K FID-DINOv2 with REPA for training seeds 0 and 1 and their mean; the MAE rows are the ladder of \Cref{fig:what_matters}(b). The centered variance explained by the DINOv2 linear split is about $2.7\%$. LAP-P is in \Cref{tab:lapp}, the iREPA operations in \Cref{tab:irepa_interaction}, and FID-50K for these targets in \Cref{tab:adm,tab:seeds_adm}. Shaded rows use LAP; bold marks the lowest FID within each teacher and column.}
\label{tab:dev_ladder}
\begin{tabular}{llrrr}
\toprule
Teacher & Target & Seed 0 & Seed 1 & Mean \\
\midrule
-- & No alignment & 1652.4 & 1649.9 & 1651.2 \\
\midrule
\multirow{3}{*}{MAE} & Raw & 1641.4 & 1643.2 & 1642.3 \\
 & \ours LAP-L & \ours 1504.3 & \ours 1512.3 & \ours 1508.3 \\
 & \ours LAP-N & \ours \textbf{1486.2} & \ours \textbf{1482.6} & \ours \textbf{1484.4} \\
\midrule
\multirow{3}{*}{CLIP} & Raw & 1548.9 & 1553.0 & 1551.0 \\
 & \ours LAP-L & \ours \textbf{1517.1} & \ours 1513.9 & \ours \textbf{1515.5} \\
 & \ours LAP-N & \ours 1537.6 & \ours \textbf{1498.1} & \ours 1517.9 \\
\midrule
\multirow{4}{*}{DINOv2} & Raw & 1551.2 & 1518.6 & 1534.9 \\
 & \ours LAP-L & \ours 1546.0 & \ours \textbf{1487.8} & \ours \textbf{1516.9} \\
 & \ours LAP-N & \ours \textbf{1508.4} & \ours 1536.2 & \ours 1522.3 \\
 & Predictable-only & 1679.9 & 1641.6 & 1660.8 \\
\bottomrule
\end{tabular}
\end{table}

\Needspace{23\baselineskip}
\subsection{Standard FID by Seed}

\Cref{tab:seeds_adm,tab:seeds_adm2} report the standard FID-50K runs behind \Cref{tab:adm,tab:compose,tab:tokenizers,tab:mech_full,tab:adm_full,tab:tok_full}, in the order of the main text. These values use the same base SiT-B/2, batch-64, 400K recipe. The larger-setting evaluations are listed separately in \Cref{tab:regime}.

\begin{table}[H]
\centering\small
\setlength{\tabcolsep}{5pt}
\caption{\textbf{Standard FID-50K by training seed: teachers and interventions.} Individual values behind \Cref{tab:adm,tab:mech_full}; ADM protocol, SDE-250, no guidance. Shaded rows use LAP; bold marks the lowest FID within each teacher and training seed.}
\label{tab:seeds_adm}
\begin{tabular}{llrr}
\toprule
Teacher & Target & Seed 0 & Seed 1 \\
\midrule
-- & No alignment & 53.95 & 53.90 \\
\midrule
\multirow{3}{*}{MAE ViT-L} & Raw & 53.68 & 52.41 \\
 & \ours LAP-L & \ours 44.81 & \ours 45.48 \\
 & \ours LAP-N & \ours \textbf{43.06} & \ours \textbf{41.53} \\
\midrule
\multirow{3}{*}{AIMv2 ViT-L} & Raw & 48.08 & 47.87 \\
 & \ours LAP-L & \ours 44.40 & \ours \textbf{44.22} \\
 & \ours LAP-N & \ours \textbf{43.98} & \ours 45.14 \\
\midrule
\multirow{3}{*}{CLIP ViT-L} & Raw & 44.73 & 44.74 \\
 & \ours LAP-L & \ours \textbf{42.08} & \ours 43.30 \\
 & \ours LAP-N & \ours 42.32 & \ours \textbf{42.11} \\
\midrule
\multirow{3}{*}{DINOv3 ViT-B} & Raw & \textbf{42.12} & 43.34 \\
 & \ours LAP-L & \ours 43.23 & \ours \textbf{40.92} \\
 & \ours LAP-N & \ours 43.10 & \ours 43.44 \\
\midrule
\multirow{3}{*}{DINOv2 ViT-B} & Raw & 42.14 & \textbf{41.35} \\
 & \ours LAP-L & \ours \textbf{40.61} & \ours 42.33 \\
 & \ours LAP-N & \ours 41.90 & \ours 42.33 \\
\midrule
\multicolumn{4}{l}{\emph{Interventions of \Cref{sec:mechanism}, DINOv2 ViT-B}} \\
\multicolumn{2}{l}{REPA $+$ prediction loss} & 46.67 & 46.90 \\
\multicolumn{2}{l}{Predictable-only target} & 57.00 & 56.42 \\
\bottomrule
\end{tabular}
\end{table}

\begin{table}[H]
\centering\small
\setlength{\tabcolsep}{5pt}
\caption{\textbf{Standard FID-50K by training seed: recipes and tokenizers.} Individual values behind \Cref{tab:compose,tab:tokenizers,tab:adm_full,tab:tok_full}; same protocol as \Cref{tab:seeds_adm}. Shaded rows use LAP; bold marks the lowest FID within each teacher--recipe or tokenizer group and training seed.}
\label{tab:seeds_adm2}
\begin{tabular}{llrr}
\toprule
Setting & Target & Seed 0 & Seed 1 \\
\midrule
\multicolumn{4}{l}{\emph{REPA-family recipes, MAE ViT-L}} \\
\multirow{3}{*}{iREPA} & Raw & 43.50 & 42.62 \\
 & \ours LAP-L & \ours 43.38 & \ours 42.81 \\
 & \ours LAP-N & \ours \textbf{41.12} & \ours \textbf{41.52} \\
\cmidrule(lr){1-4}
\multirow{3}{*}{VA-REPA} & Raw & 53.46 & 52.19 \\
 & \ours LAP-L & \ours 42.98 & \ours 43.88 \\
 & \ours LAP-N & \ours \textbf{42.24} & \ours \textbf{40.95} \\
\cmidrule(lr){1-4}
\multirow{2}{*}{REG} & Raw & 53.29 & 52.54 \\
 & \ours LAP-N & \ours \textbf{41.70} & \ours \textbf{41.61} \\
\cmidrule(lr){1-4}
\multirow{2}{*}{sREPA} & Raw & 50.06 & 50.80 \\
 & \ours LAP-N & \ours \textbf{38.90} & \ours \textbf{38.02} \\
\midrule
\multicolumn{4}{l}{\emph{REPA-family recipes, DINOv2 ViT-B}} \\
\multirow{3}{*}{iREPA} & Raw & 41.45 & 41.18 \\
 & \ours LAP-L & \ours \textbf{40.28} & \ours 40.83 \\
 & \ours LAP-N & \ours 41.49 & \ours \textbf{40.57} \\
\cmidrule(lr){1-4}
\multirow{3}{*}{VA-REPA} & Raw & 40.81 & 40.61 \\
 & \ours LAP-L & \ours \textbf{40.37} & \ours \textbf{40.40} \\
 & \ours LAP-N & \ours 42.33 & \ours 41.44 \\
\cmidrule(lr){1-4}
\multirow{3}{*}{REG} & Raw & 28.66 & 29.82 \\
 & \ours LAP-L & \ours \textbf{28.48} & \ours \textbf{28.85} \\
 & \ours LAP-N & \ours 29.37 & \ours 29.64 \\
\cmidrule(lr){1-4}
\multirow{3}{*}{sREPA} & Raw & 38.14 & 38.13 \\
 & \ours LAP-L & \ours \textbf{37.03} & \ours 38.24 \\
 & \ours LAP-N & \ours 37.21 & \ours \textbf{36.62} \\
\midrule
\multicolumn{4}{l}{\emph{Frozen tokenizers, MAE ViT-L}} \\
\multirow{3}{*}{EQ-VAE} & No alignment & 52.47 & 52.53 \\
 & Raw & 51.44 & 53.38 \\
 & \ours LAP-N & \ours \textbf{40.19} & \ours \textbf{39.83} \\
\cmidrule(lr){1-4}
\multirow{3}{*}{REPA-E VAE} & No alignment & 32.27 & 32.33 \\
 & Raw & 33.42 & 34.06 \\
 & \ours LAP-N & \ours \textbf{27.84} & \ours \textbf{27.28} \\
\midrule
\multicolumn{4}{l}{\emph{Frozen tokenizers, DINOv2 ViT-B}} \\
\multirow{3}{*}{REPA-E VAE} & Raw & 25.88 & 26.33 \\
 & \ours LAP-L & \ours 25.98 & \ours 25.39 \\
 & \ours LAP-N & \ours \textbf{25.03} & \ours \textbf{24.85} \\
\bottomrule
\end{tabular}
\end{table}